\documentclass{article} 
\usepackage{iclr2024_conference,times}

\usepackage{amsmath,amsfonts,bm}

\def\eqref#1{equation~\ref{#1}}

\def\1{\bm{1}}

\DeclareMathAlphabet{\mathsfit}{\encodingdefault}{\sfdefault}{m}{sl}
\SetMathAlphabet{\mathsfit}{bold}{\encodingdefault}{\sfdefault}{bx}{n}

\usepackage{hyperref}
\usepackage{url}
\usepackage{graphicx}

\usepackage[utf8]{inputenc} 
\usepackage[T1]{fontenc}    
\usepackage{hyperref}       
\usepackage{url}            
\usepackage{booktabs}       
\usepackage{amsfonts}       
\usepackage{nicefrac}       
\usepackage{microtype}      
\usepackage{xcolor}         

\usepackage{amsmath}
\usepackage{amssymb}
\usepackage{mathtools}
\usepackage{amsthm}

\usepackage{wrapfig}

\usepackage{algorithm}
\usepackage[noend]{algpseudocode}

\usepackage[capitalize,noabbrev]{cleveref}

\theoremstyle{plain}

\theoremstyle{definition}

\theoremstyle{remark}

\hypersetup{
    colorlinks=true,
    linkcolor=blue,
    filecolor=magenta,      
    urlcolor=cyan,
    citecolor=cyan,
}

\definecolor{mygreen}{HTML}{00cf00}
\definecolor{myblue}{HTML}{0000cf}
\definecolor{myorange}{HTML}{e74e00}
\definecolor{mypurple}{HTML}{CC0066}
\definecolor{myred}{HTML}{A60033}

\usepackage{url}            
\usepackage{booktabs}       
\usepackage{amsfonts}       
\usepackage{nicefrac}       
\usepackage{microtype}      

\usepackage{float}
\usepackage{mathtools}
\usepackage{amsthm}
\usepackage{amssymb}
\usepackage{amsfonts}
\usepackage{graphicx}
\usepackage{booktabs}
\usepackage{siunitx}
\usepackage{arydshln}
\usepackage{caption}
\usepackage{subcaption}
\usepackage{soul}
\usepackage{epstopdf}
\usepackage{fancyvrb}
\usepackage{cancel}
\usepackage{makecell}
\usepackage{graphicx}

\newcommand{\boldtheta}{{\boldsymbol{\theta}}}
\newcommand{\boldepsilon}{\boldsymbol{\epsilon}}
\newcommand{\boldxi}{\boldsymbol{\xi}}
\newcommand{\boldphi}{\boldsymbol{\phi}}

\newcommand{\boldmu}{\boldsymbol{\mu}}

\newcommand{\boldx}{\mathbf{x}}

\newcommand{\bolda}{\mathbf{a}}

\newcommand{\boldc}{\mathbf{c}}
\newcommand{\boldA}{\mathbf{A}}
\newcommand{\boldB}{\mathbf{B}}

\newcommand{\bolds}{\mathbf{s}}

\newcommand{\boldh}{\mathbf{h}}

\newcommand{\boldg}{\boldsymbol{g}}

\newcommand{\boldI}{\mathbf{I}}

\newcommand{\boldW}{\mathbf{W}}

\newcommand{\boldzero}{\boldsymbol{0}}

\newcommand{\loss}{J}

\title{Directly Fine-Tuning Diffusion Models on Differentiable Rewards}

\author{
Kevin Clark$^*$, Paul Vicol$^*$, Kevin Swersky, David J. Fleet\\
\hspace{1mm}Google DeepMind \quad $^*$Equal contribution\\
\texttt{\{kevclark, paulvicol, kswersky, davidfleet\}}@google.com\\
}

\iclrfinalcopy
\begin{document}

\maketitle

\begin{abstract}
\vspace{-0.1cm}
We present Direct Reward Fine-Tuning (DRaFT), a simple and effective method for fine-tuning diffusion models to maximize differentiable reward functions, such as scores from human preference models.
We first show that it is possible to backpropagate the reward function gradient through the full sampling procedure, and that doing so achieves strong performance on a variety of rewards, outperforming reinforcement learning-based approaches.
We then propose more efficient variants of DRaFT: DRaFT-$K$, which truncates backpropagation to only the last $K$ steps of sampling, and DRaFT-LV, which obtains lower-variance gradient estimates for the case when $K=1$.
We show that our methods work well for a variety of reward functions and can be used to substantially improve the aesthetic quality of images generated by Stable Diffusion 1.4.
Finally, we draw connections between our approach and prior work, providing a unifying perspective on the design space of gradient-based fine-tuning algorithms.
\vspace{-0.1cm}
\end{abstract}

\begin{figure}[b]
    \vspace{-0.1cm}
    \centering
    \includegraphics[width=0.97\linewidth]{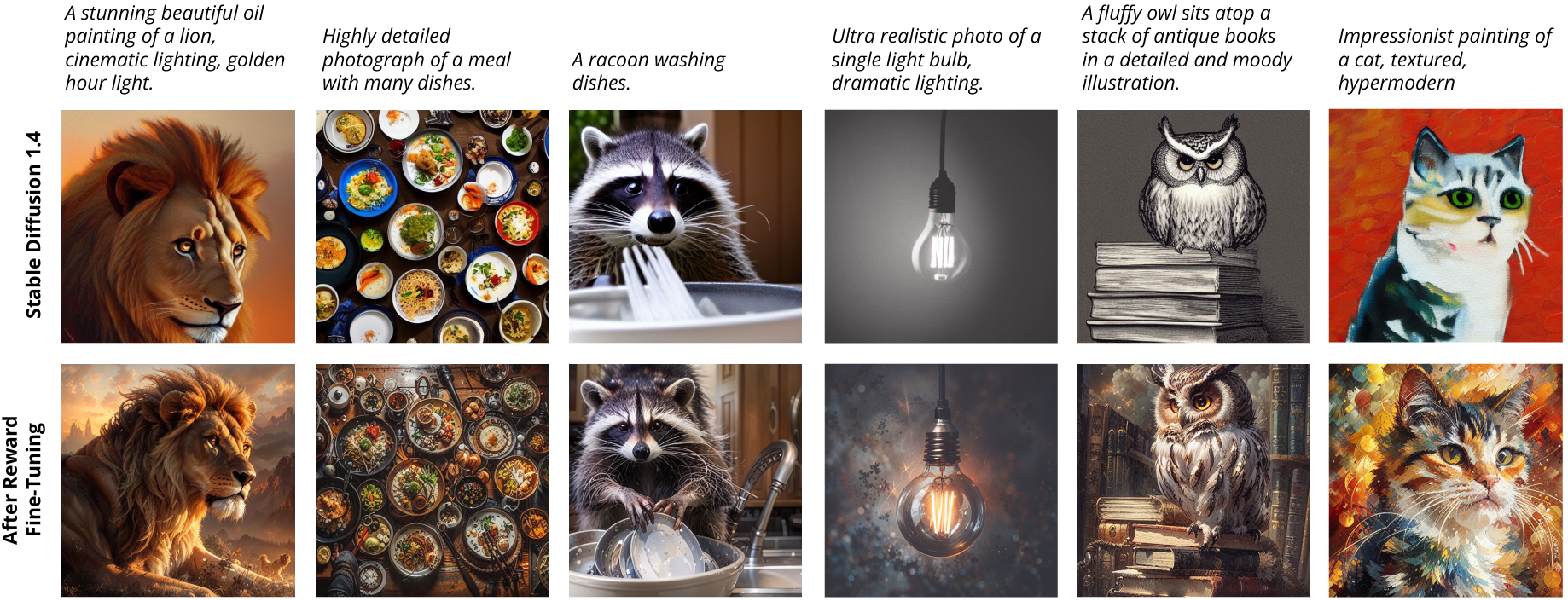}
    \vspace{-0.1cm}
    \caption{\small DRaFT fine-tuning for human preference reward models yields more detailed and stylized images than baseline Stable Diffusion. See Figure~\ref{fig:uncurated-combined} in Appendix~\ref{app:uncurated-samples} for more samples.}
    \label{fig:main-figure}
\end{figure}

\vspace{-0.1cm}
\section{Introduction}
\label{sec:introduction}
\vspace{-0.1cm}

Diffusion models~\citep{sohl2015deep,song2019generative,ho2020denoising,song2020score,kingma2021variational} have revolutionized generative modeling for continuous data, achieving impressive results across modalities including images, videos, and audio.
However, for many use cases, modeling the training data distribution exactly (e.g. diverse images from the web) does not align with the model's desired behavior (e.g. generating aesthetically pleasing images). 
To overcome this mismatch, text-to-image diffusion models commonly employ methods like classifier-free guidance \citep{ho2022classifier} to improve image-text alignment, and are fine-tuned on curated datasets such as LAION Aesthetics \citep{laionaesthetics2022} to improve image quality.  

Within the language modeling community, aligning model behavior with human preferences is widely accomplished through fine-tuning with reinforcement learning (RL), such as with RL from human feedback (RLHF; \citealt{ziegler2019fine,stiennon2020learning,ouyang2022training,bai2022training}).
Inspired by RLHF, supervised and reinforcement learning methods have been developed to train diffusion models on human preference models or other rewards \citep{lee2023aligning,wu2023better,dong2023raft,black2023training,fan2023dpok}. 
While showing some success at improving text alignment or image aesthetics on narrow domains, these approaches are sample inefficient and have not been scaled up to the level of generally improving generation style across diverse prompts.

We propose a simple and efficient approach for gradient-based reward fine-tuning based on differentiating through the diffusion sampling process.
We first introduce Direct Reward Fine-Tuning (DRaFT), a conceptually straightforward method that backpropagates the reward through the full sampling chain.
To keep memory and compute costs low, we use gradient checkpointing~\citep{chen2016training} and optimize LoRA weights~\citep{hu2021lora} rather than the full set of model parameters.
Then, we introduce modifications to DRaFT that further improve its efficiency and performance.
First, we propose DRaFT-$K$, a variant that backpropagates through only the last $K$ steps of sampling to compute the gradient.
We show empirically that full backpropagation can result in exploding gradients, and that using the truncated gradient performs substantially better given the same number of training steps.
Then, we further improve efficiency by introducing DRaFT-LV, a variant of DRaFT-1 that computes lower-variance gradient estimates by averaging over multiple noise samples.

We apply DRaFT to Stable Diffusion 1.4 \citep{rombach2022high} and evaluate it on a variety of reward functions and prompt sets.
As our methods leverage gradients, they are substantially more efficient than RL-based fine-tuning baselines, for example maximizing scores from the LAION Aesthetics Classifier \citep{laionaesthetics2022} $>\!200\times$ faster than the RL algorithms from \citet{black2023training}.
DRaFT-LV is particularly efficient, learning roughly $2 \times$ faster than ReFL \citep{xu2023imagereward}, a previous gradient-based fine-tuning method.
We scale up DRaFT to train on hundreds of thousands of prompts and show that it improves Stable Diffusion on human preference rewards such as PickScore \citep{kirstain2023pick} and Human Preference Score v2 \citep{wu2023human}.
We further show that we can interpolate between the pre-trained and fine-tuned models by scaling the LoRA weights, and we can combine the effects of multiple rewards by linearly combining fine-tuned LoRA weights.
Finally, we showcase how DRaFT can be applied to diverse reward functions including image compressibility and incompressibility, object detection and removal, and generating adversarial examples.

\vspace{-0.2cm}
\section{Related Work}
\label{sec:related-work}
\vspace{-0.2cm}

We provide an extended discussion of related work in Appendix~\ref{app:extended-related-work}.

\vspace{-0.4cm}
\paragraph{Learning human preferences.} 
Human preference learning trains models on judgements of which behaviors people prefer, rather than on human demonstrations directly \citep{knox2009interactively,akrour2011preference}.
This training is commonly done by learning a reward model reflecting human preferences and then learning a policy that maximizes the reward \citep{christiano2017deep,ibarz2018reward}. 
We apply DRaFT to optimize scores from existing preference models, such as PickScore \citep{kirstain2023pick} and Human Preference Score v2 \citep{wu2023human}, which are trained on human judgements between pairs of images generated by diffusion models for the same prompt.

\textbf{Guidance}~\citep{song2020score,dhariwal2021diffusion}
steers sampling towards images that satisfy a desired objective by adding an auxiliary term to the score function.
Various pre-trained models can be used to provide guidance signals, including classifiers, facial recognition models, and object detectors~\citep{bansal2023universal}.
Noisy examples obtained during sampling may be out-of-distribution for the guidance model.
Two approaches to mitigate this are: (1) training the guidance model on noisy data~\citep{dhariwal2021diffusion} and (2) applying a guidance model to predicted clean images from one-step denoising~\citep{li2022upainting}.
However, (1) precludes the use of off-the-shelf pre-trained guidance functions and (2) applies guidance to out-of-distribution blurry images for high noise levels.
DRaFT avoids these issues by only applying the reward function to the final generated image. 
\vspace{-0.2cm}
\paragraph{Backpropagation through diffusion sampling.}

\citet{watson2022learning} differentiate through sampling to learn sampler hyperparameters (rather than model parameters) in order to reduce inference cost while maintaining perceptual image quality.
\citet{fan2023optimizing} also propose an approach to speed up sampling: they consider doing this by backpropagating through diffusion sampling, but focus on an RL-based approach due to memory and exploding/vanishing gradient concerns. Both~\citet{fan2023optimizing} and~\citet{watson2022learning} aim to improve sampling speed, rather than optimize arbitrary differentiable rewards, which is our focus.
Like DRaFT, Direct Optimization of Diffusion Latents (DOODL; \citealt{wallace2023end}) backpropagates through sampling to optimize differentiable rewards; however DOODL optimizes the \textit{initial noise sample} $\boldx_T$ rather than the model parameters.
While this does not require a training phase, it is much slower at inference time because the optimization must be redone for each prompt and metric.
Outside of image generation, \citet{wang2023diffusion} train diffusion model policies for reinforcement learning by backpropagating through sampling. 

\vspace{-0.2cm}
\paragraph{Reward Fine-Tuning with Supervised Learning.}
\citet{lee2023aligning} and \citet{wu2023better} use supervised approaches to fine-tune diffusion models on rewards. These methods generate images with the pre-trained model and then fine-tune on the images while weighting examples according to the reward function or discarding low-reward examples. Unlike DRaFT or RL methods, the model is not trained online on examples generated by the current policy.
However, \citet{dong2023raft} use an online version of this approach where examples are re-generated over multiple rounds of training, which can be viewed as a simple kind of reinforcement learning.

\vspace{-0.2cm}
\paragraph{Reward Fine-Tuning with Reinforcement Learning.}
\citet{fan2023optimizing} interpret the denoising process as a multi-step decision-making task and use policy gradient algorithms to fine-tune diffusion samplers.
Building on it, \citet{black2023training} and~\citet{fan2023dpok} use policy gradient algorithms to fine-tune diffusion models for arbitrary black-box objectives.
Rather than optimizing model parameters, \citet{hao2022optimizing} apply RL to improve the input prompts.
RL approaches are flexible because they do not require differentiable rewards. However, in practice many reward functions are  differentiable, or can be implemented or approximated in a differentiable way, and thus analytic gradients are often available. In such cases, using reinforcement learning discards useful information.

\vspace{-0.2cm}
\paragraph{Reward Feedback Learning (ReFL).}
ReFL \citep{xu2023imagereward} uses reward function gradients to fine-tune a diffusion model.
It evaluates the reward on the one-step predicted clean image, $r(\hat{\boldx}_0, \boldc)$ from a randomly-chosen step $t$ along the denoising trajectory rather than on the final image, as DRaFT does.
While ReFL performs similarly to DRaFT-1, DRaFT-LV is substantially more efficient, training approximately $2 \times$ faster.
We further discuss how ReFL relates to DRaFT in Section~\ref{sec:method}.

\vspace{-0.2cm}
\section{Background on Diffusion Models}
\label{sec:task-setup}
\vspace{-0.2cm}

\paragraph{Diffusion Models.}
Diffusion models are latent variable models based on the principle of iterative denoising: samples are generated starting from pure noise $\boldx_T \sim \mathcal{N}(\boldzero, \boldI)$ via a sequence of applications of a learned denoising function $\boldepsilon_{\boldtheta}$ that gradually removes noise over $T$ timesteps.
Diffusion models often learn the denoiser $\boldepsilon_{\boldtheta}$ by minimizing the following re-weighted variational lower bound of the marginal likelihood~\citep{ho2020denoising}:
\begin{align}  \label{eq:simple}
\mathcal{L}_{\text{Simple}}(\boldtheta)
=
\mathbb{E}_{t \sim U(0, T), \boldx_0 \sim p_{\text{data}}, \boldepsilon \sim \mathcal{N}(\boldzero, \boldI)} \left[ \| \boldepsilon - \boldepsilon_{\boldtheta}\left( \alpha_t \boldx_0 + \sigma_t \boldepsilon, t \right) \|^2 \right]
\end{align}
where $\sigma_t = \sqrt{1 - \alpha^2_t}$ is an increasing noise schedule.\footnote{We use VDM notation \citep{kingma2021variational}, where $\alpha_t$ corresponds to $\sqrt{\bar{\alpha}_t}$ in the \citet{ho2020denoising} notation.}
Here, $\boldepsilon_{\boldtheta}$ is trained to predict the noise $\boldepsilon$ added to the clean datapoint $\boldx_0$.
For image data, $\boldepsilon_{\boldtheta}$ is typically parameterized by a UNet~\citep{ronneberger2015u}.
We use conditional diffusion models, which include a context $\boldc$ such as a text prompt passed to the denoising function $\boldepsilon_{\boldtheta}(\boldx_t, \boldc, t)$.
For inference (i.e.,\@ sampling), one draws a noise sample $\boldx_T$, and then iteratively uses $\boldepsilon_{\theta}$ to estimate the noise and compute the next latent sample $\boldx_{t - 1}$.

\vspace{-0.2cm}
\paragraph{Classifier-Free Guidance.}
Classifier-free guidance (CFG) uses a linear combination of the conditional and unconditional score estimates, denoising according to $(1 + w(t))\boldepsilon(\boldx_t, \boldc, t) - w(t)\boldepsilon(\boldx_t, \varnothing, t)$ where $w(t)$ is the guidance weight and $\varnothing$ indicates an empty conditioning signal.
It requires that $\boldc$ is replaced with $\varnothing$ part of the time during training so that the unconditional score function is learned.

\vspace{-0.2cm}
\section{Method}
\label{sec:method}
\vspace{-0.2cm}

\begin{figure}[t]
    \centering
    \includegraphics[width=0.95\linewidth]{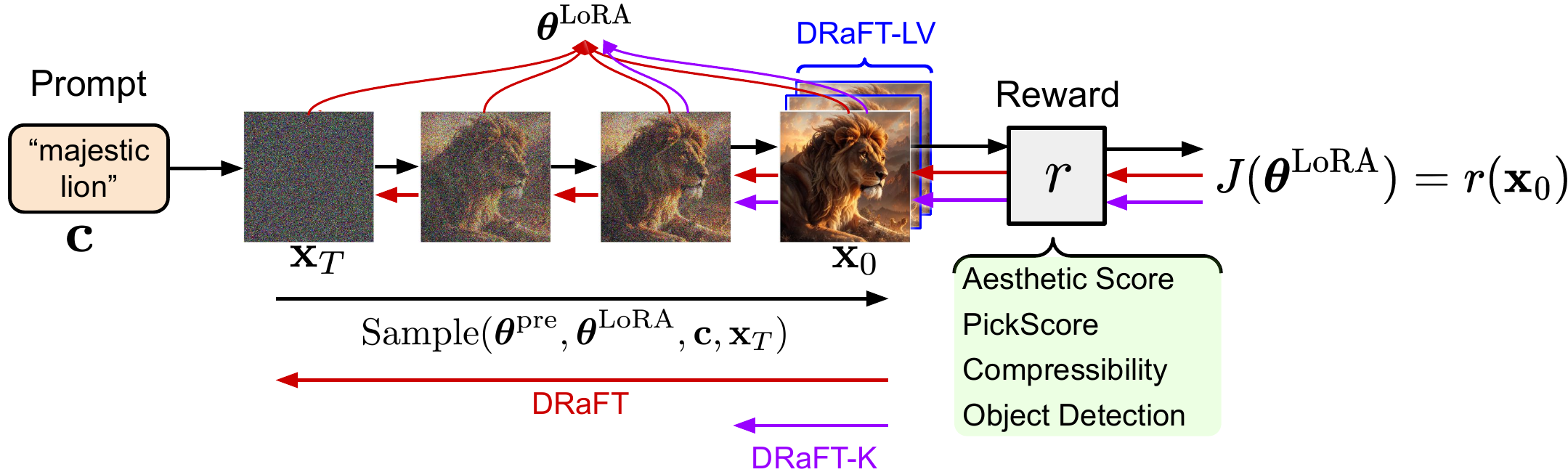}
    \vspace{-0.2cm}
    \caption{\small {\color{myred}DRaFT} backpropagates reward gradients through the sampling process into LoRA weights. {\color{mypurple}DRaFT-$K$} truncates the backward pass, differentiating through only the last $K$ steps of sampling. {\color{myblue}DRaFT-LV} improves the efficiency of DRaFT-1 by generating multiple examples in the last step to reduce variance.}
    \label{fig:method}
    \vspace{-0.4cm}
\end{figure}

We propose a simple approach for fine-tuning diffusion models for differentiable reward functions.
Our goal is to fine-tune the parameters $\boldtheta$ of a pre-trained diffusion model such that images generated by the sampling process maximize a differentiable reward function $r$: 
\begin{align} \label{eq:reward-objective}
\loss(\boldtheta) = \mathbb{E}_{\boldc \sim p_{\boldc}, \boldx_T \sim \mathcal{N}(\boldzero, \boldI)} \left[ r(\text{sample}(\boldtheta, \boldc, \boldx_T), \boldc) \right]
\end{align}
where $\text{sample}(\boldtheta, \boldc, \boldx_T)$ denotes the sampling process from time $t = T \to 0$ with context $\boldc$.
\vspace{-0.2cm}
\paragraph{DRaFT.}
First, we consider solving Eq.~\ref{eq:reward-objective} by computing $\nabla_{\boldtheta} r(\text{sample}(\boldtheta, \boldc, \boldx_T), \boldc)$ and using gradient ascent.
Computing this gradient requires backpropagation through multiple diffusion model calls in the sampling chain, similar to backpropagation through time in a recurrent neural network.
We use two approaches to reduce the memory cost of DRaFT: 1) low-rank adaptation (LoRA)~\citep{hu2021lora}; and 2) gradient checkpointing~\citep{chen2016training}.

\vspace{-0.2cm}
\paragraph{LoRA.}
Rather than fine-tuning the full set of model parameters, Low-Rank Adaptation (LoRA)~\citep{hu2021lora} freezes the weights of the pre-trained model, and injects new low-rank weight matrices alongside the original model weights, whose contributions are summed to produce the adapted model outputs.
Mathematically, for a layer with base parameters $\boldW_0$ whose forward pass yields $\boldh = \boldW_0 \boldx$, the LoRA adapted layer is $\boldh = \boldW_0 \boldx + \boldB \boldA \boldx$, where $\boldB \boldA$ is a low-rank matrix.
LoRA dramatically reduces the number of parameters that need to be optimized, which reduces the memory requirements of fine-tuning.
Another benefit is that, because the information learned by the fine-tuned model is contained within the LoRA parameters, we can conveniently combine fine-tuned models by taking linear combinations of the LoRA parameters (Figure~\ref{fig:reward-comparison}, right) and interpolate between the original and fine-tuned model by re-scaling the LoRA parameters (Figure~\ref{fig:lora-scaling}).

\begin{wrapfigure}[28]{l}{0.49\linewidth}
\vspace{-0.7cm}
\begin{minipage}{\linewidth}
\begin{algorithm}[H]
  \caption{$\text{DRaFT (with DDIM sampling)}$
  }
  \label{alg:sampling}
    \begin{algorithmic}
    \State \textbf{Inputs:} pre-trained diffusion model weights $\boldtheta$, reward $r$, prompt dataset $p_{\boldc}$
    \While {not converged}
        \State $t_\text{truncate} = \begin{cases}
          \text{randint}(1, m)  & \text{\textbf{if} \color{myorange} ReFL} \\
          T  & \text{\textbf{if} \color{myred} DRaFT} \\
          K  & \text{\textbf{if} \color{mypurple} DRaFT-$K$} \\
          1  & \text{\textbf{if} \color{myblue} DRaFT-LV} \\
        \end{cases}$
        
        \State $\boldc \sim p_{\boldc}$, $\boldx_T \sim \mathcal{N}(\boldzero, \boldI)$   %
        
        \For {$t = T, \dots, 1$}
            \If {{\color{mypurple} $t = t_\text{truncate}$}}
                \State {\color{mypurple} $\boldx_{t} = \texttt{stop\_grad}(\boldx_{t})$}
            \EndIf
            \State \smash{$\hat{\boldx}_0 = (\boldx_t - \sigma_t \boldepsilon_{\boldtheta}(\boldx_t, \boldc, t)) / \alpha_t$}
            \If {{\color{myorange} $t = t_\text{truncate}$ and ReFL}}
                \State {\color{myorange} $\boldx_0 \approx \hat{\boldx}_0$}
                \State {\color{myorange} \textbf{break}}
            \EndIf
            \State $\boldx_{t-1} = \alpha_{t - 1}\hat{\boldx}_0 + \sigma_{t - 1}\boldepsilon_{\boldtheta}(\boldx_t, \boldc, t)$
        \EndFor
        \State $\boldg = \nabla_{\boldtheta} r(\boldx_0, \boldc)$ 
        \If {{\color{myblue} DRaFT-LV}}
            {\color{myblue}
            \For {$i = 1, \dots, n$}
                \State $\boldepsilon \sim \mathcal{N}(\boldzero, \boldI)$
                \State $\boldx_1 = \alpha_1 \texttt{stop\_grad}(\boldx_0) + \sigma_1 \boldepsilon$
                \State \smash{$\hat{\boldx}_0 = (\boldx_1 - \sigma_1 \boldepsilon_{\boldtheta}(\boldx_1, \boldc, 1)) / \alpha_1$}
                \State $\boldg = \boldg + \nabla_{\boldtheta} r(\hat{\boldx}_0, \boldc)$
            \EndFor
            }
        \EndIf
        \State $\boldtheta \gets \boldtheta - \eta \boldg$ 
    \EndWhile
    \State \textbf{return} $\boldtheta$
    \end{algorithmic}
\end{algorithm}
\end{minipage}
\end{wrapfigure}
\vspace{-0.4cm}
\paragraph{Gradient Checkpointing.}
Gradient checkpointing~\citep{gruslys2016memory,chen2016training} reduces the memory cost of storing activations for use during backprop at the cost of increased compute, by storing only a subset of activations in memory and re-computing the others on the fly.
Specifically, we only store the input latent for each denoising step, and re-materialize the UNet activations during backprop.
Implementing checkpointing in JAX~\citep{jax2018github} is as simple as adding \texttt{@jax.checkpoint} to the body of the sampling loop.

\vspace{-0.2cm}
\paragraph{DRaFT-$K$.}
While gradient checkpointing makes it possible to backpropagate through the full sampling chain, we found that optimization speed and overall performance can be substantially improved by truncating backprop through only the last $K$ sampling steps.
We call the resulting approach {\color{mypurple}DRaFT-$K$}, illustrated in Figure~\ref{fig:method}.
Truncating backprop reduces the compute needed per step by decreasing the number of backward passes through the UNet.
Surprisingly, it also improves training efficiency per-step, which we investigate further in Section~\ref{sec:understanding-k}.
For small $K$ (e.g., $K=1$), the memory cost of unrolling is small, and we do not need gradient checkpointing.

\vspace{-0.2cm}
\paragraph{DRaFT-LV.}
Empirically, we found that simply setting $K=1$ (i.e.,\@ only differentiating through the last sampling step) results in the best reward vs. compute tradeoff. 
Here, we propose a method for improving the efficiency of DRaFT-1 further by reducing the variance of the gradient estimate.
We call this low-variance estimator {\color{myblue}DRaFT-LV}.
The key idea is to use the forward diffusion process to produce additional examples to train on without re-generating new images.
Specifically, we noise the generated image $n$ times, and use the summed reward gradient over these examples.
Although DRaFT-LV adds $n$ additional forward and backward passes through the UNet and reward model, in practice this is fairly little overhead compared to the $T$ UNet calls already needed for sampling.
We found that using $n=2$ is around $2\times$ more efficient than DRaFT-1 while adding around 10\% compute overhead for our reward functions.
 
\vspace{-0.2cm}
\paragraph{General Reward Fine-Tuning Algorithm.}
Algorithm~\ref{alg:sampling} presents a unified framework encompassing several gradient-based approaches for diffusion fine-tuning.
In particular, it includes as special cases DRaFT, DRaFT-$K$, DRaFT-LV, and ReFL~\citep{xu2023imagereward}, which differ in how the gradient is computed.
{\color{mypurple} DRaFT-$K$} inserts a stop-gradient operation $\boldx_t = \texttt{stop\_grad}(\boldx_t)$ at sampling iteration $K$, to ensure that the gradient is only computed through the last $K$ steps (as the gradient does not flow through \texttt{stop\_grad}); vanilla DRaFT can be obtained by setting $K = T$.
{\color{myorange}ReFL} inserts a stop gradient and breaks out of the sampling loop early, returning a one-step predicted clean image. Additionally, it does so at a random timestep towards the end of sampling (we use $m=20$ for 50-step sampling).
In this unifying framework, {\color{myorange}ReFL} with $m=1$ is equivalent to {\color{mypurple}DRaFT-$1$}.
Lastly, {\color{myblue}DRaFT-LV} averages the last-step gradient over multiple examples with different noises to reduce variance.
Note that in practice we use Adam and LoRA, not shown in the algorithm for simplicity.

\vspace{-0.2cm}
\section{Experiments}
\label{sec:experiments}
\vspace{-0.2cm}

In this section, we show that DRaFT can be applied to a wide range of reward functions, and that it substantially outperforms prior reward fine-tuning methods.
We used Stable Diffusion 1.4 as the base diffusion model.
As it is a latent diffusion model \citep{rombach2022high}, applying DRaFT involves backpropagating through both the sampling process that produces the final latent representation and the decoder used to obtain the image.
DRaFT is applicable to any differentiable sampler; in our experiments, we used DDIM \citep{song2020denoising}, but we found that ancestral sampling performs similarly. We used 50 sampling steps, so DRaFT-50 corresponds to backpropagating through the full sampling chain (i.e.,\@ vanilla DRaFT).
We used classifier-free guidance weight 7.5.
Experimental details, hyperparameters, and additional results are provided in Appendices~\ref{app:exp-details} and~\ref{app:additional-results}.

\vspace{-0.2cm}
\subsection{Fine-Tuning for Aesthetic Quality}
\label{sec:aesthetics}
\vspace{-0.2cm}

\begin{figure}[h]
    \vspace{-0.3cm}
    \centering
    \begin{minipage}{0.36\textwidth}
    \centering
    \includegraphics[width=\textwidth]{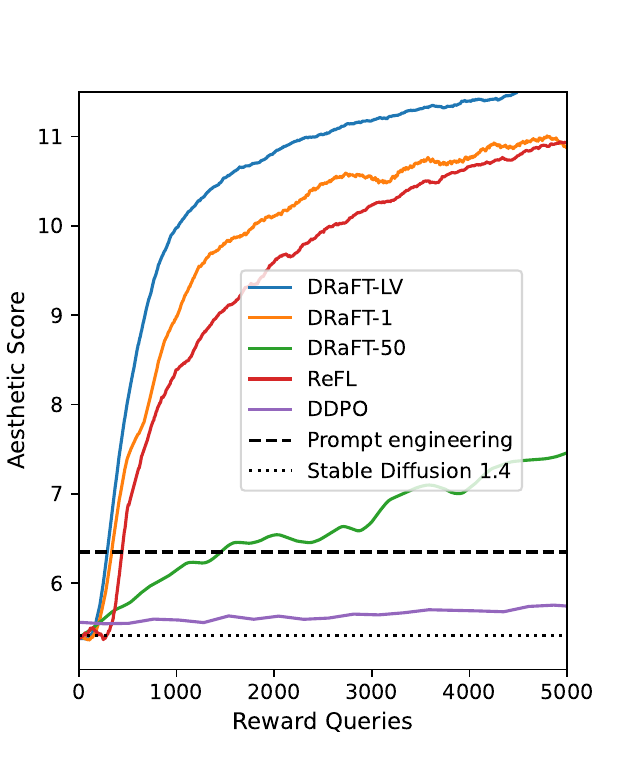}
    \end{minipage}
    \quad
    \begin{minipage}{0.46\textwidth}
    \centering
    \includegraphics[width=\textwidth]{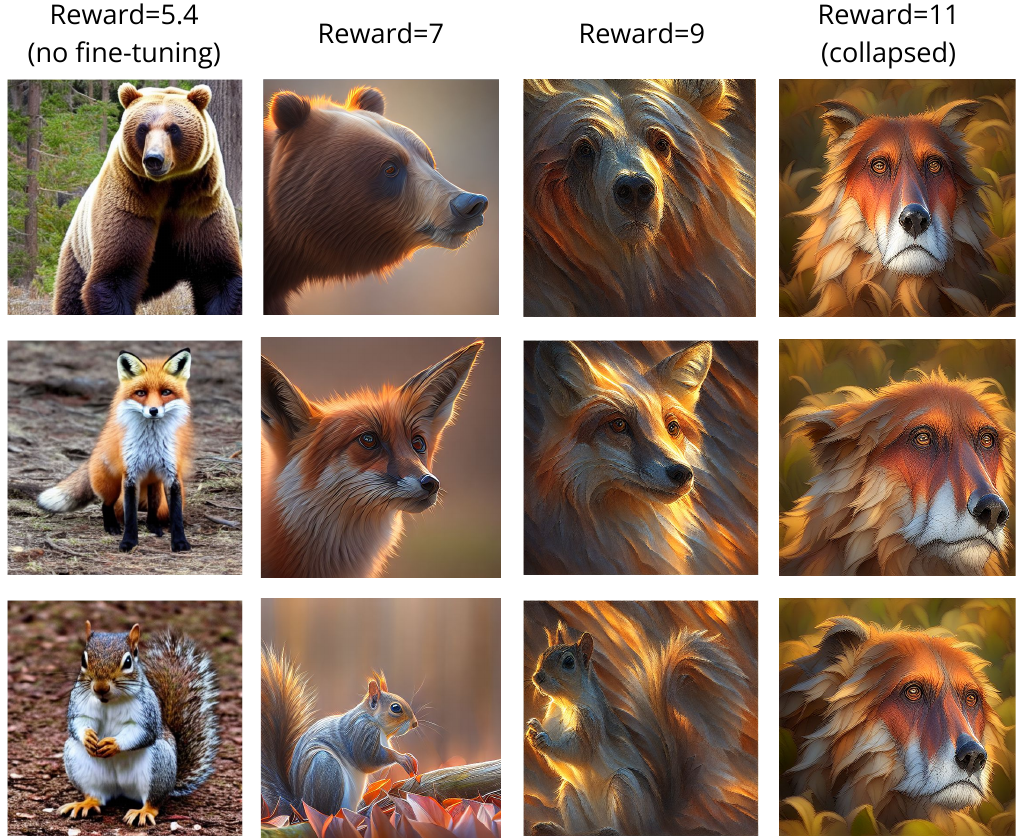}
    \label{fig:image2}
  \end{minipage}
  \vspace{-0.4cm}
  \caption{\small
  \underline{Left}: Sample efficiency of methods when maximizing scores from the LAION Aesthetic Classifier. DDPO \citep{black2023training} results were obtained from correspondence with the authors, as the results in the original paper use a buggy implementation; DDPO achieves a reward of 7.4 after 50k reward queries. \underline{Right}: Qualitative comparison of generations as training progresses.}
  \label{fig:aesthetic}
\end{figure}

First, we compared fine-tuning methods to improve aesthetic quality scores given by the LAION aesthetic predictor, which is trained to rate images on a scale of 1 through 10.
We used the simple set of 45 prompts from \citet{black2023training}, where each prompt is the name of a common animal.
We compared against the DDPO \citep{black2023training} RL method, ReFL \citep{xu2023imagereward}, and a prompt engineering baseline (see Appendix~\ref{app:baselines} for more details).
Results are shown in Figure~\ref{fig:aesthetic}.
Because it does not make use of gradients, reinforcement learning is much less sample-efficient than DRaFT. 
Interestingly, truncating DRaFT to a single backwards step substantially improves sample efficiency; we analyze this phenomenon further in Section~\ref{sec:understanding-k}.
Thanks to its lower-variance gradient estimate, DRaFT-LV further improves training efficiency.
We observed that the methods initially produced improved images, but eventually collapsed to producing very similar high-reward images.

\vspace{-0.2cm}
\paragraph{Reward Hacking.}
In Figure~\ref{fig:aesthetic} (Right), we observe an instance of \textit{reward hacking}, where the fine-tuned model loses diversity and collapses to generate a certain high-reward image.
Reward hacking points to deficiencies in existing rewards; efficiently finding gaps between the desired behavior and the behavior implicitly captured by the reward is a crucial step in fixing the reward.
Improved techniques for fine-tuning, such as DRaFT, can aid the development of new rewards.
We discuss reward hacking further in Appendix~\ref{app:over-optimization}.

\vspace{-0.2cm}
\subsection{Fine-Tuning on Human Preferences}
\label{sec:preferences}

\begin{figure}
    \vspace{-0.1cm}
    \centering
    \includegraphics[width=0.9\linewidth]{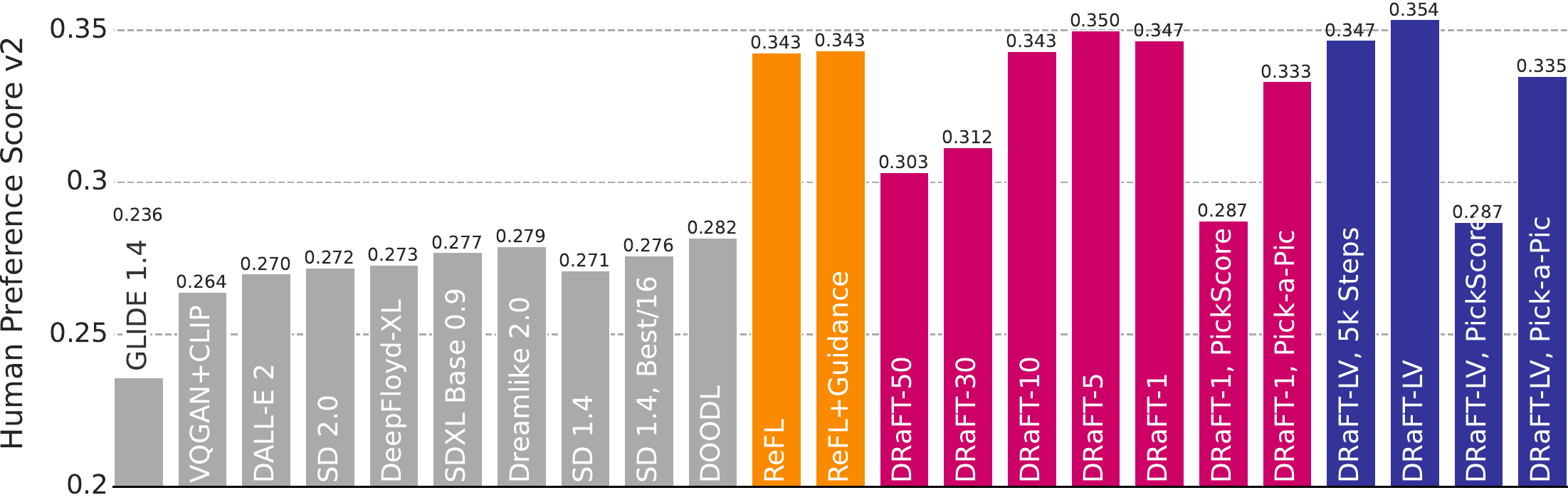}
    \vspace{-0.1cm}
    \caption{\small Comparison of the HPSv2 rewards achieved by a variety of baselines ({\color{gray}denoted by gray bars}), {\color{myorange}ReFL}, {\color{mypurple}DRaFT-$K$}, and {\color{myblue}DRaFT-LV} on the HPDv2 test set.
    All baselines except for DOODL and best of 16 are taken from~\citet{wu2023human}. Unless indicated otherwise, models are Stable Diffusion 1.4 fine-tuned for 10k steps.
    }
    \label{fig:bar-chart}
    \vspace{-0.3cm}
\end{figure}

Next, we applied DRaFT to two reward functions trained on human preference data: Human Preference Score v2 (HPSv2; \citealt{wu2023human}) and PickScore~\citep{laionopenclip2022}.

\begin{figure}[h]
    \centering
    \includegraphics[width=0.85\linewidth]{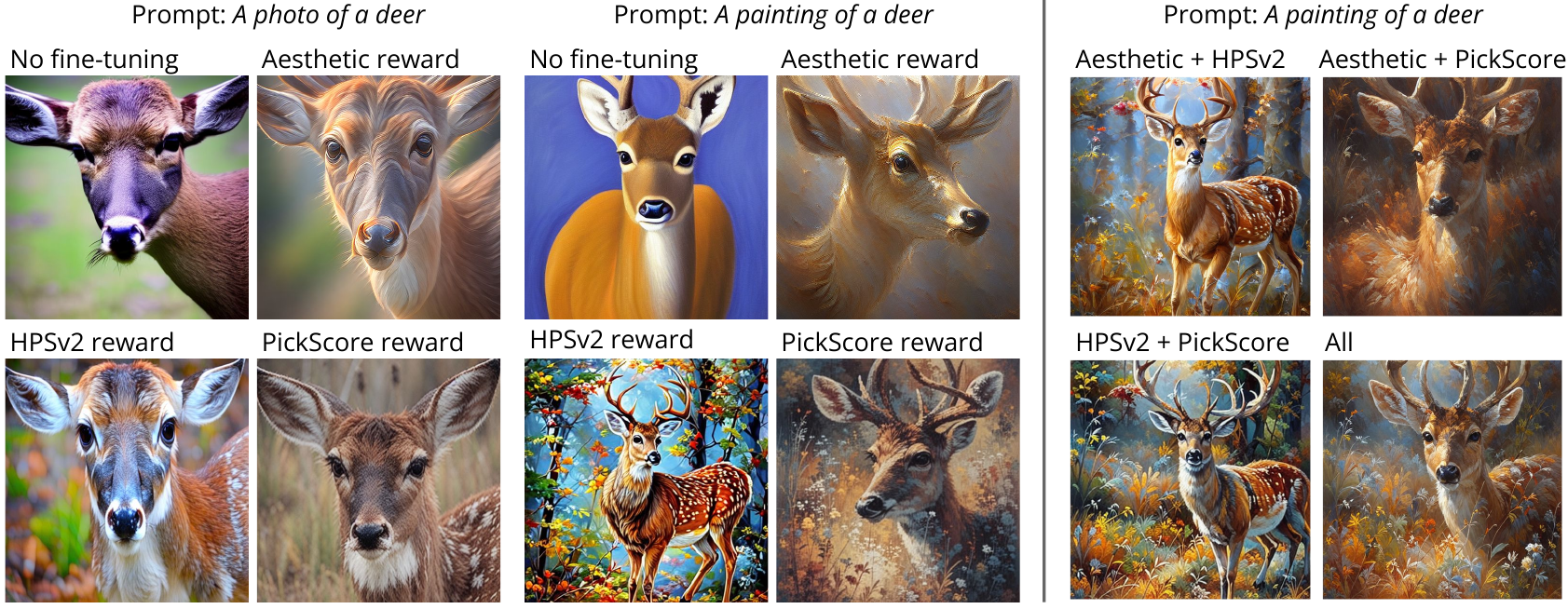}
    \vspace{-0.2cm}
    \caption{\small \underline{Left:} Comparison of generations  (using the same random seed) from DRaFT models fine-tuned using the LAION Aesthetic Classifier, Human Preference Score v2, and PickScore reward functions. \underline{Right:} Outputs of DRaFT models combined by mixing the LoRA weights for different reward functions.}
    \label{fig:reward-comparison}
    \vspace{-0.2cm}
\end{figure}

\vspace{-0.2cm}
\paragraph{Qualitative comparison of reward functions.}
First, we trained DRaFT models using the aesthetic, PickScore, and HPSv2 rewards on the same set of 45 animal prompts.
As the reward models behave differently for photographs and paintings, we trained two versions of each model. 
Qualitatively, we found that the different reward functions led to quite different generated images (see Figure~\ref{fig:reward-comparison}, Left).
HPSv2 generally encourages more colorful but less photorealistic generations.

\begin{figure}[h]
    \centering
    \includegraphics[width=0.95\linewidth]{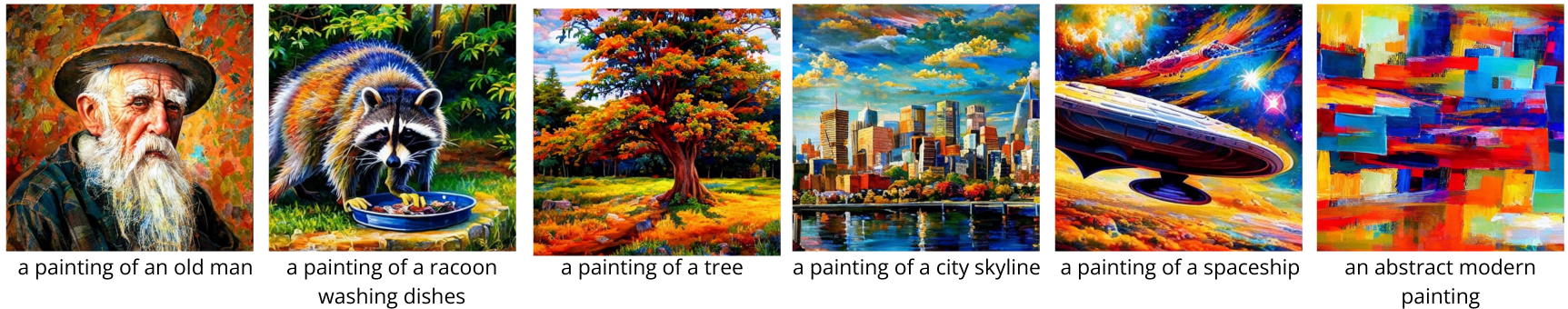}
    \vspace{-0.2cm}
    \caption{\small A DRaFT model trained on a simple prompt set of 45 animals generalizes well to other image subjects while preserving the learned style (Human Preference Score v2 reward).}
    \label{fig:hpsv2-generalization}
    \vspace{-0.2cm}
\end{figure}

\vspace{-0.2cm}
\paragraph{Human Preference Score Comparison.}
Next, we compared DRaFT-finetuned models to several baselines on the Human Preference Score v2 benchmark~\citep{wu2023human}.
We used DRaFT to fine-tune for HPSv2, using the Human Preference Dataset v2 (HPDv2) training set prompts.
In Figure~\ref{fig:bar-chart}, we report the mean performance over the four test set categories; Table~\ref{table:quantitative} in Appendix~\ref{app:quantitative-table} provides a detailed breakdown of results by category.
We compared DRaFT to a selected set of baselines from~\citet{wu2023human}, as well as to DOODL~\citep{wallace2023end} and ReFL~\citep{xu2023imagereward}.
As ReFL does not use guidance during training, we also experimented with adding guidance to it. Interestingly, the model with guidance performs similarly; possibly fine-tuning compensates for the lack of guidance similarly to how guided diffusion models can be distilled into unguided ones \citep{meng2023distillation}.
DRaFT-1 slightly outperforms ReFL, while being simpler to implement. DRaFT-LV achieves the best reward value, learning approximately $2 \times$ faster than DRaFT-1 (see the ``DRaFT-LV, 5k steps'' bar).
Figure~\ref{fig:bar-chart} also provides an ablation comparing DRaFT-$K$ for $K \in \{ 1, 5, 10, 30, 50 \}$.
We found that using smaller $K$ was beneficial, both in terms of training time and final performance.

\vspace{-0.2cm}
\paragraph{Evaluating Generalization.}
Next, we investigated several types of generalization: 1) \textit{prompt generalization}, where we fine-tuned for HPSv2 on one prompt dataset (Pick-a-Pic) and evaluated performance on another prompt dataset (HPDv2)---these results are shown by the ``DRaFT, Pick-a-Pic'' bars in Figure~\ref{fig:bar-chart}; and 2) \textit{reward generalization}, where we fine-tuned using a different reward (PickScore) than we used for evaluation (HPSv2)---these results are shown by the ``DRaFT, PickScore'' bars.
DRaFT-1 and DRaFT-LV transfer well across prompt sets and fairly well across reward functions.
We also show qualitative prompt generalization results in Figure~\ref{fig:hpsv2-generalization}: we found that a model fine-tuned using only 45 animal prompts learns a style that generalizes well to other prompts.

\vspace{-0.2cm}
\paragraph{Large-Scale Multi-Reward Model.}
We pushed the scale of DRaFT-LV further by performing a $2\times$ longer training run on the larger HPDv2 dataset, optimizing for a weighted combination of rewards, $\text{PickScore}=10, \text{HPSv2}=2, \text{Aesthetic}=0.05$.\footnote{The gradients of the different rewards have substantially different norms, leading to the wide range of coefficients. The particular values were chosen based on sample quality after experimenting with a few settings.}
Images generated by the model are shown in Figure~\ref{fig:main-figure}.
Overall, the fine-tuned model produces more detailed and stylized images than baseline Stable Diffusion. 
In some cases, fine-tuning improves the model's image-text alignment as well as aesthetics, because the PickScore and HPSv2 rewards measure the similarity between the prompt and image embeddings. 
For example, the fine-tuned model almost always generates images faithful to the prompt ``a raccoon washing dishes,'' while baseline Stable Diffusion does so less than half the time.
However, we also found that the model sometimes overgeneralizes the reward function, generating detailed images even when the prompt calls for a simpler style (see Figure~\ref{fig:overgeneralization} in Appendix~\ref{app:additional-results}).

\begin{figure}[h]
    \centering
    \includegraphics[width=0.6\linewidth]{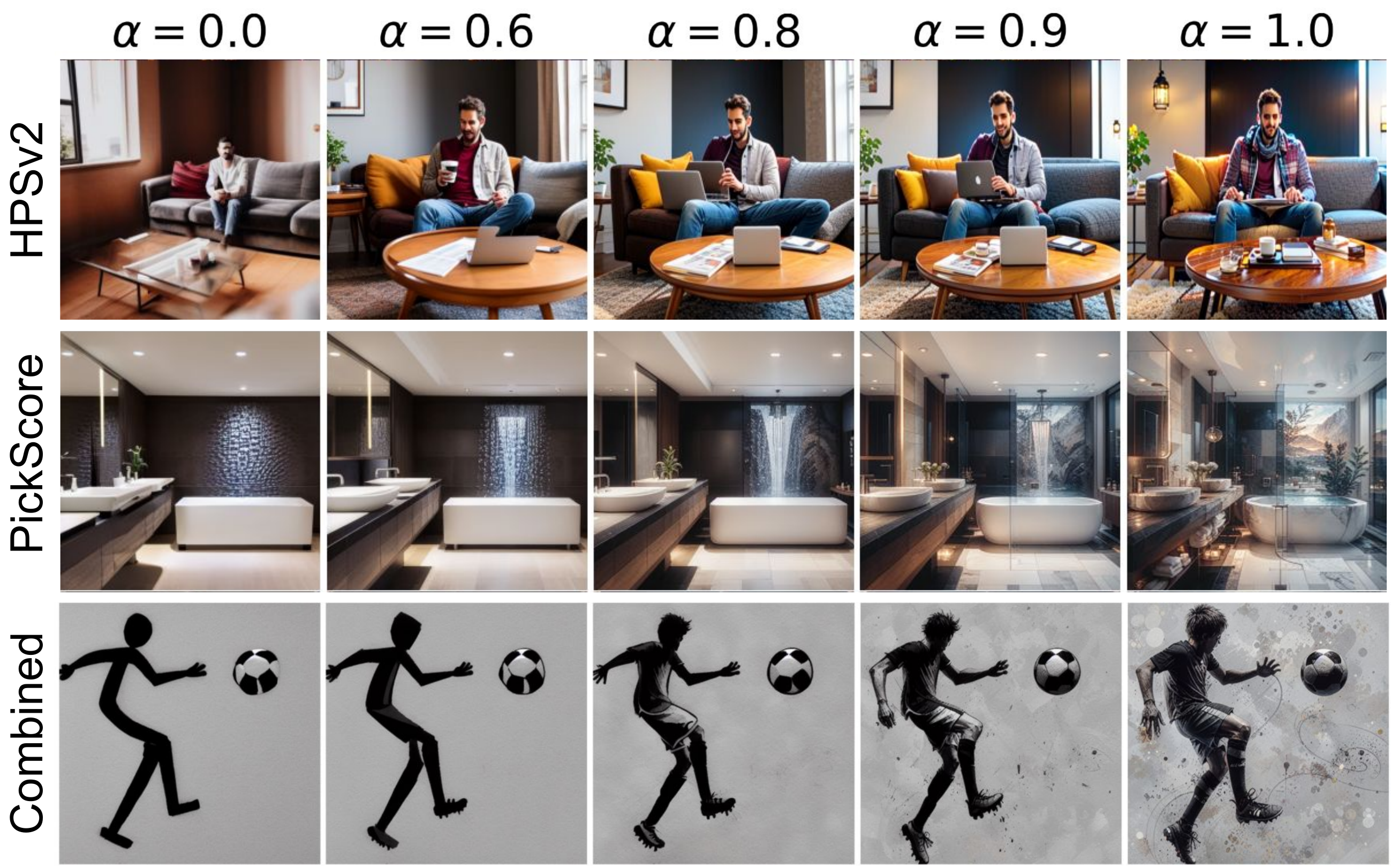}
    \hfill
    \includegraphics[width=0.38\linewidth]{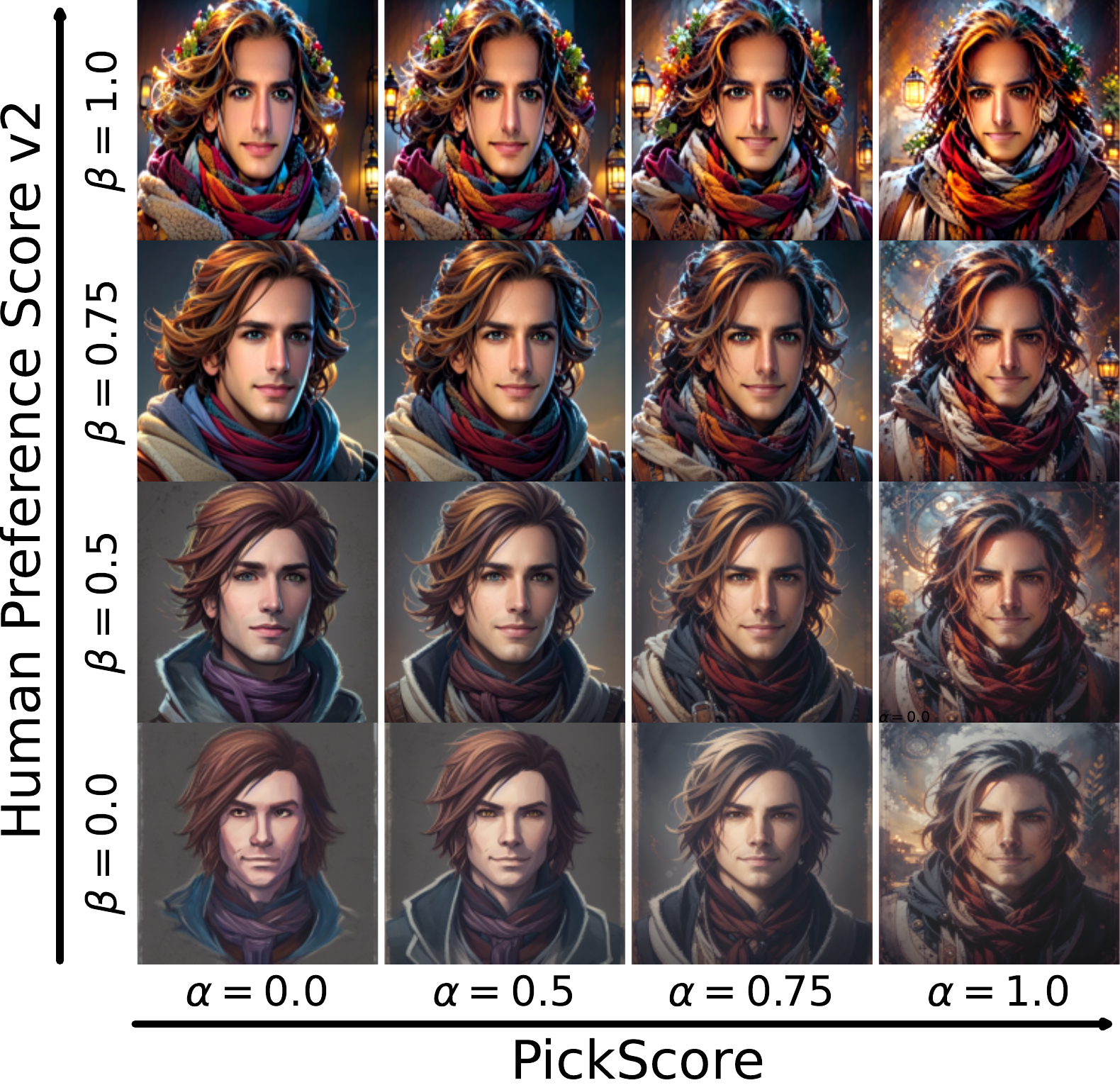}
    \vspace{-0.2cm}
    \caption{\small \textbf{Interpolating LoRA parameters.} \underline{Left:} Generations from DRaFT models with different scalar multipliers applied to the LoRA weights.
    Scaling LoRA parameters yields semantic interpolations between the pre-trained and fine-tuned models.
    \underline{Right:} Combining LoRA parameters that have been fine-tuned independently for two different rewards—PickScore and HPSv2—allows for fine-grained control over style without additional training.
    For a fixed noise sample, we show images generated using $\alpha \boldtheta^{\text{PickScore}}_\text{LoRA} + \beta \boldtheta^{\text{HPSv2}}_\text{LoRA}$.
    }
    \label{fig:lora-scaling}
    \vspace{-0.2cm}
\end{figure}

\vspace{-0.2cm}
\paragraph{Scaling and Mixing LoRA Weights.}
It is possible to control the strength of the fine-tuning simply by scaling the LoRA parameters: multiplying them by a scalar $\alpha < 1$ moves the adapted parameters closer to the original, pre-trained model.
Figure ~\ref{fig:lora-scaling} (Left) shows that this method smoothly interpolates between the pre-trained and DRaFT models.
The LoRA scale could become a useful hyperparameter for controlling generations from DRaFT models similarly to how guidance strength is currently used.
While we experimented with KL regularization and early stopping as alternative ways of reducing reward overfitting (see Figure~\ref{fig:reward-generalization} in Appendix~\ref{app:over-optimization}), we found LoRA scaling to work best. 

We also probed the compositional properties of LoRA weights adapted for different rewards.
One can combine the effects of multiple reward functions by interpolating the LoRA weights of the respective DRaFT models (Figure~\ref{fig:reward-comparison}, Right).
In Figure~\ref{fig:lora-scaling} (Right), we show images generated using linear combinations of LoRA parameters trained separately for the PickScore and HPSv2 rewards, using scaling coefficients $\alpha$ and $\beta$.
The resulting images mix the semantics of HPSv2 and PickScore, and are simultaneously more saturated and detailed than the pre-trained model outputs.
While it would also be possible to instead directly sum the score functions of the models (similar to \citealt{liu2022compositional}), this would increase generation time proportionally to the number of combined models.

\vspace{-0.2cm}
\paragraph{Understanding the Impact of $K$.}
\label{sec:understanding-k}

\begin{wrapfigure}[9]{r}{0.25\linewidth}
    \vspace{-0.4cm}
    \centering
    \includegraphics[width=\linewidth]{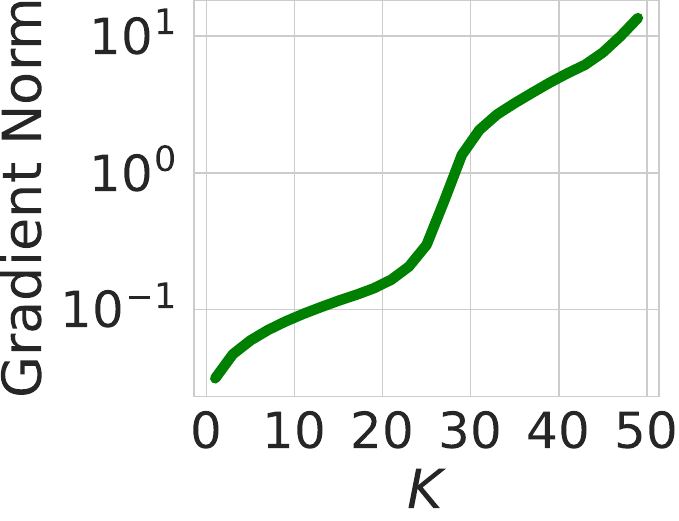}
    \vspace{-0.6cm}
    \caption{\footnotesize Gradient norms}
    \label{fig:draft-k-grad-norms}
\end{wrapfigure}

\begin{figure*}[htb]
    \centering
    \begin{minipage}[t]{.5\textwidth}
        \centering
        \includegraphics[width=\linewidth]{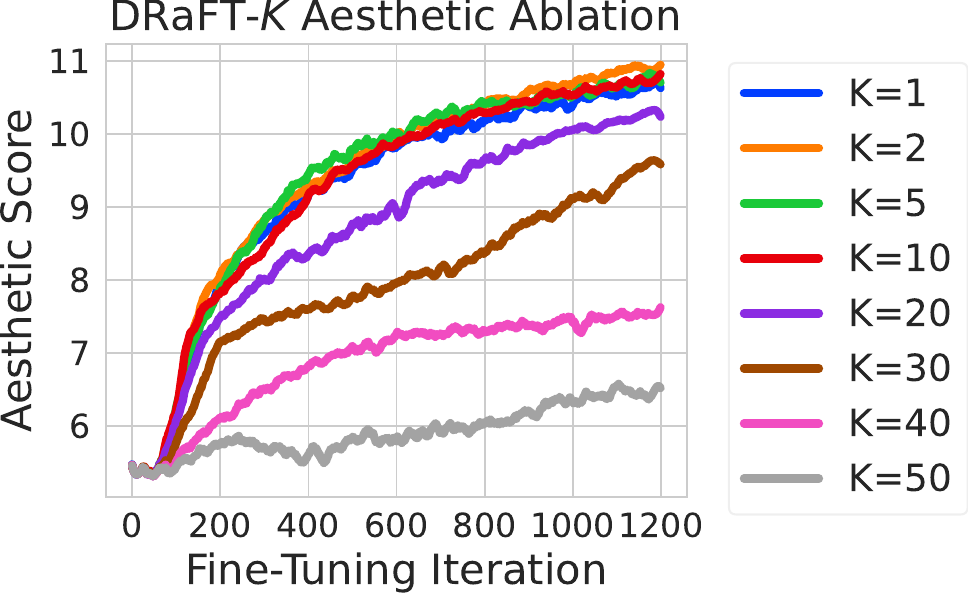}
        \vspace{-0.5cm}
        \caption{\small Ablation over $K$ for aesthetic reward fine-tuning.}
        \label{fig:aesthetic-ablation-k}
    \end{minipage}
    \qquad
    \begin{minipage}[t]{.38\textwidth}
        \centering
        \includegraphics[width=\linewidth]{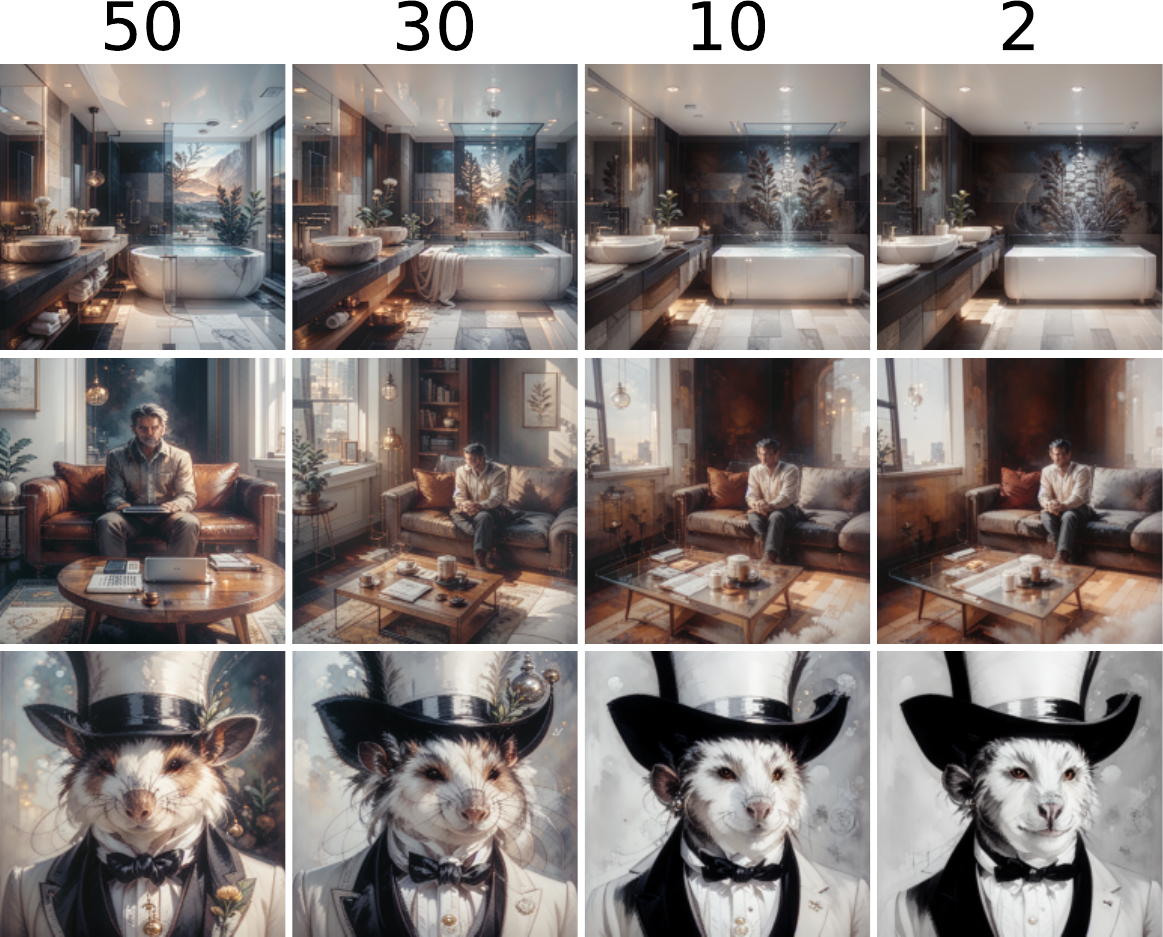}
        \vspace{-0.5cm}
        \caption{\small Images generated with different LoRA start steps $M \in \{ 50, 30, 10, 2 \}$.}
        \label{fig:lora-mouse-start}
    \end{minipage}  
    \label{fig:1-2}
\end{figure*}

Interestingly, we found that truncated backpropagation improves sample efficiency as well as compute efficiency.
Our analysis suggests that gradients explode as $K$ increases, which may lead to optimization difficulties for large $K$ (see Figure~\ref{fig:draft-k-grad-norms}).
We found that gradient clipping somewhat alleviates this issue (see Figure~\ref{fig:gradient-clipping-ablation} in Appendix~\ref{app:understanding-k}), but does not fully close the gap.
In Figure~\ref{fig:aesthetic-ablation-k}, we show an ablation over $K$ for aesthetic fine-tuning: performance degrades as $K$ increases for $K > 10$.
In Figure~\ref{fig:lora-mouse-start}, we show that although DRaFT-1 only computes gradients through the final generation step, it generalizes across time and affects the whole sampling chain.
We compare samples generated with different \textit{LoRA start iterations} $M$, where we applied the LoRA-adapted diffusion parameters for the last $M$ steps of the sampling chain, while in the first $T - M$ steps we used the pre-trained Stable Diffusion parameters without LoRA adaptation.
We observe that adaptation does not happen only at the end of sampling (also see Figure~\ref{fig:lora-start-step} in App.~\ref{app:understanding-k}).
We also investigated the opposite scenario, where the LoRA-adapted parameters are used in the first $M$ steps, after which LoRA is ``turned off'' and we use only the original pre-trained parameters for the remaining $T-M$ steps (see Figure~\ref{fig:lora-end-step} App.~\ref{app:understanding-k}), which shows that it is also important to apply the LoRA parameters early in sampling.

\vspace{-0.4cm}
\subsection{Other Reward Functions}
\vspace{-0.1cm}

\vspace{-0.1cm}
\paragraph{Compressibility and Incompressibility.}
\label{sec:compressibility}
\begin{wrapfigure}[14]{r}{0.55\linewidth}
    \vspace{-0.8cm}
    \centering
    \includegraphics[width=\linewidth,trim={0 0 4.6cm 0.7cm},clip]{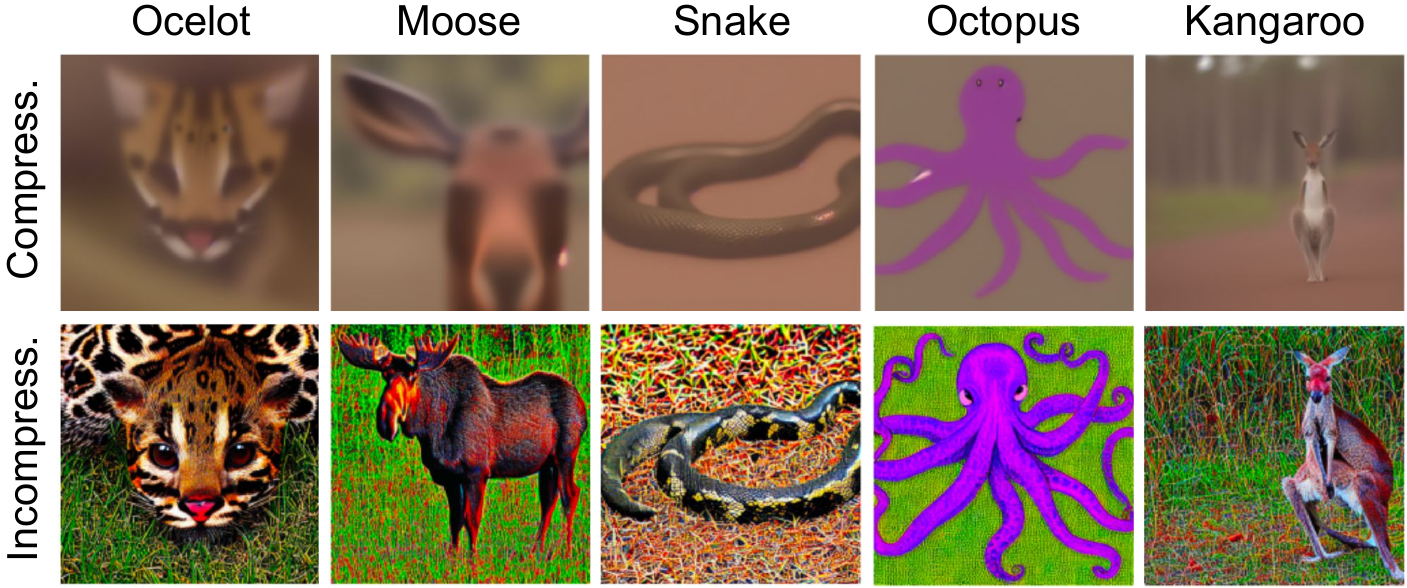}
    \vspace{-0.6cm}
    \caption{
    \small \underline{Top:}
    Images generated by a diffusion model fine-tuned for JPEG compressibility, with reward $r(\boldx_0) = -\| \boldx_0 - d(c(\boldx_0)) \|_2^2$ where $\boldx_0 = \text{sample}(\boldtheta, \boldc, \boldx_T)$.
    \underline{Bottom:} Using the \textit{negation} of the compressibility loss yields images that are difficult to compress.
    }
    \vspace{-0.2cm}
    \label{fig:compressibility-results}
\end{wrapfigure}
Inspired by~\cite{black2023training}, we also fine-tuned a diffusion model to generate easily-compressible images.
While the JPEG compression and decompression algorithms available in common libraries are not differentiable by default, they can be implemented in a differentiable manner using deep learning libraries like PyTorch~\citep{paszke2019pytorch} or JAX~\citep{jax2018github}.
Given a prompt $\boldc$, we pass the output of the diffusion model, $\boldx_0 = \text{sample}(\boldtheta, \boldc, \boldx_T)$ through differentiable compression and decompression algorithms to obtain $d(c(\boldx_0))$, and minimize the Euclidean distance between the original and reconstructed images.
This encourages the diffusion model to produce simple images.
The results using DRaFT to fine-tune for JPEG compressibility are shown in Figure~\ref{fig:compressibility-results} (Top): these images have simple, uniform-color backgrounds and low-frequency foregrounds.
In addition, negating the compressibility loss yields \textit{incompressible} images, shown in Figure~\ref{fig:compressibility-results} (Bottom), that have complex, high-frequency foregrounds and backgrounds. Additional details are provided in Appendix~\ref{app:compress-details}.

\vspace{-0.1cm}
\paragraph{Object Detection and Removal.}
\label{sec:object-detection}
\vspace{-0.2cm}

\begin{figure}[h]
    \centering
    \includegraphics[width=0.9\linewidth]{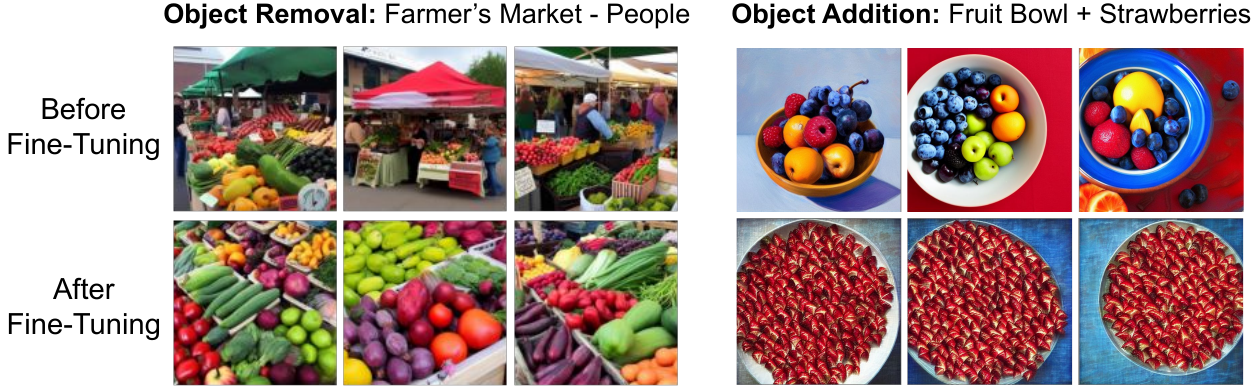}
    \vspace{-0.1cm}
    \caption{\small \textbf{Fine-tuning for object removal or addition using the OWL-ViT object detector.}
    \underline{Left:} The prompt to the diffusion model is \texttt{a farmer market} and the queries for OWL-ViT are $\mathcal{Q} = \{ \text{\texttt{a person}, \texttt{people}} \}$.
    Before fine-tuning, most of the generated images contain people in the background; after finetuning, all images contain only vegetables. \underline{Right:} Adding strawberries to a fruit bowl.}
    \label{fig:detection-results}
    \vspace{-0.1cm}
\end{figure}

Here, we explore the use of a pre-trained object detection model to bias a diffusion model's generations to include or exclude a certain object class.
We use OWL-ViT~\citep{minderer2022simple}, an open-world localization model that takes arbitrary text queries and returns bounding boxes with scores for the corresponding objects in the image.
We pass images generated by the diffusion model through a pre-trained OWL-ViT model, together with a set of queries $\mathcal{Q}$ that we wish to exclude from the generated images (see Figure~\ref{fig:object-detection} in Appendix~\ref{app:additional-object-detection} for details).
As the reward, we use the sum of scores for the localized objects corresponding to all queries $q \in \mathcal{Q}$, as well as the areas of their bounding boxes.
In Figure~\ref{fig:detection-results} (Left), we show the results of fine-tuning to minimize detection of people.
Similarly, we can maximize the sum of scores for localized objects to encourage generations to \textit{include} a certain class: in Figure~\ref{fig:detection-results} (Right), we use the prompt ``a bowl of fruit'' and queries $\mathcal{Q} = \{ \text{strawberry} \}$.

$$
$$

\vspace{-1.1cm}
\paragraph{Adversarial Examples.}
\label{sec:adversarial}
\vspace{-0.2cm}

\begin{wrapfigure}[15]{r}{0.4\linewidth}
    \vspace{-0.8cm}
    \centering
    \includegraphics[width=\linewidth,trim={0 0 5.6cm 0},clip]{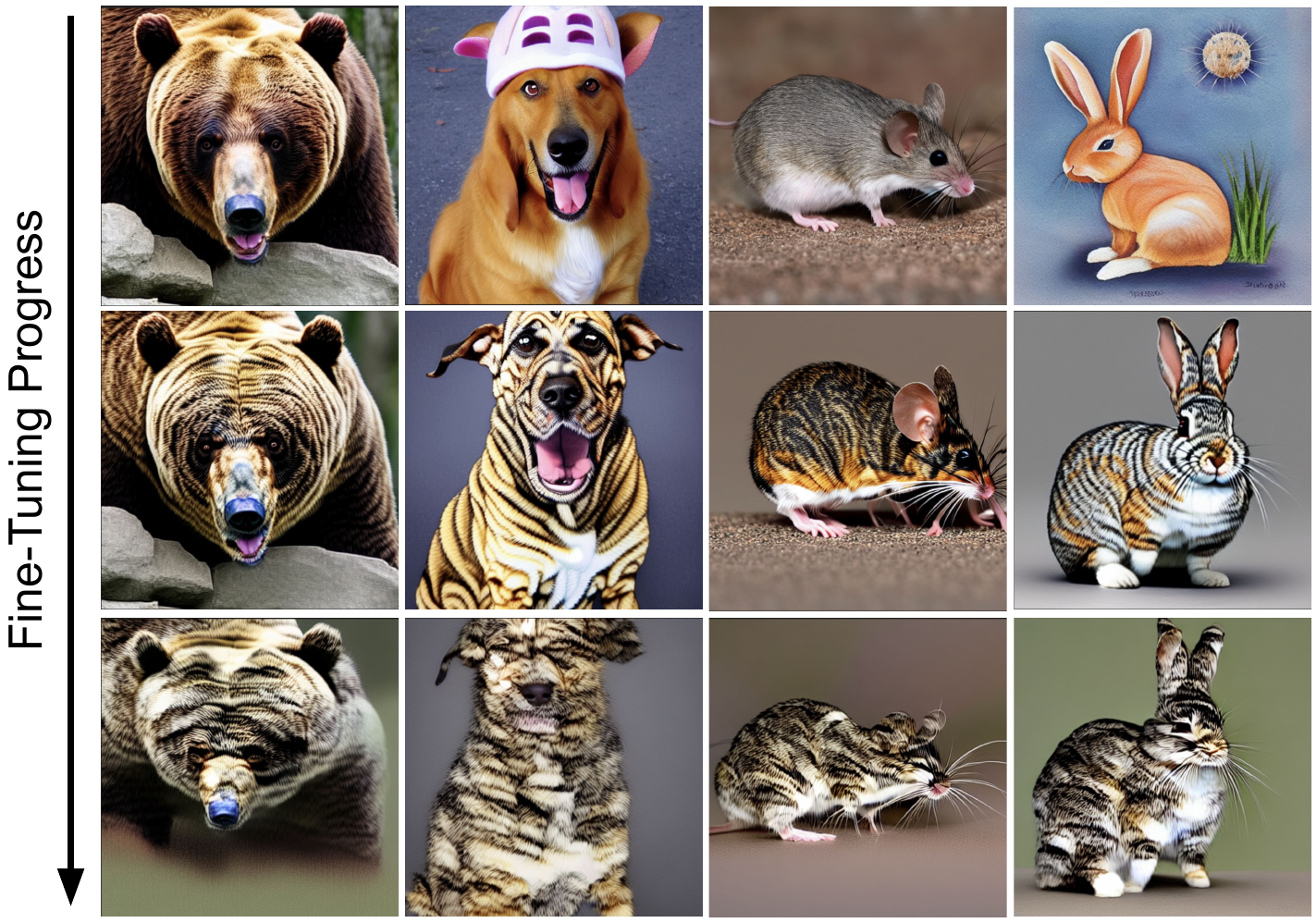}
    \vspace{-0.6cm}
    \caption{\small Diffusion adversarial examples for prompts \texttt{bear, dog, mouse} with target class \texttt{cat}. 
    }
    \label{fig:adversarial-results}
\end{wrapfigure}
Our framework also provides an avenue to study the inductive biases of both diffusion models and pre-trained classifiers: we can fine-tune a diffusion model such that images generated based on a prompt for a class $y$ (e.g., ``mouse'') are classified as a different class $y'$ (e.g., ``cat'') by a ResNet-50 pretrained on ImageNet.
As the fine-tuning reward, we use the negative cross-entropy to the target class.
From the qualitative results shown in Figure~\ref{fig:adversarial-results}, we observe that this classifier is texture-biased~\citep{geirhos2018imagenet}, as the images generated by the fine-tuned model incorporate cat-like textures while keeping the animal shapes mostly unchanged.

\vspace{-0.3cm}
\section{Conclusion}
\label{sec:conclusion}
\vspace{-0.3cm}

We introduced an efficient framework for fine-tuning diffusion models on differentiable rewards by leveraging reward gradients.
Our first method, Direct Reward Fine-Tuning (DRaFT), backpropagates through the full sampling procedure, and achieves strong performance on a variety of reward functions.
We then propose DRaFT-$K$ and DRaFT-LV, variants which improve efficiency via truncated  backpropagation through time. In addition, we draw connections between our approach and prior work such as ReFL~\citep{xu2023imagereward}, to provide a unifying perspective on the design space of gradient-based reward fine-tuning algorithms.
We show that various methods can be obtained as special cases of a more general algorithm, depending on the whether---and where---\texttt{stop\_gradient} operations are included.
We hope that our work will inspire the development of improved techniques for reward fine-tuning; just as RLHF has become crucial for deploying large language models, we believe that reward fine-tuning may become a key step for improving image generation models.

\section*{acknowledgements}
\label{sec:acknowledgements}

We thank Daniel Watson, Jason Baldridge, Hartwig Adam, Robert Geirhos, and the anonymous reviewers for their thoughtful comments and suggestions.
We thank Xiaoshi Wu for help with the HPSv2 dataset, Jiazheng Xu for answering our questions about ReFL, and Kevin Black for answering our questions about DDPO and re-running their aesthetic reward experiments.

\bibliography{references}

\begin{thebibliography}{68}
\providecommand{\natexlab}[1]{#1}
\providecommand{\url}[1]{\texttt{#1}}
\expandafter\ifx\csname urlstyle\endcsname\relax
  \providecommand{\doi}[1]{doi: #1}\else
  \providecommand{\doi}{doi: \begingroup \urlstyle{rm}\Url}\fi

\bibitem[Akrour et~al.(2011)Akrour, Schoenauer, and
  Sebag]{akrour2011preference}
Riad Akrour, Marc Schoenauer, and Michele Sebag.
\newblock Preference-based policy learning.
\newblock In \emph{Machine Learning and Knowledge Discovery in Databases:
  European Conference}, pp.\  12--27, 2011.

\bibitem[Askell et~al.(2021)Askell, Bai, Chen, Drain, Ganguli, Henighan, Jones,
  Joseph, Mann, DasSarma, et~al.]{askell2021general}
Amanda Askell, Yuntao Bai, Anna Chen, Dawn Drain, Deep Ganguli, Tom Henighan,
  Andy Jones, Nicholas Joseph, Ben Mann, Nova DasSarma, et~al.
\newblock A general language assistant as a laboratory for alignment.
\newblock \emph{arXiv preprint arXiv:2112.00861}, 2021.

\bibitem[Bai et~al.(2022{\natexlab{a}})Bai, Jones, Ndousse, Askell, Chen,
  DasSarma, Drain, Fort, Ganguli, Henighan, et~al.]{bai2022training}
Yuntao Bai, Andy Jones, Kamal Ndousse, Amanda Askell, Anna Chen, Nova DasSarma,
  Dawn Drain, Stanislav Fort, Deep Ganguli, Tom Henighan, et~al.
\newblock Training a helpful and harmless assistant with reinforcement learning
  from human feedback.
\newblock \emph{arXiv preprint arXiv:2204.05862}, 2022{\natexlab{a}}.

\bibitem[Bai et~al.(2022{\natexlab{b}})Bai, Kadavath, Kundu, Askell, Kernion,
  Jones, Chen, Goldie, Mirhoseini, McKinnon, et~al.]{bai2022constitutional}
Yuntao Bai, Saurav Kadavath, Sandipan Kundu, Amanda Askell, Jackson Kernion,
  Andy Jones, Anna Chen, Anna Goldie, Azalia Mirhoseini, Cameron McKinnon,
  et~al.
\newblock {Constitutional AI: Harmlessness from AI feedback}.
\newblock \emph{arXiv preprint arXiv:2212.08073}, 2022{\natexlab{b}}.

\bibitem[Bansal et~al.(2023)Bansal, Chu, Schwarzschild, Sengupta, Goldblum,
  Geiping, and Goldstein]{bansal2023universal}
Arpit Bansal, Hong-Min Chu, Avi Schwarzschild, Soumyadip Sengupta, Micah
  Goldblum, Jonas Geiping, and Tom Goldstein.
\newblock Universal guidance for diffusion models.
\newblock In \emph{Conference on Computer Vision and Pattern Recognition}, pp.\
   843--852, 2023.

\bibitem[Beaumont et~al.(2022)Beaumont, Wightman, Wang, Dayma, Wortsman,
  Blinkdl, Schuhmann, Jitsev, Ludwig, Cherti, and Mostaque]{laionopenclip2022}
Romain Beaumont, Ross Wightman, Phil Wang, Boris Dayma, Mitchell Wortsman,
  Blinkdl, Cristoph Schuhmann, Jenia Jitsev, Schmidt Ludwig, Mehdi Cherti, and
  Emad Mostaque.
\newblock Large scale {OpenCLIP}: {L}/14, {H}/14 and {G}/14 trained on
  {LAION}-{2B}.
\newblock \emph{laion.ai}, 2022.

\bibitem[Black et~al.(2023)Black, Janner, Du, Kostrikov, and
  Levine]{black2023training}
Kevin Black, Michael Janner, Yilun Du, Ilya Kostrikov, and Sergey Levine.
\newblock Training diffusion models with reinforcement learning.
\newblock \emph{arXiv preprint arXiv:2305.13301}, 2023.

\bibitem[Bradbury et~al.(2018)Bradbury, Frostig, Hawkins, Johnson, Leary,
  Maclaurin, Necula, Paszke, Vander{P}las, Wanderman-{M}ilne, and
  Zhang]{jax2018github}
James Bradbury, Roy Frostig, Peter Hawkins, Matthew~James Johnson, Chris Leary,
  Dougal Maclaurin, George Necula, Adam Paszke, Jake Vander{P}las, Skye
  Wanderman-{M}ilne, and Qiao Zhang.
\newblock {JAX}: composable transformations of {P}ython+{N}um{P}y programs,
  2018.
\newblock URL \url{http://github.com/google/jax}.

\bibitem[Chen et~al.(2016)Chen, Xu, Zhang, and Guestrin]{chen2016training}
Tianqi Chen, Bing Xu, Chiyuan Zhang, and Carlos Guestrin.
\newblock Training deep nets with sublinear memory cost.
\newblock \emph{arXiv preprint arXiv:1604.06174}, 2016.

\bibitem[Chen et~al.(2023)Chen, Wang, Changpinyo, Piergiovanni, Padlewski,
  Salz, Goodman, Grycner, Mustafa, Beyer, et~al.]{chen2022pali}
Xi~Chen, Xiao Wang, Soravit Changpinyo, AJ~Piergiovanni, Piotr Padlewski,
  Daniel Salz, Sebastian Goodman, Adam Grycner, Basil Mustafa, Lucas Beyer,
  et~al.
\newblock {PaLi}: A jointly-scaled multilingual language-image model.
\newblock In \emph{International Conference on Learning Representations}, 2023.

\bibitem[Chen et~al.(2015)Chen, Fang, Lin, Vedantam, Gupta, Doll{\'a}r, and
  Zitnick]{chen2015microsoft}
Xinlei Chen, Hao Fang, Tsung-Yi Lin, Ramakrishna Vedantam, Saurabh Gupta, Piotr
  Doll{\'a}r, and C~Lawrence Zitnick.
\newblock {Microsoft COCO captions: Data collection and evaluation server}.
\newblock \emph{arXiv preprint arXiv:1504.00325}, 2015.

\bibitem[Christiano et~al.(2017)Christiano, Leike, Brown, Martic, Legg, and
  Amodei]{christiano2017deep}
Paul~F Christiano, Jan Leike, Tom Brown, Miljan Martic, Shane Legg, and Dario
  Amodei.
\newblock Deep reinforcement learning from human preferences.
\newblock \emph{Advances in Neural Information Processing Systems}, 30, 2017.

\bibitem[Dhariwal \& Nichol(2021)Dhariwal and Nichol]{dhariwal2021diffusion}
Prafulla Dhariwal and Alexander Nichol.
\newblock {Diffusion models beat GANs on image synthesis}.
\newblock \emph{Advances in Neural Information Processing Systems},
  34:\penalty0 8780--8794, 2021.

\bibitem[Dong et~al.(2023)Dong, Xiong, Goyal, Pan, Diao, Zhang, Shum, and
  Zhang]{dong2023raft}
Hanze Dong, Wei Xiong, Deepanshu Goyal, Rui Pan, Shizhe Diao, Jipeng Zhang,
  Kashun Shum, and Tong Zhang.
\newblock {RAFT: Reward ranked finetuning for generative foundation model
  alignment}.
\newblock \emph{arXiv preprint arXiv:2304.06767}, 2023.

\bibitem[Fan \& Lee(2023)Fan and Lee]{fan2023optimizing}
Ying Fan and Kangwook Lee.
\newblock {Optimizing DDPM Sampling with Shortcut Fine-Tuning}.
\newblock In \emph{International Conference on Machine Learning (ICML)}, 2023.

\bibitem[Fan et~al.(2023)Fan, Watkins, Du, Liu, Ryu, Boutilier, Abbeel,
  Ghavamzadeh, Lee, and Lee]{fan2023dpok}
Ying Fan, Olivia Watkins, Yuqing Du, Hao Liu, Moonkyung Ryu, Craig Boutilier,
  Pieter Abbeel, Mohammad Ghavamzadeh, Kangwook Lee, and Kimin Lee.
\newblock {DPOK: Reinforcement learning for fine-tuning text-to-image diffusion
  models}.
\newblock In \emph{Advances in Neural Information Processing Systems}, 2023.

\bibitem[Geirhos et~al.(2019)Geirhos, Rubisch, Michaelis, Bethge, Wichmann, and
  Brendel]{geirhos2018imagenet}
Robert Geirhos, Patricia Rubisch, Claudio Michaelis, Matthias Bethge, Felix~A
  Wichmann, and Wieland Brendel.
\newblock {ImageNet-trained CNNs are biased towards texture; increasing shape
  bias improves accuracy and robustness}.
\newblock In \emph{International Conference on Learning Representations}, 2019.

\bibitem[Glaese et~al.(2022)Glaese, McAleese, Tr{\k{e}}bacz, Aslanides, Firoiu,
  Ewalds, Rauh, Weidinger, Chadwick, Thacker, et~al.]{glaese2022improving}
Amelia Glaese, Nat McAleese, Maja Tr{\k{e}}bacz, John Aslanides, Vlad Firoiu,
  Timo Ewalds, Maribeth Rauh, Laura Weidinger, Martin Chadwick, Phoebe Thacker,
  et~al.
\newblock Improving alignment of dialogue agents via targeted human judgements.
\newblock \emph{arXiv preprint arXiv:2209.14375}, 2022.

\bibitem[Gruslys et~al.(2016)Gruslys, Munos, Danihelka, Lanctot, and
  Graves]{gruslys2016memory}
Audrunas Gruslys, R{\'e}mi Munos, Ivo Danihelka, Marc Lanctot, and Alex Graves.
\newblock Memory-efficient backpropagation through time.
\newblock \emph{Advances in Neural Information Processing Systems}, 29, 2016.

\bibitem[Gu et~al.(2023)Gu, Trevithick, Lin, Susskind, Theobalt, Liu, and
  Ramamoorthi]{gu2023nerfdiff}
Jiatao Gu, Alex Trevithick, Kai-En Lin, Joshua~M Susskind, Christian Theobalt,
  Lingjie Liu, and Ravi Ramamoorthi.
\newblock {NerfDiff: Single-image view synthesis with NeRF-guided distillation
  from 3d-aware diffusion}.
\newblock In \emph{International Conference on Machine Learning}, pp.\
  11808--11826, 2023.

\bibitem[Hao et~al.(2022)Hao, Chi, Dong, and Wei]{hao2022optimizing}
Yaru Hao, Zewen Chi, Li~Dong, and Furu Wei.
\newblock Optimizing prompts for text-to-image generation.
\newblock \emph{arXiv preprint arXiv:2212.09611}, 2022.

\bibitem[Ho \& Salimans(2021)Ho and Salimans]{ho2022classifier}
Jonathan Ho and Tim Salimans.
\newblock Classifier-free diffusion guidance.
\newblock \emph{NeurIPS Workshop on Deep Generative Models and Downstream
  Applications}, 2021.

\bibitem[Ho et~al.(2020)Ho, Jain, and Abbeel]{ho2020denoising}
Jonathan Ho, Ajay Jain, and Pieter Abbeel.
\newblock Denoising diffusion probabilistic models.
\newblock \emph{Advances in Neural Information Processing Systems},
  33:\penalty0 6840--6851, 2020.

\bibitem[Ho et~al.(2022)Ho, Chan, Saharia, Whang, Gao, Gritsenko, Kingma,
  Poole, Norouzi, Fleet, et~al.]{ho2022imagen}
Jonathan Ho, William Chan, Chitwan Saharia, Jay Whang, Ruiqi Gao, Alexey
  Gritsenko, Diederik~P Kingma, Ben Poole, Mohammad Norouzi, David~J Fleet,
  et~al.
\newblock Imagen video: High definition video generation with diffusion models.
\newblock \emph{arXiv preprint arXiv:2210.02303}, 2022.

\bibitem[Hu et~al.(2022)Hu, Shen, Wallis, Allen-Zhu, Li, Wang, Wang, and
  Chen]{hu2021lora}
Edward~J Hu, Yelong Shen, Phillip Wallis, Zeyuan Allen-Zhu, Yuanzhi Li, Shean
  Wang, Lu~Wang, and Weizhu Chen.
\newblock {LoRA: Low-rank adaptation of large language models}.
\newblock In \emph{International Conference on Learning Representations}, 2022.

\bibitem[Ibarz et~al.(2018)Ibarz, Leike, Pohlen, Irving, Legg, and
  Amodei]{ibarz2018reward}
Borja Ibarz, Jan Leike, Tobias Pohlen, Geoffrey Irving, Shane Legg, and Dario
  Amodei.
\newblock {Reward learning from human preferences and demonstrations in Atari}.
\newblock \emph{Advances in Neural Information Processing Systems}, 31, 2018.

\bibitem[Kingma et~al.(2021)Kingma, Salimans, Poole, and
  Ho]{kingma2021variational}
Diederik Kingma, Tim Salimans, Ben Poole, and Jonathan Ho.
\newblock Variational diffusion models.
\newblock \emph{Advances in Neural Information Processing Systems},
  34:\penalty0 21696--21707, 2021.

\bibitem[Kirstain et~al.(2023)Kirstain, Polyak, Singer, Matiana, Penna, and
  Levy]{kirstain2023pick}
Yuval Kirstain, Adam Polyak, Uriel Singer, Shahbuland Matiana, Joe Penna, and
  Omer Levy.
\newblock Pick-a-pic: An open dataset of user preferences for text-to-image
  generation.
\newblock \emph{arXiv preprint arXiv:2305.01569}, 2023.

\bibitem[Knox \& Stone(2009)Knox and Stone]{knox2009interactively}
W~Bradley Knox and Peter Stone.
\newblock Interactively shaping agents via human reinforcement: The tamer
  framework.
\newblock In \emph{Proceedings of the fifth international Conference on
  Knowledge capture}, pp.\  9--16, 2009.

\bibitem[Lee et~al.(2023)Lee, Liu, Ryu, Watkins, Du, Boutilier, Abbeel,
  Ghavamzadeh, and Gu]{lee2023aligning}
Kimin Lee, Hao Liu, Moonkyung Ryu, Olivia Watkins, Yuqing Du, Craig Boutilier,
  Pieter Abbeel, Mohammad Ghavamzadeh, and Shixiang~Shane Gu.
\newblock Aligning text-to-image models using human feedback.
\newblock \emph{arXiv preprint arXiv:2302.12192}, 2023.

\bibitem[Li et~al.(2022)Li, Xu, Xiao, Liu, Yang, Li, Wang, Feng, She, Lyu,
  et~al.]{li2022upainting}
Wei Li, Xue Xu, Xinyan Xiao, Jiachen Liu, Hu~Yang, Guohao Li, Zhanpeng Wang,
  Zhifan Feng, Qiaoqiao She, Yajuan Lyu, et~al.
\newblock {UPainting: Unified text-to-image diffusion generation with
  cross-modal guidance}.
\newblock \emph{arXiv preprint arXiv:2210.16031}, 2022.

\bibitem[Lillicrap et~al.(2016)Lillicrap, Hunt, Pritzel, Heess, Erez, Tassa,
  Silver, and Wierstra]{lillicrap2015continuous}
Timothy~P Lillicrap, Jonathan~J Hunt, Alexander Pritzel, Nicolas Heess, Tom
  Erez, Yuval Tassa, David Silver, and Daan Wierstra.
\newblock Continuous control with deep reinforcement learning.
\newblock In \emph{International Conference on Learning Representations}, 2016.

\bibitem[Liu et~al.(2023)Liu, Chen, Yuan, Mei, Liu, Mandic, Wang, and
  Plumbley]{liu2023audioldm}
Haohe Liu, Zehua Chen, Yi~Yuan, Xinhao Mei, Xubo Liu, Danilo Mandic, Wenwu
  Wang, and Mark~D Plumbley.
\newblock {AudioLDM: Text-to-audio generation with latent diffusion models}.
\newblock In \emph{Proceedings of Machine Learning Research}, 2023.

\bibitem[Liu et~al.(2022)Liu, Li, Du, Torralba, and
  Tenenbaum]{liu2022compositional}
Nan Liu, Shuang Li, Yilun Du, Antonio Torralba, and Joshua~B Tenenbaum.
\newblock Compositional visual generation with composable diffusion models.
\newblock In \emph{European Conference on Computer Vision}, pp.\  423--439,
  2022.

\bibitem[Loshchilov \& Hutter(2019)Loshchilov and
  Hutter]{loshchilov2017decoupled}
Ilya Loshchilov and Frank Hutter.
\newblock Decoupled weight decay regularization.
\newblock In \emph{International Conference on Learning Representations}, 2019.

\bibitem[Meng et~al.(2023)Meng, Rombach, Gao, Kingma, Ermon, Ho, and
  Salimans]{meng2023distillation}
Chenlin Meng, Robin Rombach, Ruiqi Gao, Diederik Kingma, Stefano Ermon,
  Jonathan Ho, and Tim Salimans.
\newblock On distillation of guided diffusion models.
\newblock In \emph{Conference on Computer Vision and Pattern Recognition}, pp.\
   14297--14306, 2023.

\bibitem[Minderer et~al.(2022)Minderer, Gritsenko, Stone, Neumann, Weissenborn,
  Dosovitskiy, Mahendran, Arnab, Dehghani, Shen, et~al.]{minderer2022simple}
Matthias Minderer, Alexey Gritsenko, Austin Stone, Maxim Neumann, Dirk
  Weissenborn, Alexey Dosovitskiy, Aravindh Mahendran, Anurag Arnab, Mostafa
  Dehghani, Zhuoran Shen, et~al.
\newblock Simple open-vocabulary object detection.
\newblock In \emph{European Conference on Computer Vision}, pp.\  728--755,
  2022.

\bibitem[Nichol et~al.(2021)Nichol, Dhariwal, Ramesh, Shyam, Mishkin, McGrew,
  Sutskever, and Chen]{nichol2021glide}
Alex Nichol, Prafulla Dhariwal, Aditya Ramesh, Pranav Shyam, Pamela Mishkin,
  Bob McGrew, Ilya Sutskever, and Mark Chen.
\newblock Glide: Towards photorealistic image generation and editing with
  text-guided diffusion models.
\newblock \emph{arXiv preprint arXiv:2112.10741}, 2021.

\bibitem[Ouyang et~al.(2022)Ouyang, Wu, Jiang, Almeida, Wainwright, Mishkin,
  Zhang, Agarwal, Slama, Ray, et~al.]{ouyang2022training}
Long Ouyang, Jeffrey Wu, Xu~Jiang, Diogo Almeida, Carroll Wainwright, Pamela
  Mishkin, Chong Zhang, Sandhini Agarwal, Katarina Slama, Alex Ray, et~al.
\newblock Training language models to follow instructions with human feedback.
\newblock \emph{Advances in Neural Information Processing Systems},
  35:\penalty0 27730--27744, 2022.

\bibitem[Paszke et~al.(2019)Paszke, Gross, Massa, Lerer, Bradbury, Chanan,
  Killeen, Lin, Gimelshein, Antiga, et~al.]{paszke2019pytorch}
Adam Paszke, Sam Gross, Francisco Massa, Adam Lerer, James Bradbury, Gregory
  Chanan, Trevor Killeen, Zeming Lin, Natalia Gimelshein, Luca Antiga, et~al.
\newblock Pytorch: An imperative style, high-performance deep learning library.
\newblock \emph{Advances in Neural Information Processing Systems}, 32, 2019.

\bibitem[Poole et~al.(2023)Poole, Jain, Barron, and
  Mildenhall]{poole2022dreamfusion}
Ben Poole, Ajay Jain, Jonathan~T Barron, and Ben Mildenhall.
\newblock Dreamfusion: Text-to-3d using 2d diffusion.
\newblock In \emph{International Conference on Learning Representations}, 2023.

\bibitem[Radford et~al.(2021)Radford, Kim, Hallacy, Ramesh, Goh, Agarwal,
  Sastry, Askell, Mishkin, Clark, et~al.]{radford2021learning}
Alec Radford, Jong~Wook Kim, Chris Hallacy, Aditya Ramesh, Gabriel Goh,
  Sandhini Agarwal, Girish Sastry, Amanda Askell, Pamela Mishkin, Jack Clark,
  et~al.
\newblock Learning transferable visual models from natural language
  supervision.
\newblock In \emph{International Conference on Machine Learning}, pp.\
  8748--8763, 2021.

\bibitem[Ramesh et~al.(2021)Ramesh, Pavlov, Goh, Gray, Voss, Radford, Chen, and
  Sutskever]{ramesh2021zero}
Aditya Ramesh, Mikhail Pavlov, Gabriel Goh, Scott Gray, Chelsea Voss, Alec
  Radford, Mark Chen, and Ilya Sutskever.
\newblock Zero-shot text-to-image generation.
\newblock In \emph{International Conference on Machine Learning}, pp.\
  8821--8831, 2021.

\bibitem[Ramesh et~al.(2022)Ramesh, Dhariwal, Nichol, Chu, and
  Chen]{ramesh2022hierarchical}
Aditya Ramesh, Prafulla Dhariwal, Alex Nichol, Casey Chu, and Mark Chen.
\newblock {Hierarchical text-conditional image generation with CLIP latents}.
\newblock \emph{arXiv preprint arXiv:2204.06125}, 2022.

\bibitem[Rombach et~al.(2022)Rombach, Blattmann, Lorenz, Esser, and
  Ommer]{rombach2022high}
Robin Rombach, Andreas Blattmann, Dominik Lorenz, Patrick Esser, and Bj{\"o}rn
  Ommer.
\newblock High-resolution image synthesis with latent diffusion models.
\newblock In \emph{Proceedings of the IEEE/CVF Conference on Computer Vision
  and Pattern Recognition}, pp.\  10684--10695, 2022.

\bibitem[Ronneberger et~al.(2015)Ronneberger, Fischer, and
  Brox]{ronneberger2015u}
Olaf Ronneberger, Philipp Fischer, and Thomas Brox.
\newblock {U-Net: Convolutional networks for biomedical image segmentation}.
\newblock In \emph{Medical Image Computing and Computer-Assisted Intervention
  International Conference}, pp.\  234--241, 2015.

\bibitem[Saharia et~al.(2022)Saharia, Chan, Saxena, Li, Whang, Denton,
  Ghasemipour, Ayan, Mahdavi, Lopes, et~al.]{saharia2022photorealistic}
Chitwan Saharia, William Chan, Saurabh Saxena, Lala Li, Jay Whang, Emily
  Denton, Seyed Kamyar~Seyed Ghasemipour, Burcu~Karagol Ayan, S~Sara Mahdavi,
  Rapha~Gontijo Lopes, et~al.
\newblock Photorealistic text-to-image diffusion models with deep language
  understanding.
\newblock \emph{Advances in Neural Information Processing Systems}, 2022.

\bibitem[Schuhmann \& Beaumont(2022)Schuhmann and
  Beaumont]{laionaesthetics2022}
Christoph Schuhmann and Romain Beaumont.
\newblock {LAION}-aesthetics.
\newblock \emph{laion.ai}, 2022.

\bibitem[Silver et~al.(2014)Silver, Lever, Heess, Degris, Wierstra, and
  Riedmiller]{silver2014deterministic}
David Silver, Guy Lever, Nicolas Heess, Thomas Degris, Daan Wierstra, and
  Martin Riedmiller.
\newblock Deterministic policy gradient algorithms.
\newblock In \emph{International Conference on Machine Learning}, pp.\
  387--395, 2014.

\bibitem[Singer et~al.(2023)Singer, Polyak, Hayes, Yin, An, Zhang, Hu, Yang,
  Ashual, Gafni, et~al.]{singer2022make}
Uriel Singer, Adam Polyak, Thomas Hayes, Xi~Yin, Jie An, Songyang Zhang, Qiyuan
  Hu, Harry Yang, Oron Ashual, Oran Gafni, et~al.
\newblock Make-a-video: Text-to-video generation without text-video data.
\newblock In \emph{International Conference on Learning Representations}, 2023.

\bibitem[Sohl-Dickstein et~al.(2015)Sohl-Dickstein, Weiss, Maheswaranathan, and
  Ganguli]{sohl2015deep}
Jascha Sohl-Dickstein, Eric Weiss, Niru Maheswaranathan, and Surya Ganguli.
\newblock Deep unsupervised learning using nonequilibrium thermodynamics.
\newblock In \emph{International Conference on Machine Learning}, pp.\
  2256--2265, 2015.

\bibitem[Song et~al.(2021{\natexlab{a}})Song, Meng, and
  Ermon]{song2020denoising}
Jiaming Song, Chenlin Meng, and Stefano Ermon.
\newblock Denoising diffusion implicit models.
\newblock In \emph{International Conference on Learning Representations},
  2021{\natexlab{a}}.

\bibitem[Song \& Ermon(2019)Song and Ermon]{song2019generative}
Yang Song and Stefano Ermon.
\newblock Generative modeling by estimating gradients of the data distribution.
\newblock \emph{Advances in Neural Information Processing Systems}, 32, 2019.

\bibitem[Song et~al.(2021{\natexlab{b}})Song, Sohl-Dickstein, Kingma, Kumar,
  Ermon, and Poole]{song2020score}
Yang Song, Jascha Sohl-Dickstein, Diederik~P Kingma, Abhishek Kumar, Stefano
  Ermon, and Ben Poole.
\newblock Score-based generative modeling through stochastic differential
  equations.
\newblock In \emph{International Conference on Learning Representations},
  2021{\natexlab{b}}.

\bibitem[Stiennon et~al.(2020)Stiennon, Ouyang, Wu, Ziegler, Lowe, Voss,
  Radford, Amodei, and Christiano]{stiennon2020learning}
Nisan Stiennon, Long Ouyang, Jeffrey Wu, Daniel Ziegler, Ryan Lowe, Chelsea
  Voss, Alec Radford, Dario Amodei, and Paul~F Christiano.
\newblock Learning to summarize with human feedback.
\newblock \emph{Advances in Neural Information Processing Systems},
  33:\penalty0 3008--3021, 2020.

\bibitem[Wallace et~al.(2023)Wallace, Gokul, Ermon, and Naik]{wallace2023end}
Bram Wallace, Akash Gokul, Stefano Ermon, and Nikhil Naik.
\newblock End-to-end diffusion latent optimization improves classifier
  guidance.
\newblock \emph{arXiv preprint arXiv:2303.13703}, 2023.

\bibitem[Wang et~al.(2023{\natexlab{a}})Wang, Hunt, and
  Zhou]{wang2023diffusion}
Zhendong Wang, Jonathan~J Hunt, and Mingyuan Zhou.
\newblock Diffusion policies as an expressive policy class for offline
  reinforcement learning.
\newblock In \emph{International Conference on Learning Representations},
  2023{\natexlab{a}}.

\bibitem[Wang et~al.(2023{\natexlab{b}})Wang, Montoya, Munechika, Yang, Hoover,
  and Chau]{wang2022diffusiondb}
Zijie~J Wang, Evan Montoya, David Munechika, Haoyang Yang, Benjamin Hoover, and
  Duen~Horng Chau.
\newblock Diffusiondb: A large-scale prompt gallery dataset for text-to-image
  generative models.
\newblock In \emph{Association for Computational Linguistics},
  2023{\natexlab{b}}.

\bibitem[Watson et~al.(2022)Watson, Chan, Ho, and Norouzi]{watson2022learning}
Daniel Watson, William Chan, Jonathan Ho, and Mohammad Norouzi.
\newblock Learning fast samplers for diffusion models by differentiating
  through sample quality.
\newblock In \emph{International Conference on Learning Representations}, 2022.

\bibitem[Williams(1992)]{williams1992simple}
Ronald~J Williams.
\newblock Simple statistical gradient-following algorithms for connectionist
  reinforcement learning.
\newblock \emph{Machine Learning}, 8:\penalty0 229--256, 1992.

\bibitem[Wortsman et~al.(2022)Wortsman, Ilharco, Kim, Li, Kornblith, Roelofs,
  Lopes, Hajishirzi, Farhadi, Namkoong, et~al.]{wortsman2022robust}
Mitchell Wortsman, Gabriel Ilharco, Jong~Wook Kim, Mike Li, Simon Kornblith,
  Rebecca Roelofs, Raphael~Gontijo Lopes, Hannaneh Hajishirzi, Ali Farhadi,
  Hongseok Namkoong, et~al.
\newblock Robust fine-tuning of zero-shot models.
\newblock In \emph{Proceedings of the IEEE/CVF Conference on Computer Vision
  and Pattern Recognition}, pp.\  7959--7971, 2022.

\bibitem[Wu et~al.(2021)Wu, Ouyang, Ziegler, Stiennon, Lowe, Leike, and
  Christiano]{wu2021recursively}
Jeff Wu, Long Ouyang, Daniel~M Ziegler, Nisan Stiennon, Ryan Lowe, Jan Leike,
  and Paul Christiano.
\newblock Recursively summarizing books with human feedback.
\newblock \emph{arXiv preprint arXiv:2109.10862}, 2021.

\bibitem[Wu et~al.(2023{\natexlab{a}})Wu, Hao, Sun, Chen, Zhu, Zhao, and
  Li]{wu2023human}
Xiaoshi Wu, Yiming Hao, Keqiang Sun, Yixiong Chen, Feng Zhu, Rui Zhao, and
  Hongsheng Li.
\newblock Human preference score v2: A solid benchmark for evaluating human
  preferences of text-to-image synthesis.
\newblock \emph{arXiv preprint arXiv:2306.09341}, 2023{\natexlab{a}}.

\bibitem[Wu et~al.(2023{\natexlab{b}})Wu, Sun, Zhu, Zhao, and Li]{wu2023better}
Xiaoshi Wu, Keqiang Sun, Feng Zhu, Rui Zhao, and Hongsheng Li.
\newblock Better aligning text-to-image models with human preference.
\newblock \emph{arXiv preprint arXiv:2303.14420}, 2023{\natexlab{b}}.

\bibitem[Xu et~al.(2023)Xu, Liu, Wu, Tong, Li, Ding, Tang, and
  Dong]{xu2023imagereward}
Jiazheng Xu, Xiao Liu, Yuchen Wu, Yuxuan Tong, Qinkai Li, Ming Ding, Jie Tang,
  and Yuxiao Dong.
\newblock Imagereward: Learning and evaluating human preferences for
  text-to-image generation.
\newblock \emph{arXiv preprint arXiv:2304.05977}, 2023.

\bibitem[Zeng et~al.(2022)Zeng, Vahdat, Williams, Gojcic, Litany, Fidler, and
  Kreis]{zeng2022lion}
Xiaohui Zeng, Arash Vahdat, Francis Williams, Zan Gojcic, Or~Litany, Sanja
  Fidler, and Karsten Kreis.
\newblock {LION}: Latent point diffusion models for 3d shape generation.
\newblock In \emph{Advances in Neural Information Processing Systems
  (NeurIPS)}, 2022.

\bibitem[Zhang et~al.(2018)Zhang, Isola, Efros, Shechtman, and
  Wang]{zhang2018perceptual}
Richard Zhang, Phillip Isola, Alexei~A Efros, Eli Shechtman, and Oliver Wang.
\newblock The unreasonable effectiveness of deep features as a perceptual
  metric.
\newblock In \emph{Conference on Computer Vision and Pattern Recognition},
  2018.

\bibitem[Ziegler et~al.(2019)Ziegler, Stiennon, Wu, Brown, Radford, Amodei,
  Christiano, and Irving]{ziegler2019fine}
Daniel~M Ziegler, Nisan Stiennon, Jeffrey Wu, Tom~B Brown, Alec Radford, Dario
  Amodei, Paul Christiano, and Geoffrey Irving.
\newblock Fine-tuning language models from human preferences.
\newblock \emph{arXiv preprint arXiv:1909.08593}, 2019.

\end{thebibliography}
\bibliographystyle{iclr2024_conference}

\clearpage

\appendix

\section*{Appendix}
This appendix is structured as follows:
\begin{itemize}
    \item In Appendix~\ref{app:exp-details}, we provide more details of our experimental setup, including hyperparameters and baselines. 
    \item In Appendix~\ref{app:additional-results}, we provide additional results and analysis.
    \item In Appendix~\ref{app:extensions}, we discuss two other methods we explored that did not achieve as strong results as DRaFT-LV: a single-reward version of DRaFT-LV and applying Deterministic Policy Gradient.
    \item In Appendix~\ref{app:extended-related-work}, we provide an extended discussion of related work.
    \item In Appendix~\ref{app:uncurated-samples}, we provide uncurated samples from models fine-tuned for various reward functions, including HPSv2, PickScore, and a combined reward weighting HPSv2, PickScore, and Aesthetic score.
\end{itemize}

\section{Experimental Setup}
\label{app:exp-details}

\subsection{Additional Experimental Details and Hyperparameters}

Here we expand on training details and provide hyperparameters.
We used the same hyperparameters for ReFL and our DRaFT variants.

\vspace{-0.2cm}
\paragraph{Optimization.}
We use the AdamW optimizer \citep{loshchilov2017decoupled} with $\beta_1=0.9$, $\beta_2=0.999$, and a weight decay of 0.1. As we use LoRA, the weight decay regularizes fine-tuning by pushing parameters towards the pre-trained weights. 
For the longer training runs (10k steps on human preference reward functions), we use a square root learning rate decay, scaling the learning rate by $\min{(10 \cdot \text{step}^{-0.5}, 1)}$.
Weight decay and learning rate decay slightly improve results, but are not critical to the method.
During training, we convert the pre-trained Stable Diffusion parameters to bfloat16 to reduce the memory footprint, but use float32 for the LoRA parameters being trained. 

\vspace{-0.2cm}
\paragraph{LoRA.}
We apply LoRA to the feedforward layers as well as the cross-attention layers in the UNet, which we found to produce slightly better results. We apply LoRA only to the UNet parameters, not to the CLIP text encoder or VAE decoder mapping latents to images, as these did not improve results in initial experiments. 

\vspace{-0.2cm}
\paragraph{Sampling.}
We use 50 steps of DDIM sampling (where no noise is added during sampling beyond the initial sampled noise).
We found the results using ancestral/DDPM sampling to be similar.
We employ classifier-free guidance with a guidance weight of 7.5 both at train-time and test-time.

\vspace{-0.2cm}
\paragraph{Hyperparameters.} 
We apply DRaFT in two settings: large-scale (human preference reward functions using the HPSv2 or PickScore prompt sets) and small-scale (the other experiments). Hyperparameters are listed in Table~\ref{tab:hparams}.
Small-scale training runs take around 1.5 hours on 4 TPUv4s. Large-scale training runs take around 8 hours on 16 TPUv4s.

\begin{table}[H]
    \centering
    \small
    \begin{tabular}{l|c|c}
        \toprule
        \textbf{Hyperparameter} & \textbf{Small-Scale Experiments} & \textbf{Large-scale Experiments} \\
        \midrule
        Learning rate & 4e-4 & 2e-4 \\
        Batch size & 4 & 16 \\
        Train steps & 2k & 10k \\
        LoRA inner dimension & 8 & 32 \\ \midrule
        Weight decay & 0.1 & 0.1 \\
        DDIM steps & 50 & 50 \\
        Guidance weight & 7.5 & 7.5 \\
        DRaFT-LV inner loops $n$ & 2 & 2 \\
        ReFL max timestep $m$ & 20 & 20 \\
        \bottomrule
    \end{tabular}
    \vspace{-0.2cm}
    \caption{Training hyperparameters.}
    \vspace{-0.2cm}
    \label{tab:hparams}
\end{table}

\vspace{-0.2cm}
\subsection{Reward Function Details}
\label{app:reward-function-details}
\vspace{-0.2cm}

The \textbf{LAION aesthetic predictor} is trained to rate images on a scale of 1 through 10. The predictor consists of a feedforward neural network on top of a CLIP \citep{radford2021learning} image encoder pre-trained on a variety of human-labeled sources. 

\textbf{Human Preference Score v2} (HPSv2; \citealt{wu2023human}) is trained on prompts from DiffusionDB \citep{wang2022diffusiondb} and COCO Captions \citep{chen2015microsoft}. Images are generated from each using a wide range of generative models. Human annotators then rate which of two images is preferred for a prompt. The reward model is OpenCLIP-H \citep{laionopenclip2022} fine-tuned on the preference judgements. Unlike the aesthetic reward, HPSv2 takes into account the prompt as well as the image. 

\textbf{PickScore} \citep{kirstain2023pick} also uses OpenCLIP-H fine-tuned on preference data. However, the data is collected in a more organic way through a web app, where users can generate and then rate pairs of images generated from different diffusion models with different sampler hyperparameters.

\subsection{Baselines}
\label{app:baselines}

\vspace{-0.2cm}
\textbf{Denoising Diffusion Policy Optimization (DDPO; \citealt{black2023training})} essentially applies the REINFORCE policy gradient algorithm \citep{williams1992simple} to diffusion model generation.
The method trains the model to increase the likelihoods of actions (i.e. sampling steps) that produce high-reward samples. 
DDPO employs additional tricks such as importance weight clipping, reward normalization, and classifier-free guidance-aware training to improve results.
We show results from DDPO-IS, which extends DDPO to use an importance sampling estimator for learning on examples generated from a stale policy, enabling  multiple optimization updates per reward query.
Empirically, DDPO was found to outperform previous supervised approaches like reward weighted regression \citep{lee2023aligning}.
Rather than using the results in Figure 4 of the original paper, we obtained results from direct correspondence with the authors, as their original aesthetic reward implementation had a bug.\footnote{See {\tiny \url{https://github.com/kvablack/ddpo-pytorch/issues/3\#issuecomment-1634723127}}}

\vspace{-0.2cm}
\paragraph{Prompt Engineering.}
Users often improve the aesthetic quality of generated images through prompt engineering.
We perform semi-automated prompt search with access to the reward function, trying out various prompts and selecting the one with the best aesthetic score.
The best-scoring prompt we found is \texttt{\small A stunning beautiful oil painting of a \{:\}, cinematic lighting, golden hour light}. 

\vspace{-0.2cm}
\paragraph{Re-ranking (Best of 16).} This simple baseline generates 16 images for each caption with the pre-trained Stable Diffusion model and selects the one with the highest reward. 

\vspace{-0.2cm}
\paragraph{Direct Optimization of Diffusion Latents (DOODL).} 
DOODL \citep{wallace2023end} backpropagates through diffusion sampling to update the initial noise $\boldx_T$ so that the resulting generation produces a high reward. Rather than using an invertible diffusion algorithm to reduce memory costs as in \citet{wallace2023end}, we simply use gradient checkpointing as with DRaFT. We optimize $\boldx_T$ with Adam (which worked slightly better for us than momentum only) over 20 steps, which makes generation approximately $60\times$ slower than standard sampling.
Following \citet{wallace2023end}, we renormalize the latent at every step after the Adam update.

\vspace{-0.2cm}
\section{Additional Experimental Results}
\label{app:additional-results}
\vspace{-0.2cm}

\subsection{Over-generalization}
\label{app:overgeneralization}
Here we show examples of over-generalization, where fine-tuning on preferences trains the model to generate detailed images, even when the prompt asks for a simple drawing.
{\centering
\includegraphics[width=1.0\linewidth]{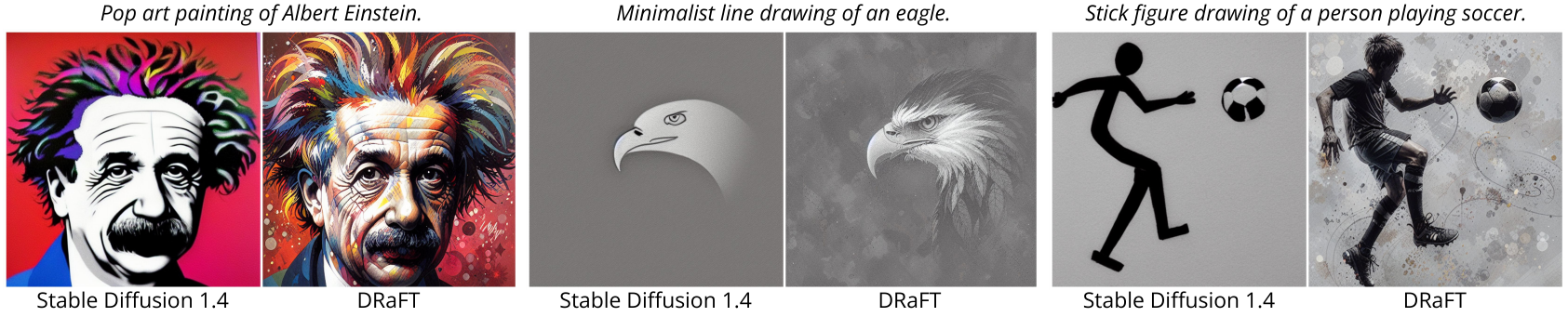}
\label{fig:overgeneralization}}

\vspace{-0.2cm}
\subsection{Full Quantitative Results}
\label{app:quantitative-table}
\vspace{-0.2cm}

In Table~\ref{table:quantitative}, we provide quantitative results comparing DRaFT to several baselines on the Human Preference Score v2 benchmark from~\citet{wu2023human}.

\begin{table}[H]
\vspace{-0.2cm}
\setlength{\tabcolsep}{5pt}
\centering
\footnotesize
\begin{tabular}{@{}lccccc@{}}
\toprule
\textbf{Model}          & \textbf{Animation}  & \textbf{Concept Art} & \textbf{Painting}                  & \textbf{Photo}      & \textbf{Averaged}   \\ \midrule
GLIDE 1.4               & $0.2334$ &
                          $0.2308$ &
                          $0.2327$ &
                          $0.2450$ &
                          $0.2355$ \\
VQGAN + CLIP            & $0.2644$ &
                          $0.2653$ &
                          $0.2647$ &
                          $0.2612$ &
                          $0.2639$ \\
DALL-E 2                & $0.2734$ &
                          $0.2654$ &
                          $0.2668$ &
                          $0.2724$ &
                          $0.2695$ \\
SD 2.0                  & $0.2748$ &
                          $0.2689$ &
                          $0.2686$ &
                          $0.2746$ &
                          $0.2717$ \\
DeepFloyd-XL            & $0.2764$ &
                          $0.2683$ &
                          $0.2686$ &
                          $0.2775$ &
                          $0.2727$ \\
SDXL Base 0.9           & $0.2842$ & $0.2763$ & $0.2760$ & $0.2729$ & $0.2773$ \\
Dreamlike Photoreal 2.0 & $0.2824$ &
                          $0.2760$ &
                          $0.2759$ &
                          $0.2799$ &
                          $0.2786$ \\
Stable Diffusion (SD 1.4)   & $0.2726$ &
                            $0.2661$ &
                            $0.2666$ &
                            $0.2727$ &
                            $0.2695$ \\
\midrule
ReFL                        & $0.3446$ & $0.3442$ & $0.3447$ & $0.3374$ & $0.3427$ \\
DRaFT-50                    & $0.3078$ & $0.3021$ & $0.3019$ & $0.3013$ & $0.3033$ \\
DRaFT-30                    & $0.3148$ & $0.3112$ & $0.3109$ & $0.3096$ & $0.3116$ \\
DRaFT-10                    & $0.3456$ & $0.3447$ & $0.3452$ & $0.3369$ & $0.3431$ \\
DRaFT-5                     & $0.3521$ & $0.3520$ & $0.3526$ & $0.3431$ & $0.3500$ \\
DRaFT-1                     & $0.3494$ & $0.3488$ & $0.3485$ & $0.3400$ & $0.3467$ \\
DRaFT-1, PickScore          & $0.2907$ & $0.2848$ & $0.2861$ & $0.2882$ & $0.2875$ \\
DRaFT-1, Pick-a-Pic         & $0.3399$ & $0.3358$ & $0.3385$ & $0.3188$ & $0.3332$ \\
DRaFT-LV                    & $0.3551$ & $0.3552$ & $0.3560$ & $0.3480$ & $0.3536$ \\
DRaFT-LV, PickScore         & $0.2880$ & $0.2833$ & $0.2846$ & $0.2923$ & $0.2870$ \\
DRaFT-LV, Pick-a-Pic        & $0.3409$ & $0.3371$ & $0.3408$ & $0.3210$ & $0.3349$ \\
\bottomrule
\end{tabular}
\vspace{-0.2cm}
\caption{\small Following~\citet{wu2023human}, we used the HPDv2 training set to fine-tune our models, and we evaluated performance on four benchmark datasets: Animation, Concept Art, Paintings, and Photos.
We report the average score achieved for each benchmark, as well as the aggregated performance across all datasets.
The results above the separator are taken from~\citet{wu2023human}; the results below were obtained from our implementations and experiments.
}
\label{table:quantitative}
\vspace{-0.4cm}
\end{table}

\vspace{-0.2cm}
\subsection{Compressibility and Incompressibility Details}
\label{app:compress-details}
\vspace{-0.2cm}

\begin{figure}[H]
    \centering
    \includegraphics[width=0.7\linewidth]{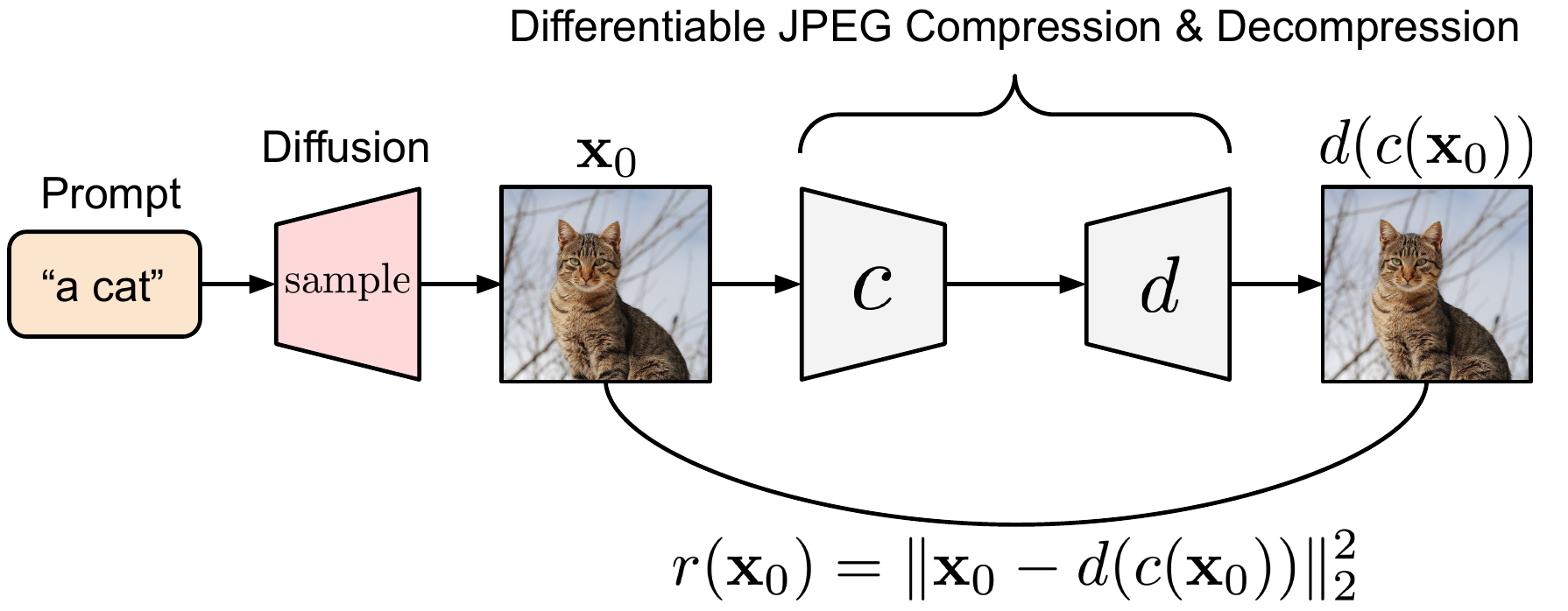}
    \vspace{-0.2cm}
    \caption{\small Fine-tuning for JPEG compressibility, using differentiable JPEG compression and decompression algorithms.}
    \label{fig:jpeg-compression}
\end{figure}

\begin{figure}[H]
    \centering
    \includegraphics[width=0.75\linewidth]{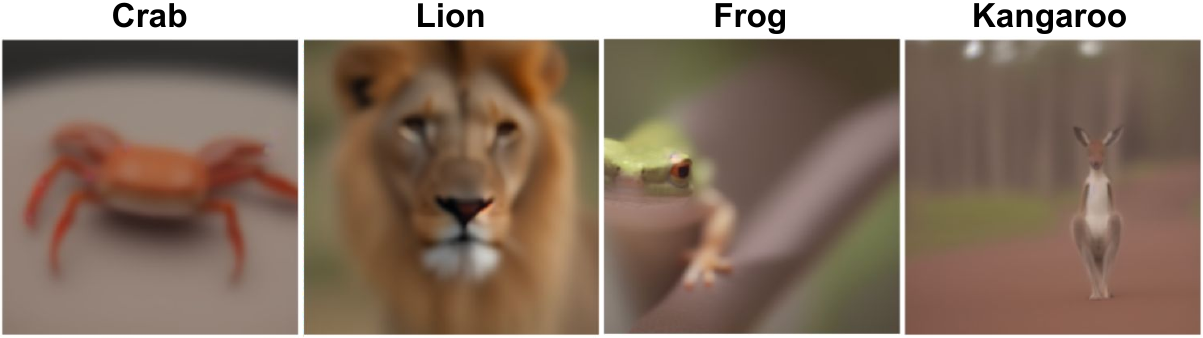}
    \vspace{-0.2cm}
    \caption{\small Images generated by a diffusion model fine-tuned for JPEG compressibility.}
    \vspace{-0.2cm}
    \label{fig:compressibility-results-appendix}
\end{figure}

\begin{figure}[H]
    \centering
    \includegraphics[width=0.75\linewidth]{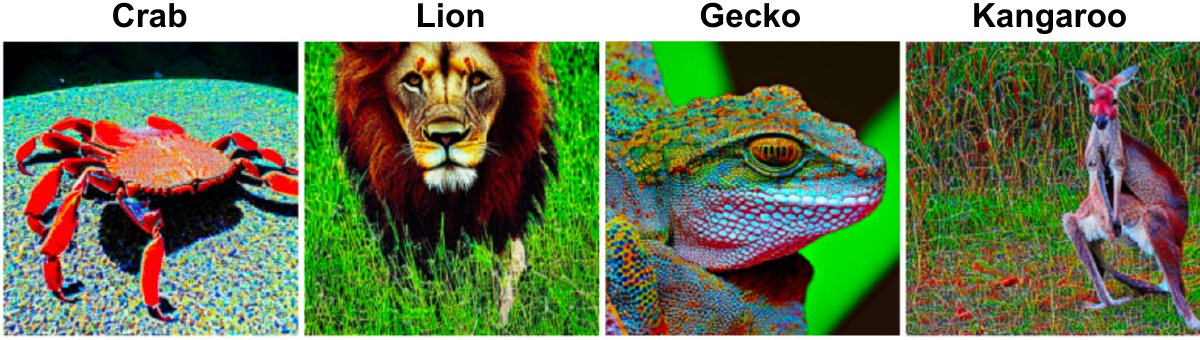}
    \vspace{-0.2cm}
    \caption{\small Images generated by a diffusion model fine-tuned using the \textit{negation} of the JPEG compressibility reward, which yields images that are \textit{difficult to compress}. We observe high-frequency foreground and background information.}
    \label{fig:incompressibility-results}
\end{figure}

\begin{figure}[H]
    \centering
    \includegraphics[width=0.75\linewidth]{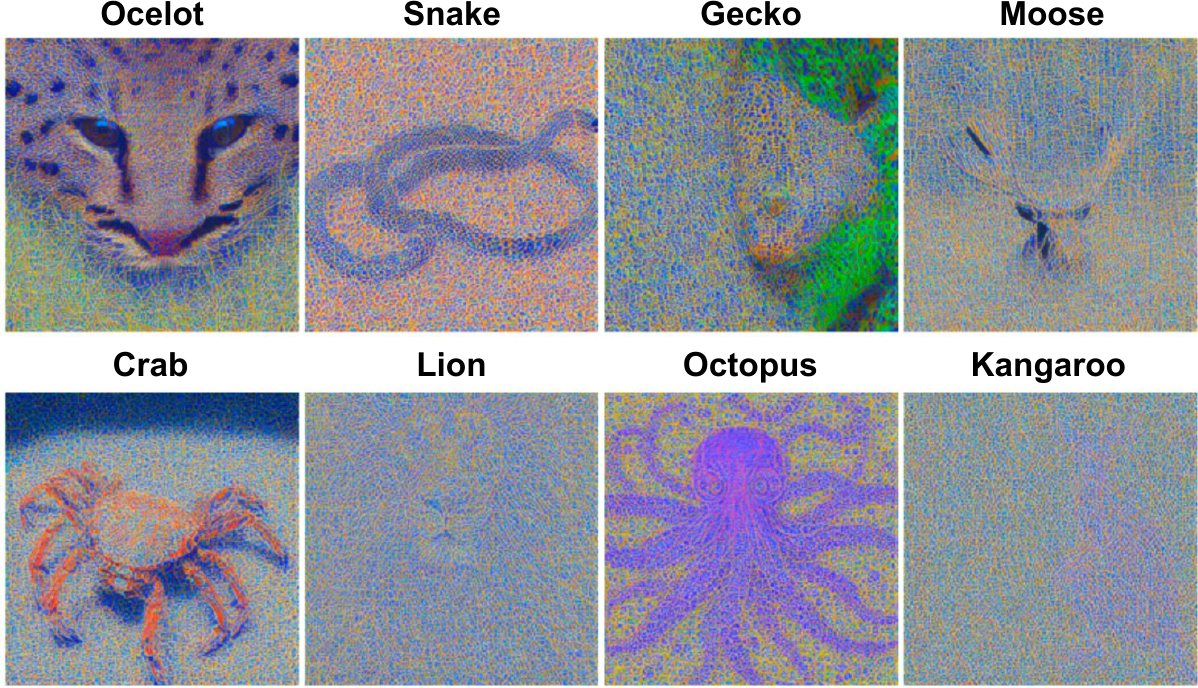}
    \vspace{-0.2cm}
    \caption{\small Over-optimization for JPEG incompressibility can lead to images with strong high-frequency texture and a loss of semantic information. Interestingly, here the octopus is generated with more than eight arms.}
    \vspace{-0.4cm}
    \label{fig:overoptim-incompressibility}
\end{figure}

Figure~\ref{fig:jpeg-compression} illustrates the setup we used to fine-tune for JPEG compressibility and incompressibility.
Given a prompt $\boldc$, we take the image generated by the diffusion model, $\boldx_0 = \text{sample}(\boldtheta, \boldc, \boldx_T)$ and pass it through JPEG compression and decompression functions, denoted by $c$ and $d$, respectively.
As the reward, we use the negative Euclidean distance between the original sample $\boldx_0$ and its reconstruction $r(\boldx_0) = - \| \boldx_0 - d(c(\boldx_0)) \|^2$.
While the JPEG implementations found in common libraries are not directly differentiable, the JPEG algorithm can be re-implemented in a differentiable way using autodiff frameworks; we re-implemented JPEG compression and decompression in JAX~\citep{jax2018github}.
Figures~\ref{fig:compressibility-results-appendix} and~\ref{fig:incompressibility-results} show additional samples generated by diffusion models fine-tuned for compressibility and incompressibility, respectively, and Figure~\ref{fig:overoptim-incompressibility} shows the results of overoptimization for incompressibility.
In our experiments, we used a fixed compression ratio of $4 \times$.

\vspace{-0.2cm}
\subsection{Adversarial Example Details}
\label{app:adv-example-details}
\vspace{-0.2cm}

\begin{figure}[H]
    \centering
    \includegraphics[width=0.8\linewidth]{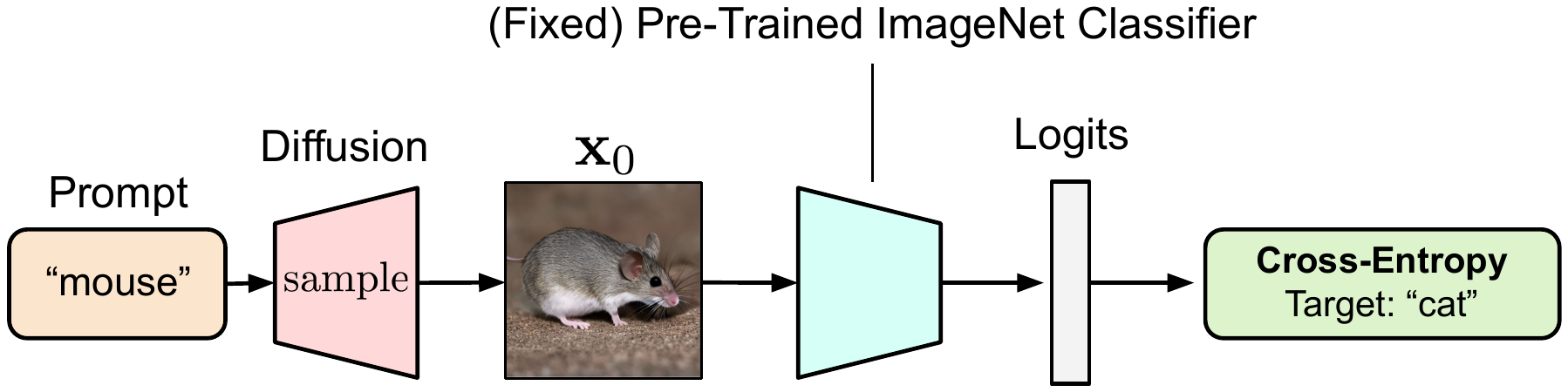}
    \caption{\small Generating ``diffusion adversarial examples'' that are conditioned on a text prompt (e.g., ``mouse'') but must be classified by a pre-trained ImageNet classifier as a different class (e.g., ``cat'').}
    \label{fig:adversarial-setup}
\end{figure}

The setup for our diffusion adversarial example experiments is illustrated in Figure~\ref{fig:adversarial-setup}.
We used a ResNet-50 model pre-trained on ImageNet-1k to classify images generated by Stable Diffusion; as the reward for fine-tuning, we used the negative cross-entropy to a fixed target class.

\vspace{-0.2cm}
\subsection{LoRA Scaling and Interpolation}
\label{app:lora-interpolation}
\vspace{-0.2cm}

\begin{figure}[H]
    \centering
    \includegraphics[width=0.9\linewidth]{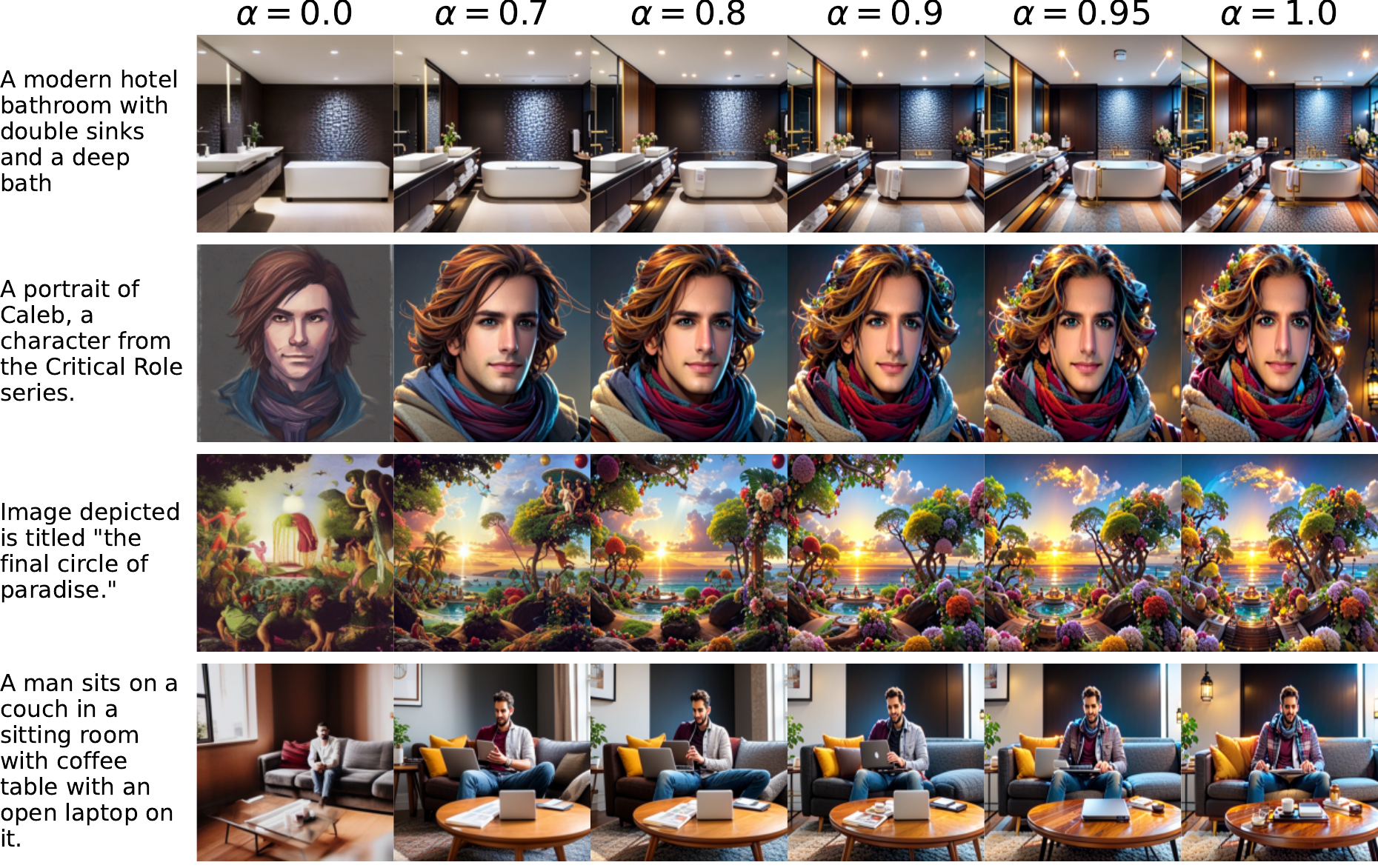}
    \vspace{-0.2cm}
    \caption{\small Scaling LoRA parameters (here adapted for Human Preference Score v2~\citep{wu2023human}) yields semantic interpolations between the original Stable Diffusion model and the fine-tuned model.}
    \label{fig:lora-interp-hpsv2}
\end{figure}

\begin{figure}[H]
    \centering
    \includegraphics[width=0.9\linewidth]{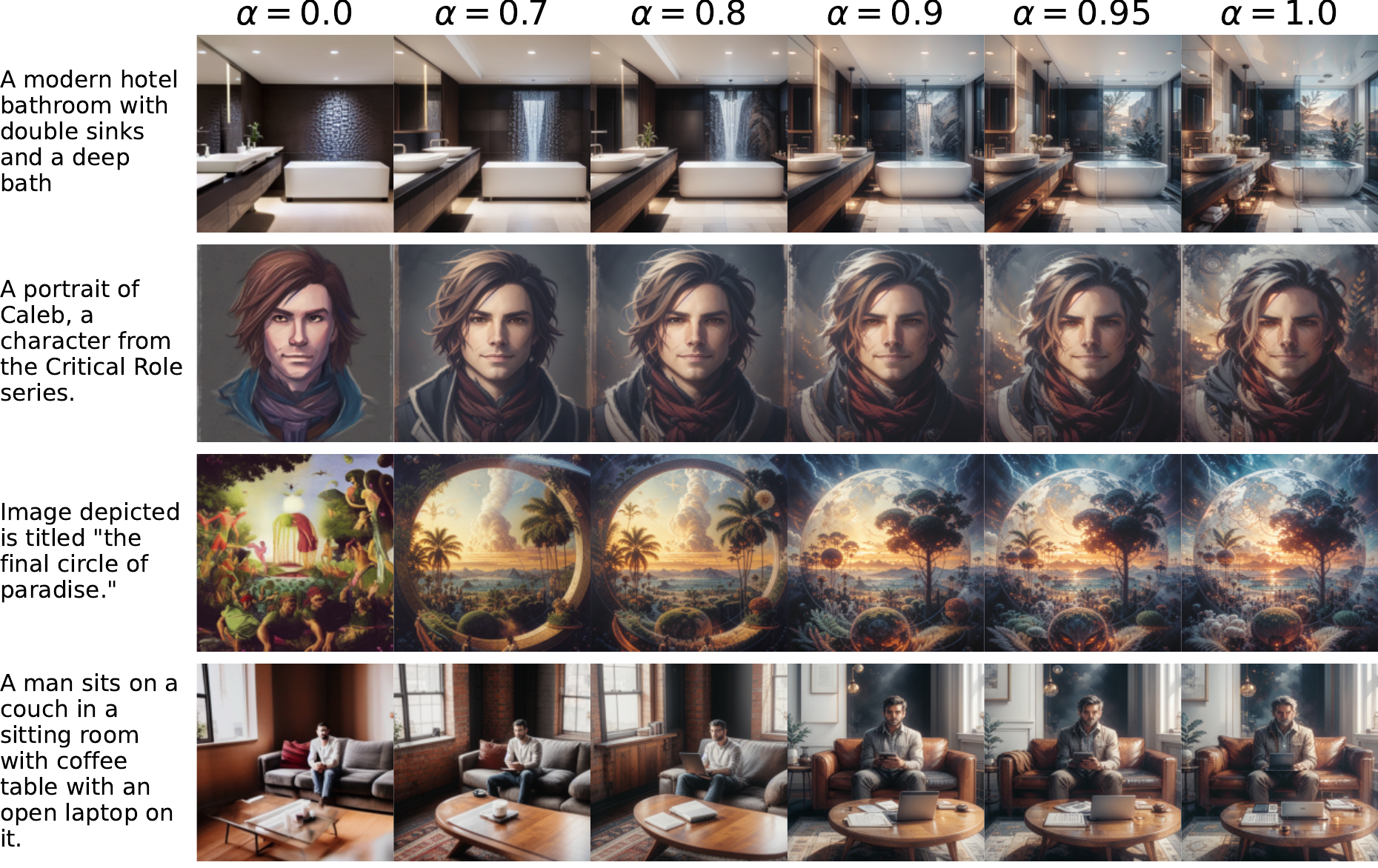}
    \vspace{-0.2cm}
    \caption{\small Scaling LoRA parameters (here adapted for PickScore~\citep{kirstain2023pick}) yields semantic interpolations between the original Stable Diffusion model and the fine-tuned model.}
    \label{fig:lora-interp-pickscore}
\end{figure}

\begin{figure}[H]
    \centering
    \includegraphics[width=0.49\linewidth]{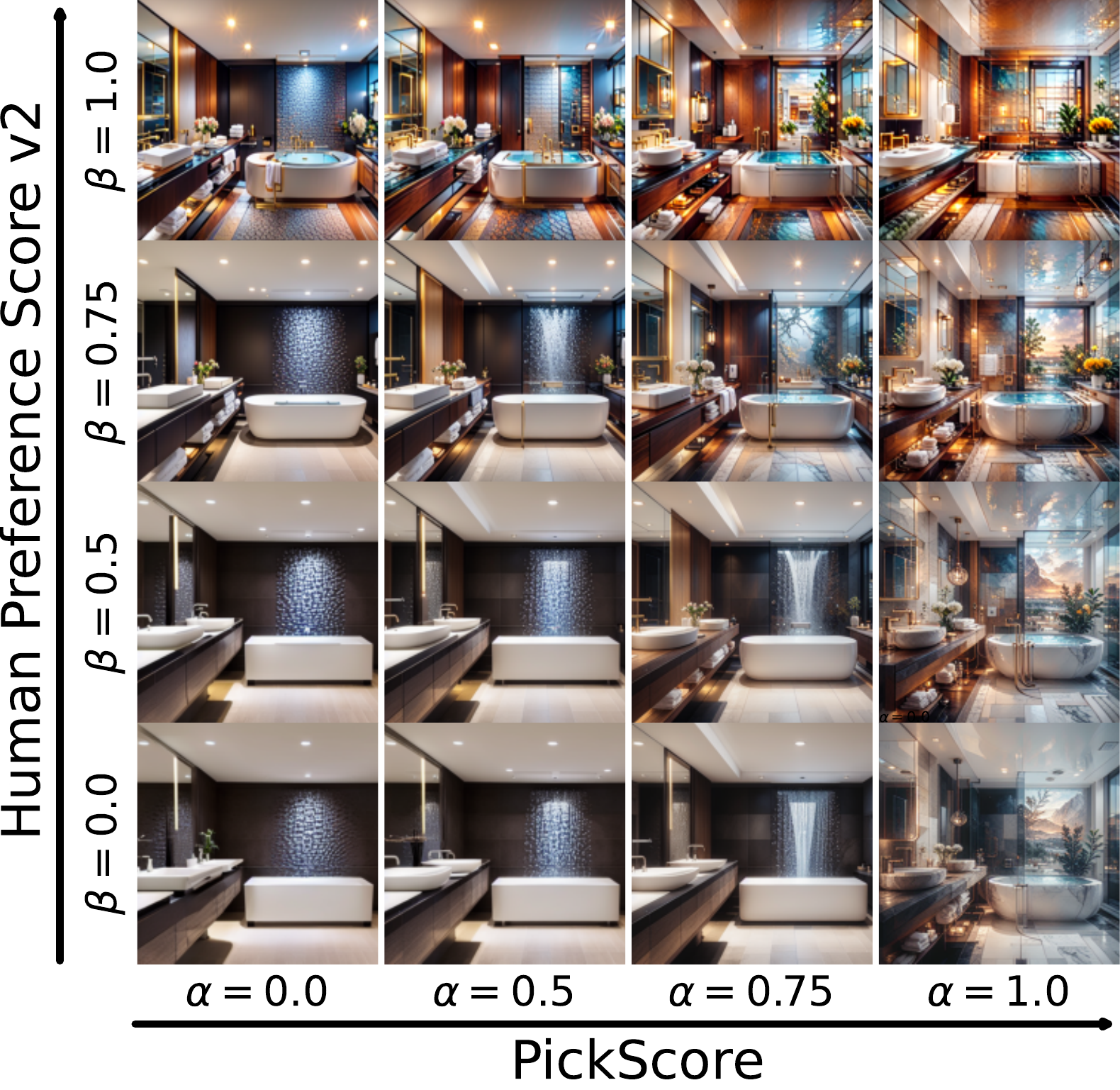}
    \includegraphics[width=0.49\linewidth]{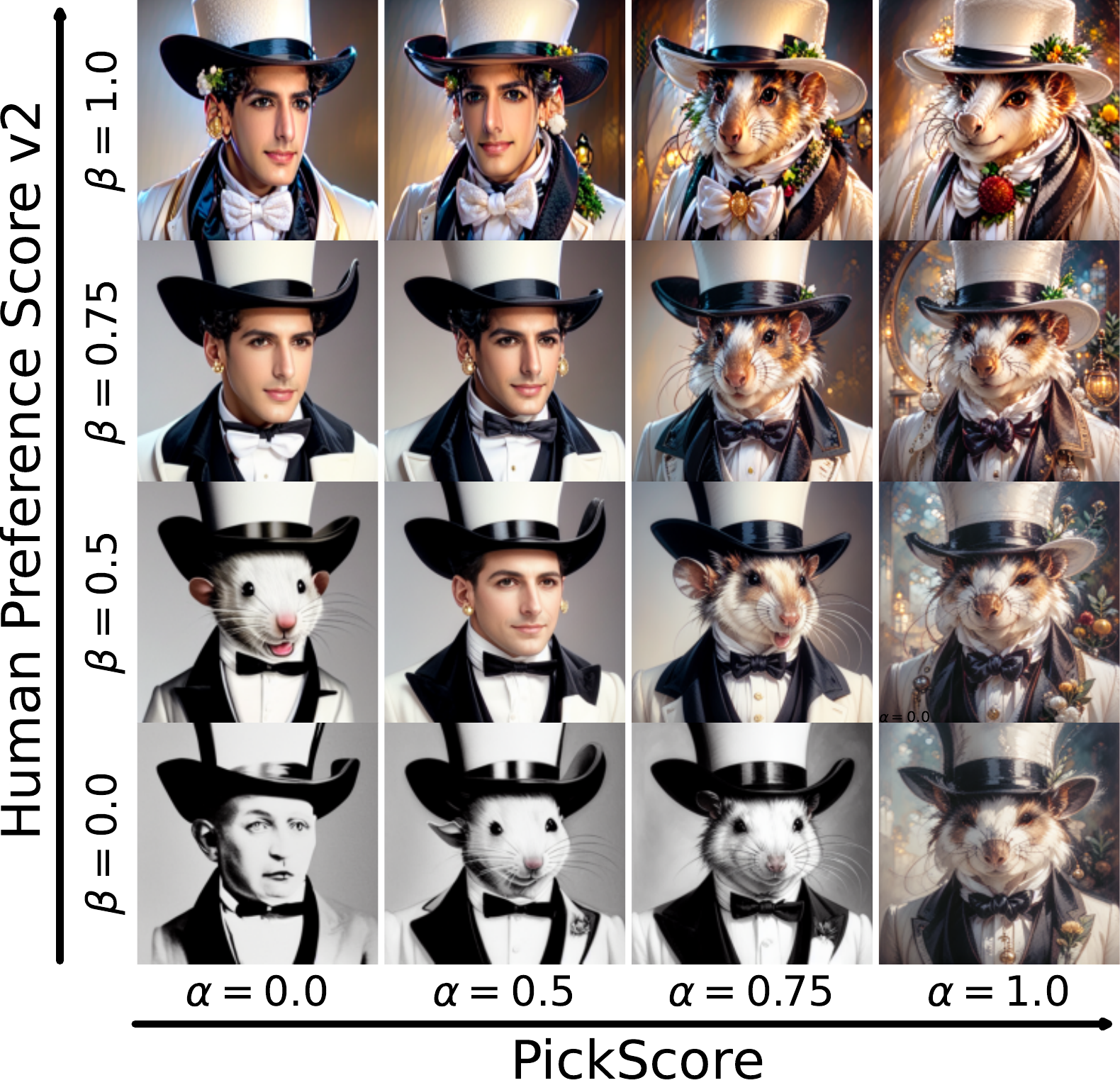}
    \vspace{-0.2cm}
    \caption{\small Additional results using linear combinations of LoRA parameters adapted for different rewards. In each subplot, we show images generated using LoRA parameters $\alpha \boldtheta^{\text{PickScore}}_{\text{LoRA}} + \beta \boldtheta^{\text{HPSv2}}_{\text{LoRA}}$ for coefficients $\alpha, \beta \in \{ 0.0, 0.5, 0.75, 1.0 \}$. We found that interpolating between LoRA weights can yield smooth transitions between different styles.}
    \label{fig:lora-multi-reward-interp4}
\end{figure}

Here, we provide additional results for scaling and interpolating LoRA parameters.
Figures~\ref{fig:lora-interp-hpsv2} and~\ref{fig:lora-interp-pickscore} show the effect of scaling LoRA parameters to control the strength of reward adaptation for the Human Preference Score v2 and PickScore rewards, respectively, and Figure~\ref{fig:lora-multi-reward-interp4} shows additional interpolations using linear combinations of LoRA weights trained separately for each of these rewards.

\subsection{Object Detection}
\label{app:additional-object-detection}

\begin{figure}[h]
    \centering
    \includegraphics[width=0.8\linewidth]{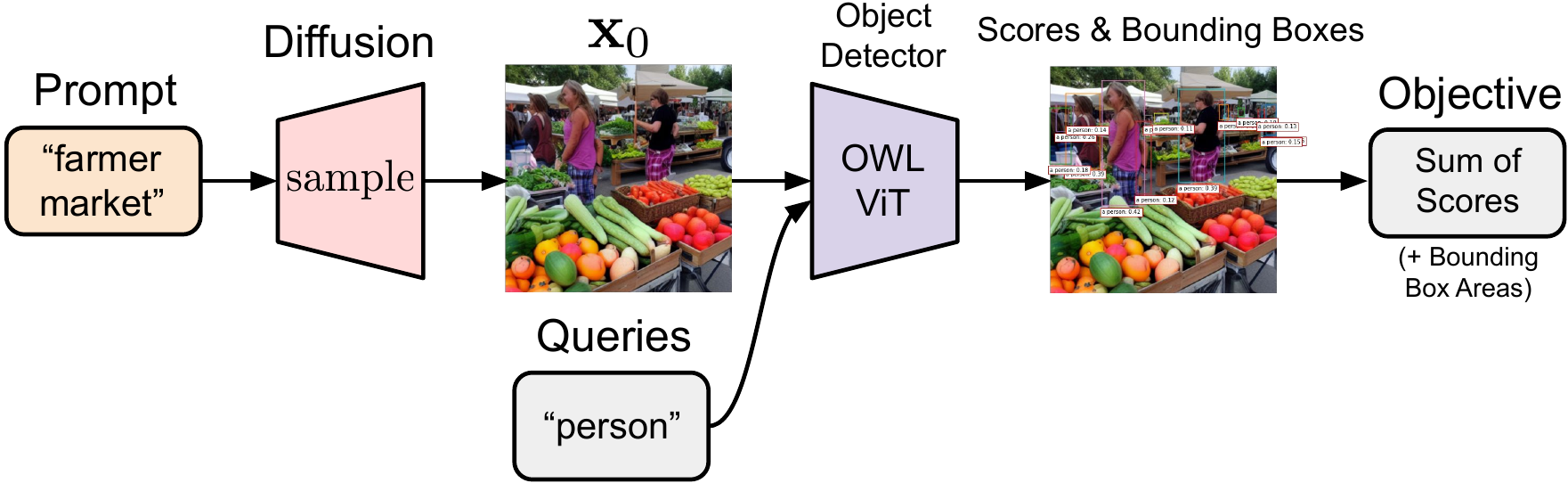}
    \caption{\small Fine-tuning for object detection using a pre-trained OWL-ViT model~\citep{minderer2022simple}.}
    \label{fig:object-detection}
\end{figure}

\begin{wrapfigure}[16]{r}{0.4\linewidth}
    \vspace{-0.6cm}
    \centering
    \includegraphics[width=\linewidth]{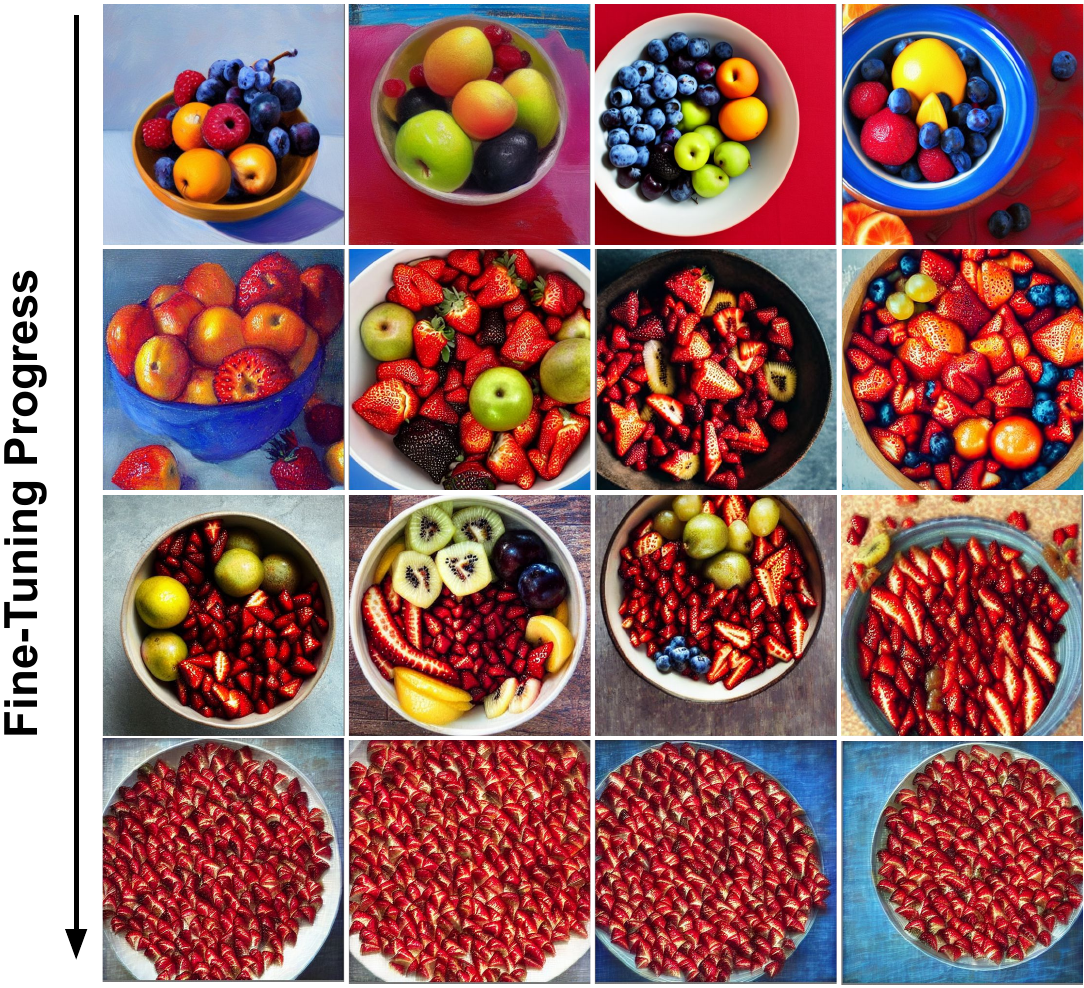}
    \vspace{-0.4cm}
    \caption{\small 
    Maximizing the object detector score for the query ``strawberry'' yields fruit bowls that contain progressively more strawberries over the course of fine-tuning.}
    \label{fig:adding-strawberries}
\end{wrapfigure}
Our setup for object detection is illustrated in Figure~\ref{fig:object-detection}.
We feed the generated images into a pre-trained Open-World Localization Vision Transformer (OWL-ViT)~\citep{minderer2022simple} model, along with a list of text queries that we wish to localize in the image.
OWL-ViT outputs bounding boxes and scores for the localized objects; we experimented with reward functions based on the sum of scores, as well as based on the areas of the bounding boxes, and found that both worked well.
Figure~\ref{fig:adding-strawberries} shows the fine-tuning progress for the diffusion prompt ``a bowl of fruit'' with OWL-ViT query ``strawberries.''

\subsection{Understanding the Impact of $K$}
\label{app:understanding-k}
\vspace{-0.2cm}

We investigated the impact of $K$ on the behavior of models fine-tuned with DRaFT-$K$ via ablations designed to answer the following questions: 1) Does adaptation occur only in the last $K$ steps (e.g., does the initial segment of the sampling chain behave similarly to the original pre-trained model, with a sudden jump occurring at the end)? and 2) Is it necessary to apply the LoRA parameters throughout the full sampling chain? What is the impact of applying the LoRA parameters to only the initial or final portions of the trajectory?

\begin{figure}[H]
    \centering
    \includegraphics[width=0.9\linewidth]{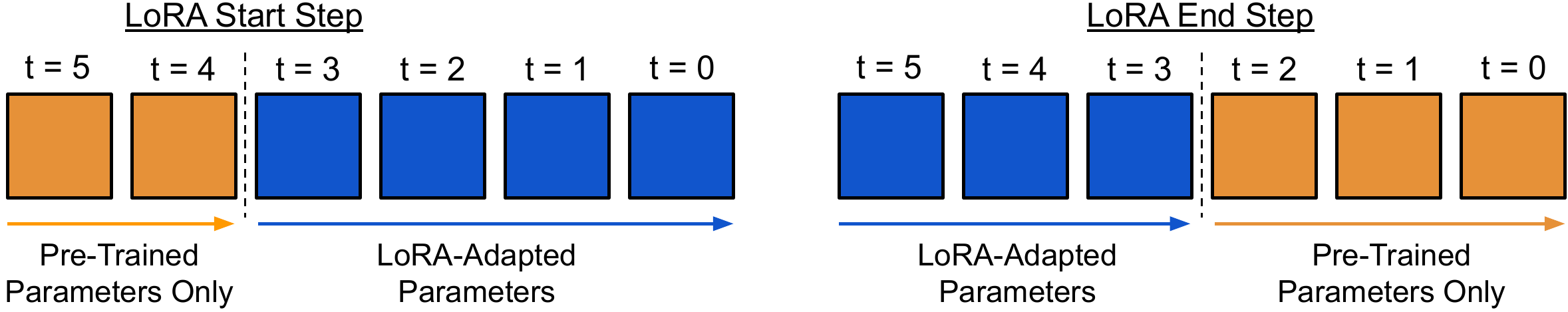}
    \vspace{-0.2cm}
    \caption{\small Illustration of sampling with different LoRA start steps and end steps.}
    \label{fig:lora-start-end-diagram}
    \vspace{-0.2cm}
\end{figure}

\begin{wrapfigure}[14]{r}{0.35\linewidth}
    \centering
    \includegraphics[width=\linewidth]{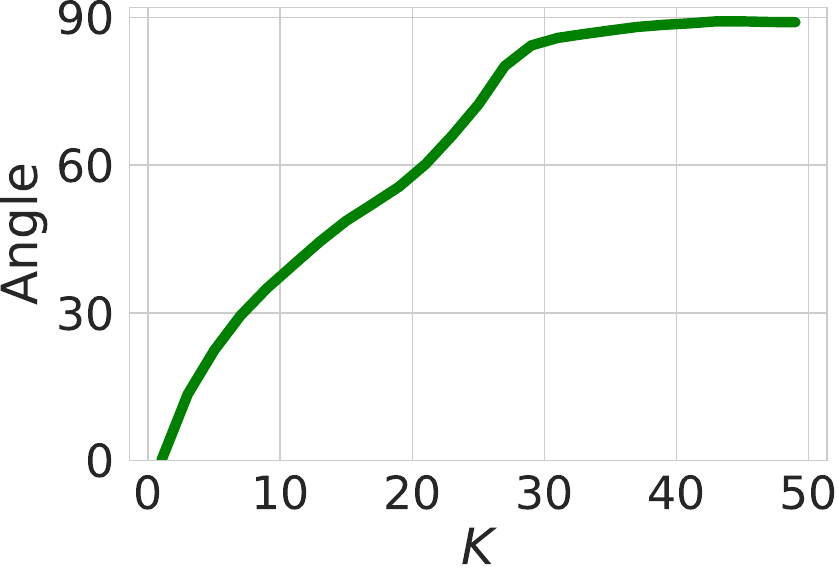}
    \vspace{-0.6cm}
    \caption{\small Angles between the DRaFT-1 gradient and DRaFT-$K$ gradients for $K \in \{ 1, \dots, 50 \}$.}
    \label{fig:draft-k-angles}
\end{wrapfigure}

\begin{figure}[H]
    \centering
    \includegraphics[width=0.7\linewidth]{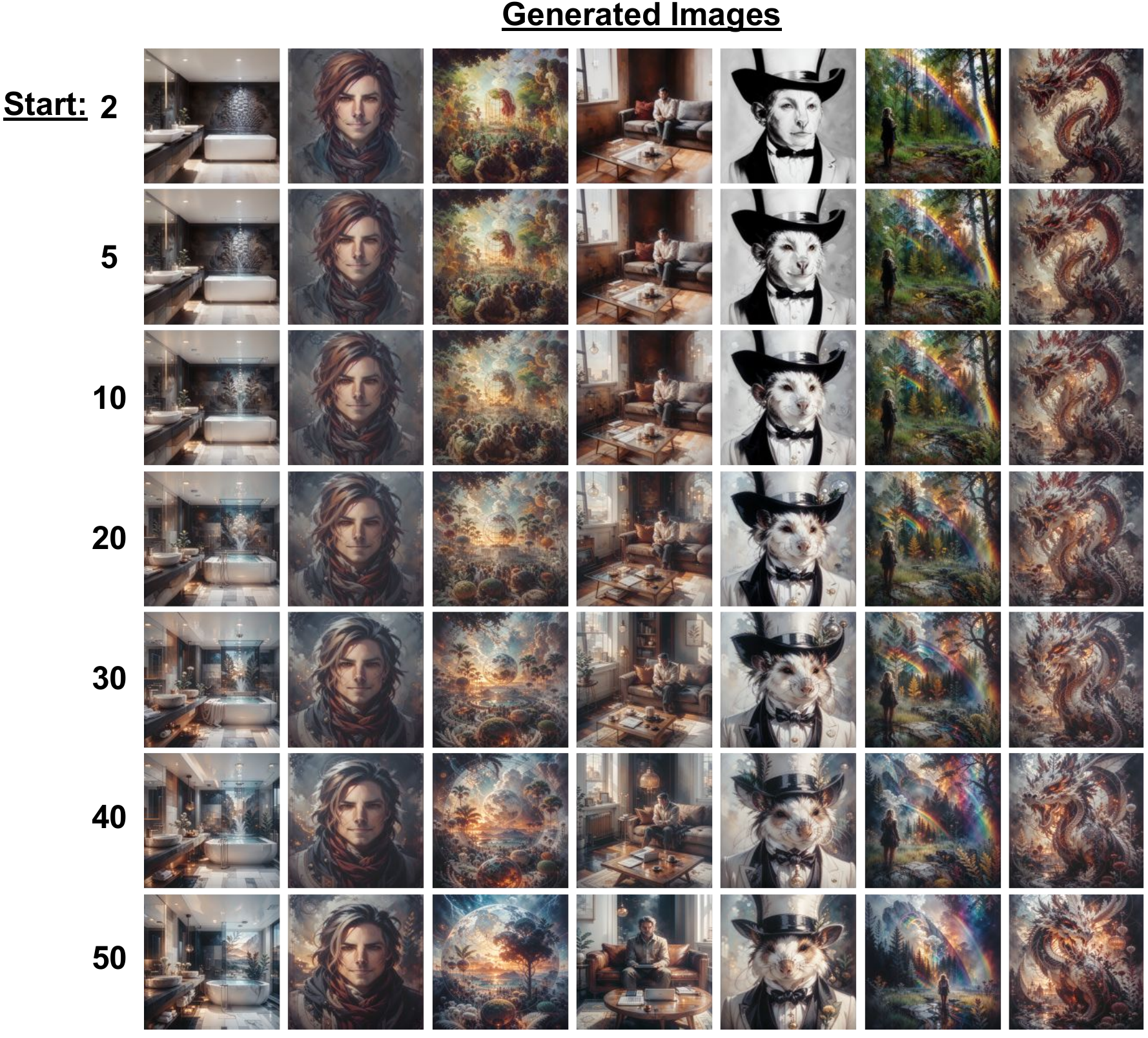}
    \vspace{-0.2cm}
    \caption{\small \textbf{Images generated with different LoRA start steps.} Here, the first $T-M$ steps of the diffusion sampling chain use the pre-trained diffusion model parameters, while the last $M$ steps use the LoRA-adapted parameters. The LoRA start step $M$ increases from top to bottom, $M \in \{ 2, 5, 10, 20, 30, 40, 50 \}$. We observe that, although the LoRA parameters are trained via truncated backpropagation through the last few steps of sampling, they affect the entire sampling chain—the images improve qualitatively when the LoRA-adapted parameters are used for many steps (e.g., $M > 10$).}
    \label{fig:lora-start-step}
\end{figure}

\begin{figure}[H]
    \centering
    \includegraphics[width=0.7\linewidth]{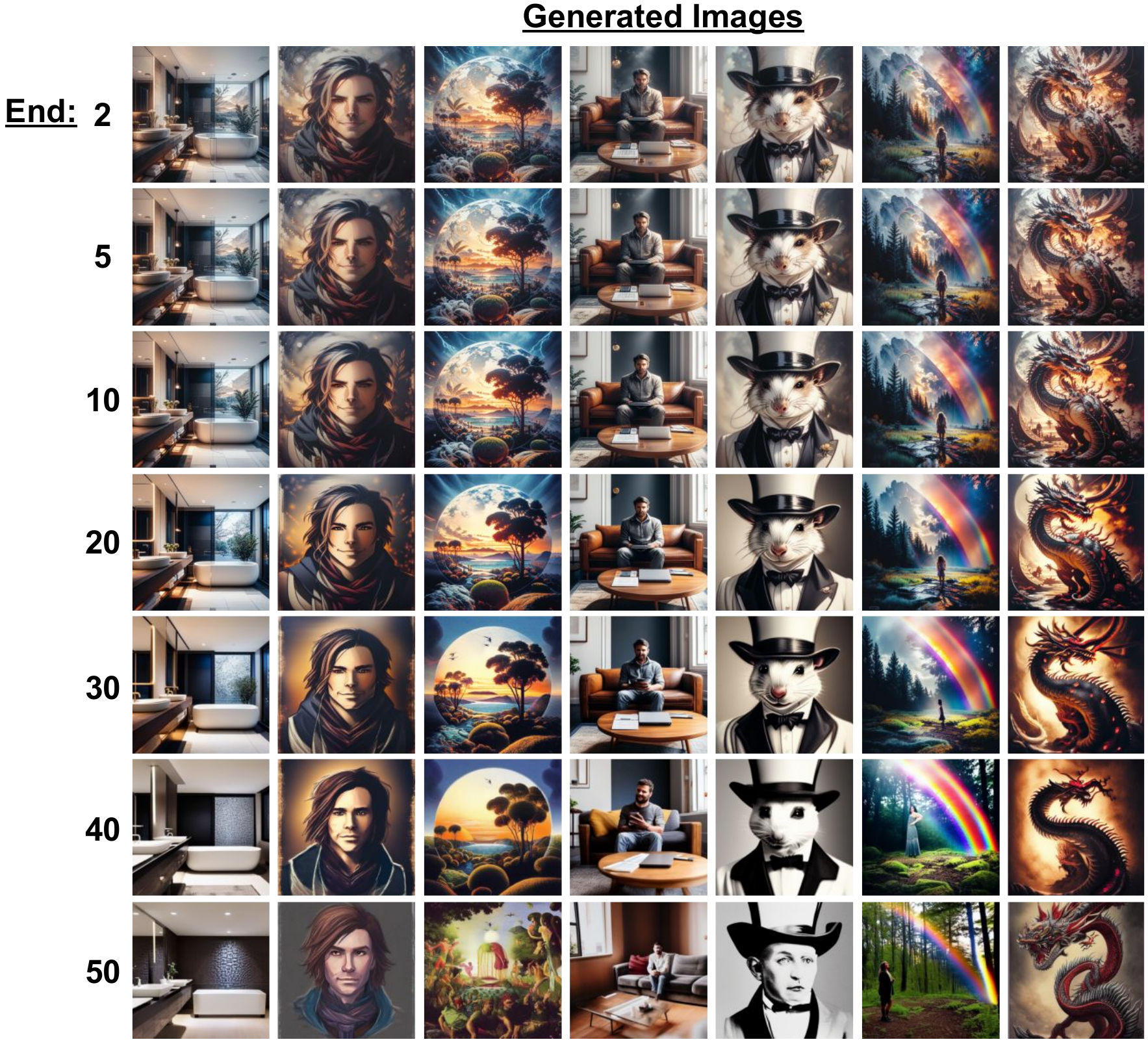}
    \vspace{-0.2cm}
    \caption{\small \textbf{Images generated with different LoRA end steps.} Here, the first $M$ steps of the diffusion sampling chain use the LoRA-adapted diffusion parameters, while the last $T-M$ steps use the un-adapted pre-trained model parameters. The LoRA end step $M$ increases from top to bottom, $M \in \{ 2, 5, 10, 20, 30, 40, 50 \}$. Interestingly, we observe that it is beneficial to apply the LoRA parameters early in sampling.}
    \label{fig:lora-end-step}
\end{figure}

To answer these questions, we applied the LoRA-adapted diffusion parameters for the last $M$ steps of the sampling chain, while in the first $T - M$ steps we used the pre-trained Stable Diffusion parameters without LoRA adaptation (see Figure~\ref{fig:lora-start-end-diagram}).
Figure~\ref{fig:lora-start-step} compares samples generated with different \textit{LoRA start iterations} $M$; interestingly, although the LoRA parameters are fine-tuned using truncated backpropagation through only the last $K$ steps of sampling, adaptation does \textit{not} happen only in the last $K$ steps; rather, the LoRA parameters need to be applied for at least 10-20 steps to yield substantial changes in the generated images.
Similarly, we also investigated the opposite scenario, where the LoRA-adapted parameters are only applied in the first $M$ steps of the sampling chain, after which LoRA is ``turned off'' and we use only the original pre-trained parameters for the remaining $T-M$ steps.
The generated images for a range of \textit{end LoRA steps} $M \in \{ 2, 5, 10, 20, 30, 40, 50 \}$ are shown in Figure~\ref{fig:lora-end-step}.
These results further demonstrate the importance of applying LoRA parameters in the early part of the sampling chain.

In Figure~\ref{fig:draft-k-angles}, we show the angle between the DRaFT-1 gradient and DRaFT-$K$ gradients for various $K$; the gradients become nearly orthogonal for $K > 30$.

\begin{figure}[H]
    \centering
    \includegraphics[width=0.6\linewidth]{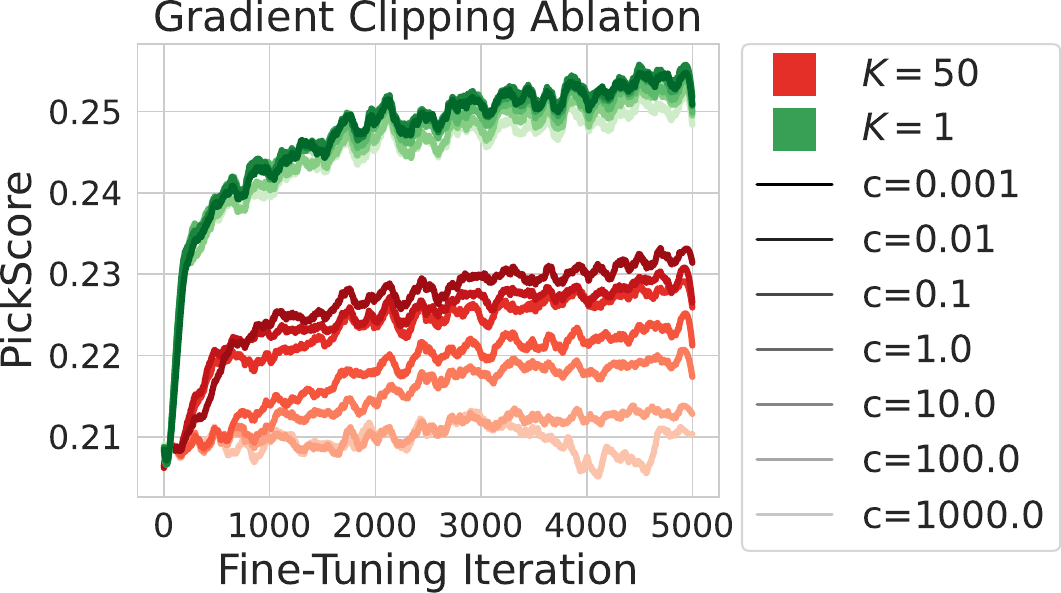}
    \vspace{-0.2cm}
    \caption{\small Ablation over the gradient clipping norm $c$ using DRaFT-$K$, fine-tuning to maximize PickScore on the HPDv2 prompt dataset. DRaFT-50 experiments are shown in red, and DRaFT-1 experiments are shown in green. In both cases, darker colors denote smaller gradient clipping norms.}
    \label{fig:gradient-clipping-ablation}
\end{figure}

\paragraph{Gradient Clipping.}
We used gradient clipping throughout all of our experiments.
In Figure~\ref{fig:gradient-clipping-ablation}, we show an ablation over the gradient clipping norm hyperparameter, which we denote $c$.
We compared the reward values achieved over the course of fine-tuning for PickScore on the HPD-v2 prompt dataset, with gradient clipping norms $c \in \{ 0.001, 0.01, 0.1, 1.0, 10.0, 100.0, 1000.0 \}$.
We found that, when using DRaFT-50, smaller gradient clipping norms ($c = 0.001$) improved optimization substantially, while large clipping norms ($c = 1000$) impeded training.
This further supports the observation that backpropagating through long diffusion sampling chains can lead to exploding gradients.
We performed a similar ablation for DRaFT-1, and found that the fine-tuning performance was nearly identical for all gradient clipping coefficients ($c = 0.001$ to $c=1000.0$).

\subsection{``Over-Optimization'' and Reward Hacking}
\label{app:over-optimization}

\paragraph{Improving Cross-Reward Generalization.}
DRaFT can cause models to overfit to the reward function, leading to high-reward but subjectively lower-quality images. 
Here, we explore regularization methods to reduce reward overfitting. 

As an approximate quantitative measure for over-optimization, we trained models on one reward function (HPSv2) and evaluated their generalization to another reward (PickScore). 
We considered three strategies to mitigate reward overfitting:
\begin{itemize}
    \item \textbf{LoRA Scaling}: multiplying the LoRA weights by a constant $<1$ (see Section~\ref{sec:preferences}), which interpolates between the pre-trained model and fine-tuned model, similarly to \citet{wortsman2022robust}.
    \item \textbf{Early Stopping}: evaluating various checkpoints across the training run instead of only the last one; early ones will achieve lower train reward but hopefully generalize better.
    \item \textbf{KL Regularization}: similarly to \citet{fan2023dpok}, we add a loss term to the model that penalizes the KL divergence between the pre-trained and fine-tuned model distributions. 
    This amounts to a squared error term between the  fine-tuned and pre-trained predicted $\boldepsilon$, that is, $\beta_{\text{KL}} \| \boldepsilon_{\text{finetuned}} - \boldepsilon_{\text{pretrained}} \|_2^2$, where $\beta_{\text{KL}}$ is the KL regularization coefficient.
    We computed the KL term using the predicted $\boldepsilon$ in the last denoising step.
    We trained several separate models, each with a different weight on the KL loss, which is a disadvantage over the previous methods that can be applied to a single training run.
\end{itemize}

\begin{figure}[H]
    \vspace{-0.2cm}
    \centering
    \includegraphics[width=0.7\linewidth]{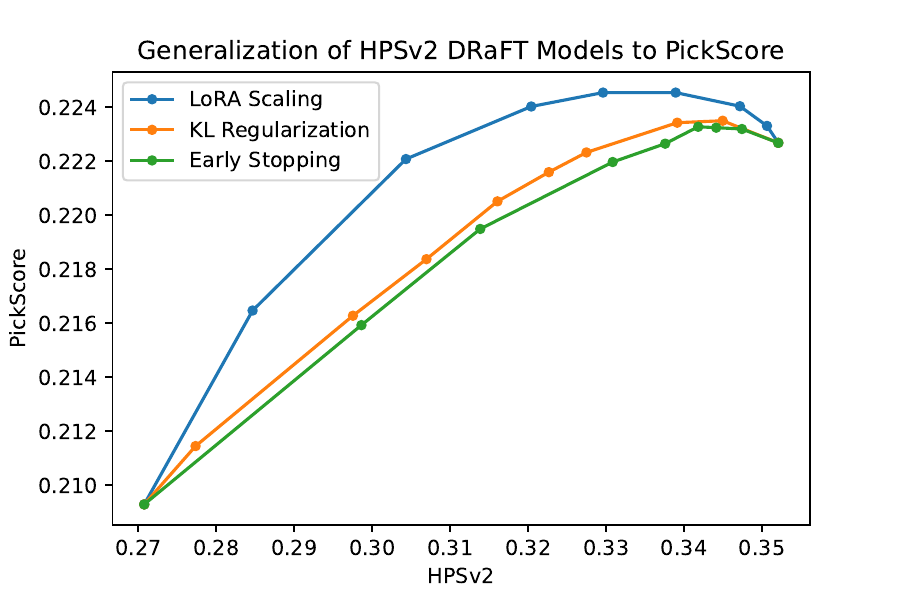}
    \vspace{-0.2cm}
    \caption{\small Generalization from HPSv2 to PickScore for various regularization methods used to prevent reward overfitting.}
    \label{fig:reward-generalization}
    \vspace{-0.4cm}
\end{figure}

All models used DRaFT-LV with the large-scale hyperparameters~(see Table~\ref{tab:hparams}).
The results are shown in Figure~\ref{fig:reward-generalization}. 
Both early stopping and KL regularization are commonly used to reduce reward overfitting in large language model preference tuning, but we found LoRA scaling to be more effective than the other approaches.

\begin{figure}[H]
    \centering
    \includegraphics[width=0.46\linewidth]{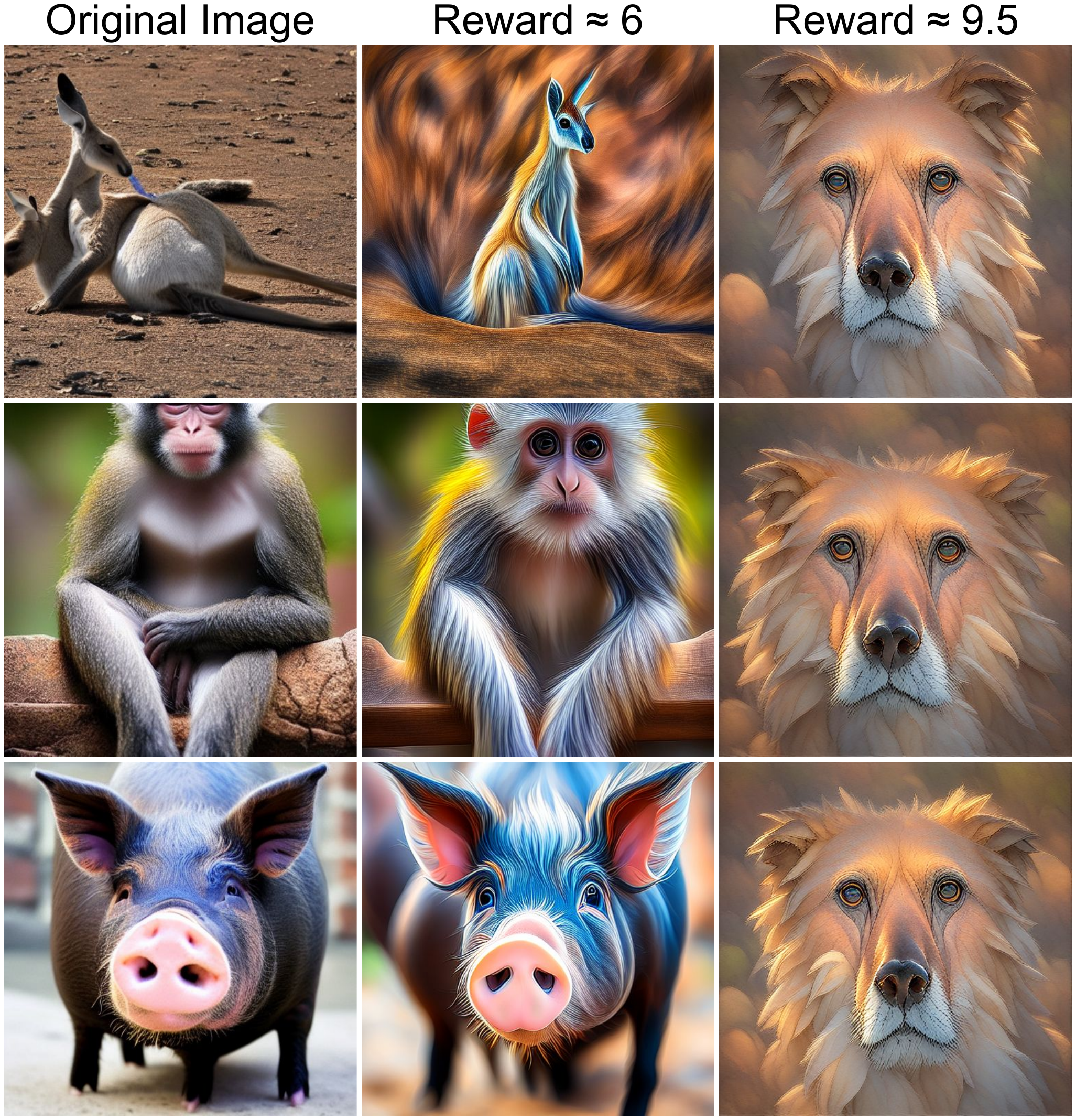}
    \hfill
    \includegraphics[width=0.46\linewidth]{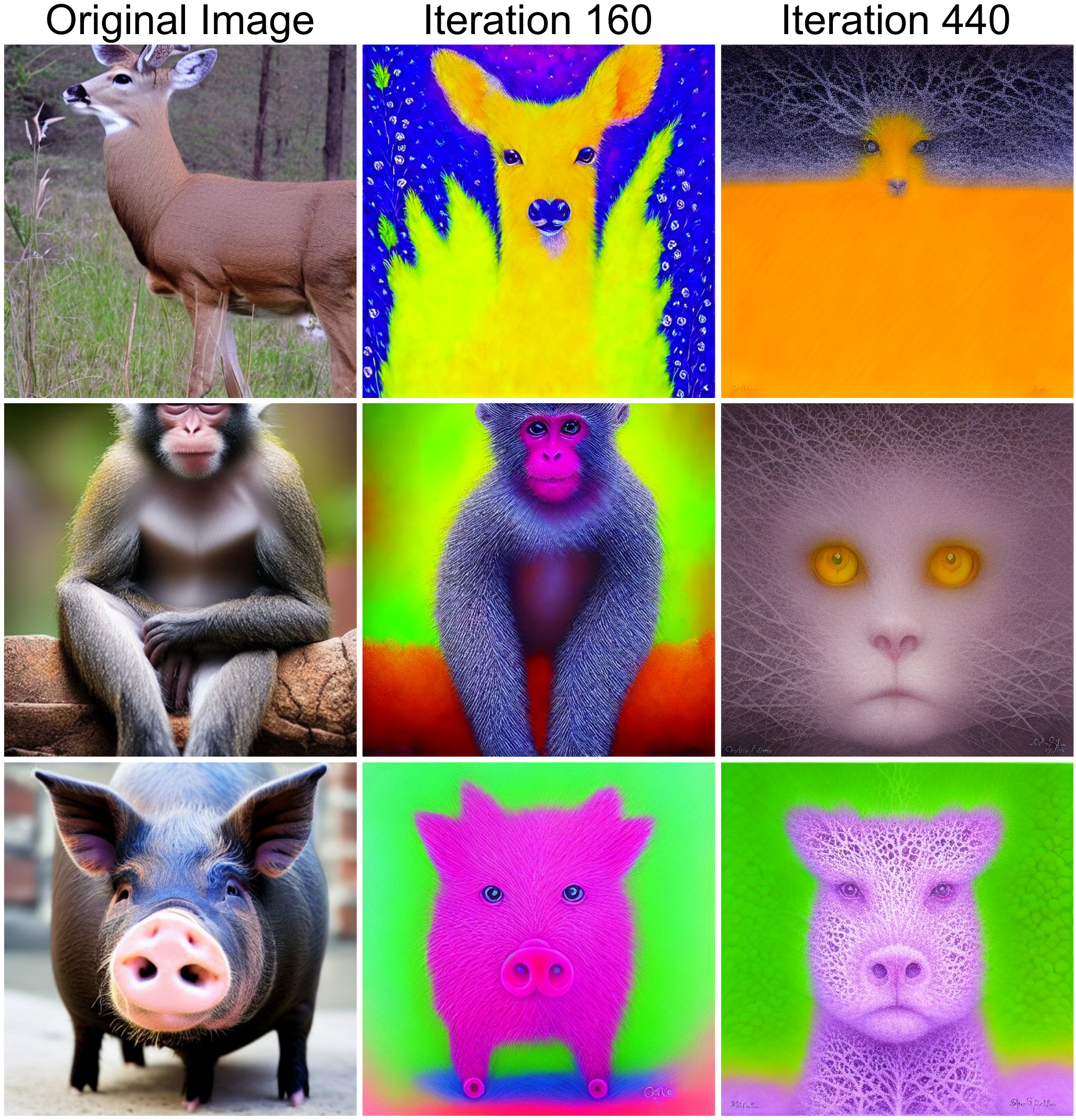}
    \vspace{-0.2cm}
    \caption{\small Reward hacking still occurs when we use large dropout rate in the Aesthetic classifier (Left). Incorporating a reward term to encourage diversity can lead to overly stylized images (Right).}
    \label{fig:aesthetic-dropout}
\end{figure}

\paragraph{Attempts to Increase Diversity.}
A common result of reward hacking is a collapse to a single, high-reward image.
Here, we describe two of our attempts to address this lack of diversity.
First, we tried adding dropout to the reward function (in this case, the aesthetic classifier); however, we found that even large dropout rates such as 0.95 led to diversity collapse, as shown in Figure~\ref{fig:aesthetic-dropout} (Left).
Next, we tried adding a term to the reward that measures the dissimilarity between images generated in the same minibatch---this encourages the model to generate different images for different prompts and noise samples.
Ideally, one would take the mean dissimilarity over all pairs of examples in a minibatch, yielding $B \choose 2$ combinations where $B$ is the minibatch size.
However, for computational tractability, we used a simpler approach, where we formed pairs by reversing the elements of a minibatch and computing elementwise dissimilarities, giving us $B / 2$ pairs.
We tried two measures of dissimilarity: 1) the Euclidean distance, and 2) the Learned Perceptual Image Patch Similarity (LPIPS)~\citep{zhang2018perceptual}.
Neither was satisfactory to increase diversity: in Figure~\ref{fig:aesthetic-dropout} (Right), we show images generated using a large weight on the LPIPS diversity term, which increased inner-batch diversity at the expense of the aesthetic score.

\subsection{Rotational Anti-Correlation Reward}
\label{sec:rotation}
\begin{wrapfigure}[10]{r}{0.35\linewidth}
    \vspace{-0.4cm}
    \centering
    \includegraphics[width=\linewidth,trim={3.2cm 0 2.6cm 0},clip]{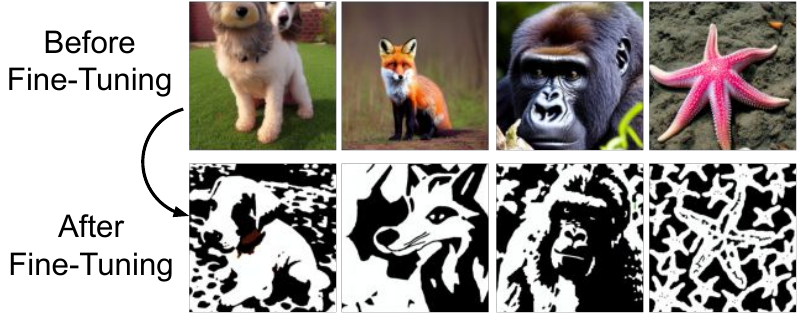}
    \vspace{-0.6cm}
    \caption{\small Rotational anti-correlation.}
    \label{fig:rotation-anticorr-results}
\end{wrapfigure}
We also explored fine-tuning for symmetry or anti-symmetry properties.
Here, we optimize for rotational anti-correlation, which encourages the diffusion model to generate images that are \textit{dissimilar} to their rotations at $\{ 90, 180, 270 \}$ degrees.
Given an image $\boldx$ and rotation operator $\text{Rot}(\boldx, 90^{\circ})$, we formulate the objective as:
$
\mathbb{E}_{\boldc \sim p_{\boldc}, \boldx \sim \text{sample}(\boldtheta, \boldc)} \Big[ \frac{1}{3} \Big( \| \boldx - \text{Rot}(\boldx, 90^{\circ}) \|^2 + \| \boldx - \text{Rot}(\boldx, 180^{\circ}) \|^2 + \| \boldx - \text{Rot}(\boldx, 270^{\circ}) \|^2 \Big) \Big]
$.
The results are shown in Figure~\ref{fig:rotation-anticorr-results}: this objective leads to interesting black-and-white, stylized images.

\subsection{CLIP Reward}
\label{sec:clip}

\begin{figure}[H]
    \centering
    \includegraphics[width=0.7\linewidth]{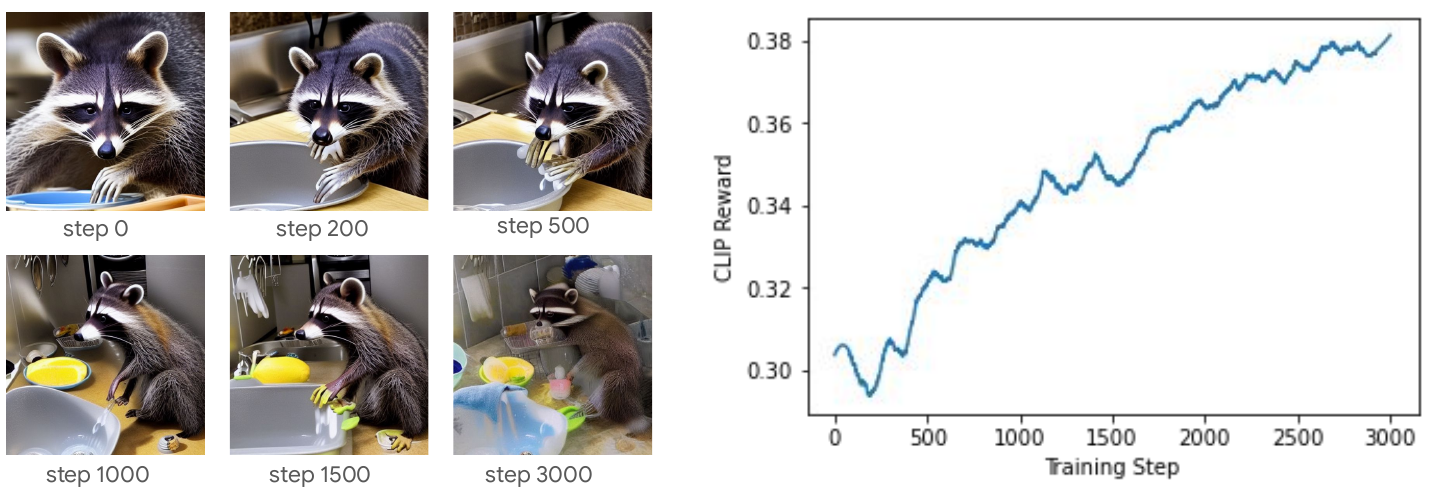}
    \vspace{-0.2cm}
    \caption{\small Qualitative examples (left) and training curve (right) for fine-tuning on CLIP reward using DRaFT for the prompt ``a raccoon washing dishes".}
    \label{fig:clip}
\end{figure}

We briefly explored using $r(\boldx_0, \boldc) = \text{CLIP-Similarity}(\boldx_0, \boldc)$ as a reward to improve image-text alignment. We found that optimization was successful in that CLIP scores improved, but that image quality degraded after many steps of training (see Figure~\ref{fig:clip}).

Both HPSv2 and ImageReward are fine-tuned CLIP models, and for HPSv2 the annotators were specifically instructed to consider alignment to the prompt in their preference judgements. 
Therefore, they already capture some aspects of image alignment, and qualitatively we did observe alignment to increase in some cases when using these rewards (e.g. in the ``a raccoon washing dishes'' example in Figure~\ref{fig:main-figure}).
However, we think improving text alignment using a powerful image captioning model such as PaLI \citep{chen2022pali} as the reward would be an interesting future direction.

\subsection{Fine-Tuning for Safety}
\label{app:safety}

The datasets on which large diffusion models are trained often contain NSFW content, which poses challenges for the deployment of safe models.
Safety measures can be implemented at various stages of the text-to-image pipeline: 1) at the text prompt level (e.g., refusing NSFW words); 2) during the generative process (e.g., encouraging the model to output safe images); and 3) post-hoc elimination of NSFW images that have been generated.
These approaches are complementary, and in practice we imagine that all three would be used in concert.
Here, we focus on stage (2), by fine-tuning diffusion models using feedback from a pre-trained NSFW classifier.
This approach is particularly useful for ambiguous prompts that may lead to either SFW or NSFW generations; for example, the prompt ``swimsuit’’ can refer to images of people wearing swimsuits, or to just the article of clothing itself (the latter being more safe-for-work).
We found that, before fine-tuning, the model mostly generated images of people wearing swimsuits, while after fine-tuning, the model instead generated images of lay-flat swimsuits.

\subsection{Learning Sampler Hyperparameters}
\label{app:learning-sampler}

$$
$$

\begin{wrapfigure}[10]{r}{0.35\linewidth}
    \vspace{-0.5cm}
    \centering
    \includegraphics[width=1.0\linewidth]{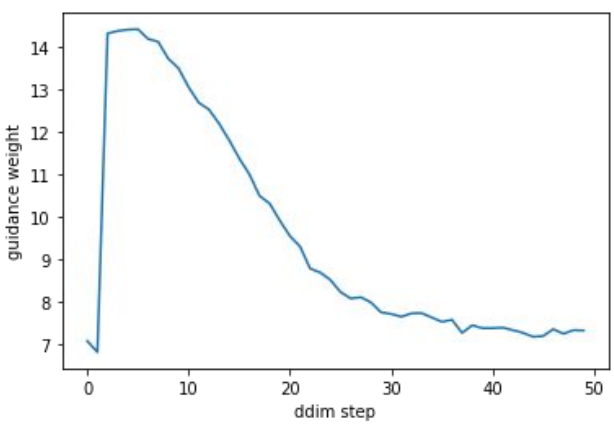}
    \vspace{-0.6cm}
    \caption{\small Guidance schedule learned by DRaFT}
    \label{fig:guidance}
\end{wrapfigure}
\vspace{-0.4cm}
As DRaFT (without truncated backprop through time) computes gradients through the denoising process end-to-end, we can use DRaFT to optimize sampler hyperparameters as well as model weights.

Learning sampler hyperparameters is similar in spirit to \citet{watson2022learning}, except we obtain gradients from a reward model rather than using kernel inception distance.
We explored learning two inputs to the sampler---the guidance weights and negative prompt embeddings---although it would be possible to extend this method to other hyperparameters such as the time schedule.
In particular, we investigated:
\begin{itemize}
    \item \textbf{Guidance weights:} We learn a separate guidance weight $w(t)$ for each step of DDIM sampling. The learned schedule can be applied to samplers with different number of steps through linearly interpolating the schedule to stretch or compress it. We use a 50x larger learning rate for training guidance than the LoRA parameters because guidance weights have a large magnitude. An example learned guidance schedule is shown in Figure~\ref{fig:guidance}.
    \item \textbf{Negative prompt:} We learn prompt-dependent text embeddings passed into the unconditional UNet used for classifier-free guidance. In particular, we parameterize the embedding as $\text{MLP}(\text{Multi-Headed-Self-Attention}(e))$ where $e$ is the text embeddings produced by the diffusion model's text encoder. 
\end{itemize}
However, in practice, we found that this did not improve results much at large scale.

\section{Method extensions}
\label{app:extensions}

Here we briefly discuss two gradient-based reward fine-tuning methods we explored, but which we found did not achieve strong results compared to DRaFT-1 and DRaFT-LV. 

\subsection{Single-Reward DRaFT-LV}
A downside of DRaFT-LV is that it requires computing the reward function gradient an additional $n$ times. While this is relatively little overhead for the reward functions we used compared to the cost of image sampling, larger reward models could make this costly. We therefore experimented with a version of DRaFT-LV that only computes the reward gradient with respect to the input pixels $\nabla_\boldx r(\boldx_0, \boldc)$  once. Then it updates the gradient in the inner loop as $\boldg = \boldg + \nabla_\boldx r(\boldx_0, \boldc) \nabla_\boldtheta \hat{\boldx}_0$. Unfortunately, we found this method to be unstable and ultimately not learn effectively, possibly because it relies on the assumption that $\nabla_\boldx r(\boldx_0, \boldc) \approx \nabla_\boldx r(\hat{\boldx}_0, \boldc)$, which may not be true in practice. 

\subsection{Deterministic Policy Gradient}

Previous approaches applying RL to reward learning do not make use of sampling gradients. Here we discuss how Deterministic Policy Gradient (DPG; \citealt{silver2014deterministic,lillicrap2015continuous}) offers a way of employing reward gradients within the RL framework.

Similarly to \citet{black2023training}, we can view DDIM sampling as applying a deterministic policy to a deterministic Markov decision process:
\begin{itemize}
    \item A state $\bolds_t$ consists of the current latent, timestep, and prompt $(\boldx_t, t, \boldc)$.
    \item The initial state $\bolds_0$ draws $\boldx_T \sim \mathcal{N}(\boldzero, \boldI), \boldc \sim p_{\boldc}$, and sets $t=T$.
    \item The policy is the learned denoiser: $\bolda_t = \boldmu_\boldtheta(\bolds_t) = \boldepsilon_{\boldtheta}(\boldx_t, \boldc, t)$
    \item The transition function performs a DDIM sampling step: \\ $f(\bolds_t, \bolda_t) = (\frac{\alpha_{t - 1}}{\alpha_t}(\boldx_t - \sigma_t \bolda_t) + \sigma_{t - 1}\bolda_t, t - 1, \boldc)$
    \item The reward $R(\bolds_t, \bolda_t)$ is $r(\boldx_0, \boldc)$ if $t = 0$ and 0 if otherwise.
\end{itemize}

DPG trains a critic $Q_{\boldphi}(\bolds_t, \bolda_t)$ and learns $\boldtheta$ with the gradient update $\nabla_\bolda Q_{\boldphi}(\bolds_t, \bolda_t) \nabla_\boldtheta \boldmu_\boldtheta(\bolds_t)$ over sampled trajectories. 
The critic is trained to minimize the squared error between its prediction and the final return (in our case $r(\boldx_0, \boldc)$).

Many of our rewards make use of neural networks with pre-trained parameters $\boldxi$, which we denote by writing $r_{\boldxi}$. We suggest applying LoRA to $\boldxi$ parameterize the critic; we denote the LoRA-adapted parameters as $\text{adapt}(\boldxi, \boldphi)$. 
We then apply the critic by (1) executing the action $\bolda_t$ on $\bolds_t$, (2) 1-step denoising the result to get a predicted clean image, and (3) applying the adapted reward model $r_{\text{adapt}(\boldxi, \boldphi)}$ to the result.
The reason for step (2) is to make the training of $\boldphi$ more efficient: we believe it is easier to adapt the reward model parameters to produce good expected return estimates for one-step-denoised inputs than it is for noisy inputs. Formally, we use the critic:
\begin{equation}
    Q_{\boldphi}(\bolds_t, \bolda_t) = r_{\text{adapt}(\boldxi, \boldphi)}((f(\bolds_t, \bolda_t) - \sigma_t \boldepsilon_{\boldtheta}(f(\bolds_t, \bolda_t), \boldc, t - 1)) / \alpha_t, \boldc)
\end{equation}

When computing gradients through $Q$, we do not backpropagate through step (2) into $\boldtheta$, to prevent it from interfering with the policy training. 

\begin{figure}[tb]
    \centering
    \includegraphics[width=0.95\linewidth]{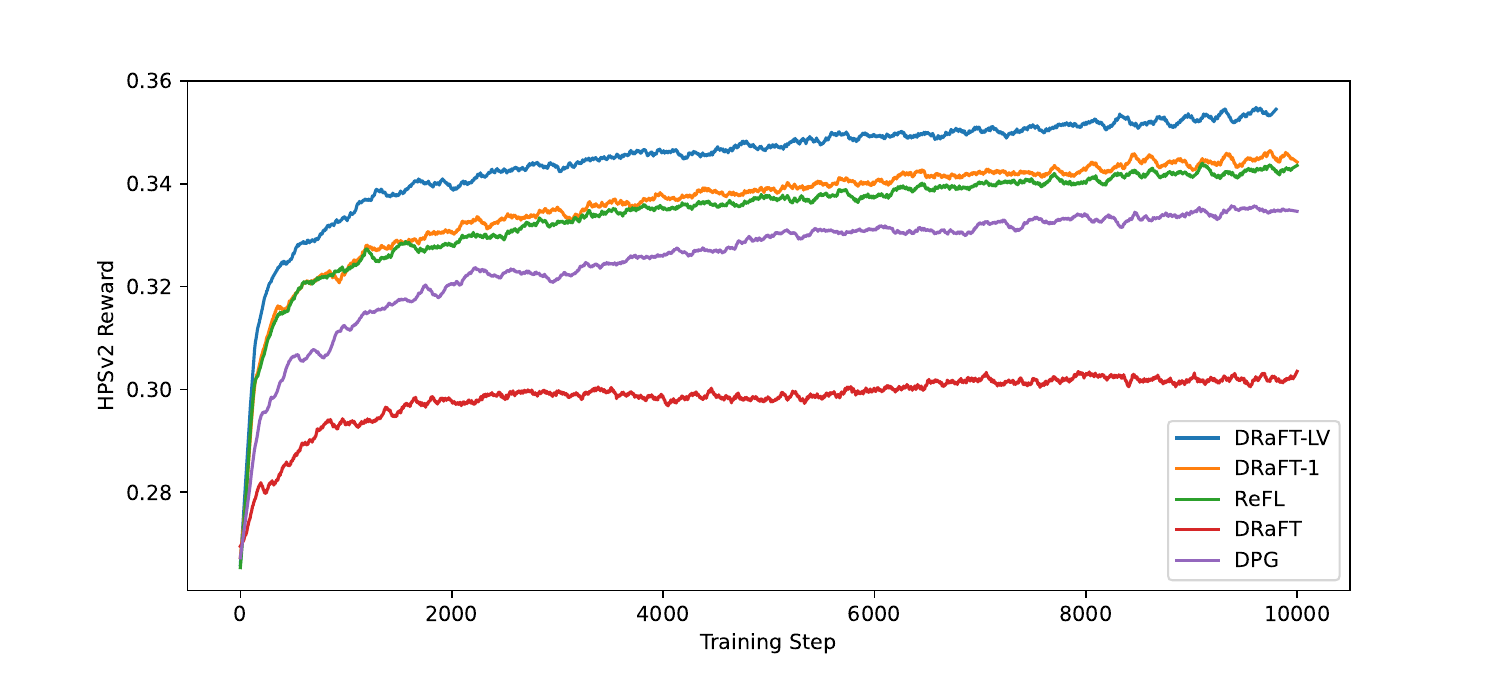}
    \caption{\small Training efficiency of reward fine-tuning methods on the HPSv2 reward.}
    \label{fig:dpg}
\end{figure}

\paragraph{Results.} The training efficiency of DPG compared with other methods is shown in Figure~\ref{fig:dpg} (using the same experimental settings as in Figure~\ref{fig:bar-chart}). DPG outperforms vanilla DRaFT, but not our more-efficient variants. 
One challenge is that $Q$ produces poor return estimates for large timesteps, which could perhaps be alleviated with a ReFL-style version that skips or rarely samples high timesteps for training the policy. 
There are many variants of DPG to explore, so we think that improving the performance of DPG for reward fine-tuning could be an interesting direction for future research; DPG is appealing because like vanilla DRaFT (but unlike ReFL and our DRaFT variants), it optimizes the full sampling process in an unbiased way.

\section{Extended Related Work}
\label{app:extended-related-work}

\vspace{-0.2cm}
\paragraph{Diffusion Models.}
Denoising diffusion probabilistic models (DDPMs; \citealt{sohl2015deep,song2019generative,song2020score,ho2020denoising}) are a class of generative models that have become the de-facto standard for most continuous data modalities, including images~\citep{ramesh2021zero}, videos~\citep{ho2022imagen,singer2022make}, audio~\citep{liu2023audioldm}, and 3D models~\citep{zeng2022lion,poole2022dreamfusion,gu2023nerfdiff}.
Text-to-image diffusion models, which generate images conditioned on text prompts, have become prominent tools with the advent of models such as GLIDE~\citep{nichol2021glide}, DALLE-2~\citep{ramesh2022hierarchical}, Imagen~\citep{saharia2022photorealistic,ho2022imagen}, and Latent Diffusion~\citep{rombach2022high}.
In this paper we focus on fine-tuning text-to-image models, but DRaFT is general and could be applied to other diffusion models or fine-tuning tasks, and to different domains such as audio or video.

\vspace{-0.2cm}
\paragraph{Learning human preferences.} 
Human preference learning  trains models on judgements of which behaviors people prefer, rather than on human demonstrations directly \citep{knox2009interactively,akrour2011preference}.
This training is commonly done by learning a reward model reflecting human preferences and then learning a policy that maximizes the reward \citep{christiano2017deep,ibarz2018reward}. 
Preference learning methods such as reinforcement learning from human feedback (RLHF) have become widely used for fine-tuning large language models so they improve at summarization \citep{stiennon2020learning,wu2021recursively}, instruction following \citep{ouyang2022training}, or general dialogue \citep{askell2021general,bai2022training,glaese2022improving,bai2022constitutional}.
We apply DRaFT to optimize scores from existing preference models, such as PickScore \citep{kirstain2023pick} and Human Preference Score v2 \citep{wu2023human}, which are trained on human judgements between pairs of images generated by diffusion models for the same prompt. 

\vspace{-0.2cm}
\paragraph{Guidance.}
Guidance steers the sampling process towards images that satisfy a desired objective by adding an auxiliary term to the score function.
The canonical example is classifier guidance~\citep{song2020score,dhariwal2021diffusion}, which can turn an unconditional diffusion model into a class-conditional model by biasing the denoising process using gradients from a pre-trained classifier.
Many other pre-trained models can also be used to provide guidance signals, including segmentation models, facial recognition models, and object detectors~\citep{bansal2023universal}.
However, care must be taken when computing the guidance signal; na\"ively, noisy examples obtained during the iterative refinement process may be out-of-distribution for the guidance model, leading to inaccurate gradients.
Two approaches for mitigating this are (1) training a guidance model from scratch on noisy data~\citep{dhariwal2021diffusion} and (2) applying a pre-trained guidance model (which has seen only non-noisy data) to the one-step-denoised images obtained by removing the predicted noise at time $t$~\citep{li2022upainting}.
However, both approaches have downsides: (1) precludes the use of off-the-shelf pre-trained guidance functions and (2) means the guidance is applied to out-of-distribution blurry images for high noise levels.
DRaFT avoids these issues by backpropagating through sampling so the reward function is only applied to the fully denoised image. 

\vspace{-0.2cm}
\paragraph{Backpropagation through diffusion sampling.}

Similarly to our method,~\citet{watson2022learning} backpropagate through sampling using the reparameterization trick and gradient checkpointing.
They learn parametric few-step samplers that aim to reduce the inference cost while maintaining perceptual image quality, using a differentiable image quality score.
\citet{fan2023optimizing} also propose an approach to speed up sampling: they consider doing this using gradients by backpropagating through diffusion sampling, but due to memory and exploding/vanishing gradient concerns, they focus on an RL-based approach. Both~\citet{fan2023optimizing} and~\citet{watson2022learning} aim to improve sampling speed, rather than optimize arbitrary differentiable rewards, which is our focus.
Like DRaFT, Direct Optimization of Diffusion Latents (DOODL; \citealt{wallace2023end}) uses backpropagation through sampling to improve image generation with respect to differentiable objectives. 
Rather than optimizing model parameters, DOODL optimizes the initial noise sample $\boldx_T$.
Although DOODL does not require a training phase, it is much slower at inference time because the optimization must be redone for each prompt and metric. In contrast, after finetuning with DRaFT, sampling has the same cost as a standard diffusion model.
Furthermore, the latent optimization methods do not support the compositionality and interpolation properties that we can achieve with different sets of LoRA parameters tuned for different objectives.

Outside of image generation, \citet{wang2023diffusion} train diffusion model policies for reinforcement learning by backpropagating the Q-value function through sampling. 

\vspace{-0.2cm}
\paragraph{Reward fine-tuning with supervised learning.}
\citet{lee2023aligning} and \citet{wu2023better} use supervised approaches to fine-tune diffusion models on rewards. These methods generate images with the pre-trained model and then fine-tune on the images while weighting examples according to the reward function or discarding low-reward examples. Unlike DRaFT or RL methods, the model is not trained online on examples generated by the current policy.
However, \citet{dong2023raft} use an online version of this approach where examples are re-generated over multiple rounds of training, which can be viewed as a simple kind of reinforcement learning.

\vspace{-0.2cm}
\paragraph{Reward fine-tuning with reinforcement learning.}
\citet{fan2023optimizing} interpret the denoising process as a multi-step decision-making task and use policy gradient algorithms to fine-tune diffusion samplers.
Building on it, \citet{black2023training} and~\citet{fan2023dpok} use policy gradient algorithms to fine-tune diffusion models for arbitrary black-box objectives.
Rather than optimizing model parameters, \citet{hao2022optimizing} apply RL to improve the input prompts.
RL approaches are flexible because they do not require differentiable rewards. However, in practice many reward functions are  differentiable, or can be re-implemented in a differentiable way, and thus analytic gradients are often available.
In such cases, using reinforcement learning discards useful information, leading to inefficient optimization.

\vspace{-0.2cm}
\paragraph{ReFL.}
\label{sec:refl}

Reward Feedback Learning (ReFL; \citealt{xu2023imagereward}) evaluates the reward on the one-step predicted clean image, $r(\hat{\boldx}_0, \boldc)$ from a randomly-chosen step $t$ along the denoising trajectory, and backpropagates through the one-step prediction w.r.t. the diffusion model parameters.
DRaFT-$K$ is conceptually simpler than ReFL, as it only differentiates through the last few steps (e.g., $K=1$) of sampling, where $K$ is deterministic; in~\citep{xu2023imagereward}, the authors randomly choose an iteration between a min and max step of the sampling chain (which incurs more hyperparameters) from which to predict the clean image.
Also, because DRaFT runs the full sampling chain, our reward functions are always evaluated on final generations. In contrast, ReFL applies the rewards to one-step denoised latents in the sampling trajectory, similar to the use of one-step denoising in guidance.
Furthermore, DRaFT-LV is substantially more efficient than ReFL, speeding up training by approximately $2 \times$ by reducing the variance of the single-denoising-step gradient estimates.

\clearpage

\vspace{-0.2cm}
\section{Uncurated Samples}
\label{app:uncurated-samples}
\vspace{-0.2cm}

\begin{figure}[H]
    \centering
    \includegraphics[width=0.95\linewidth]{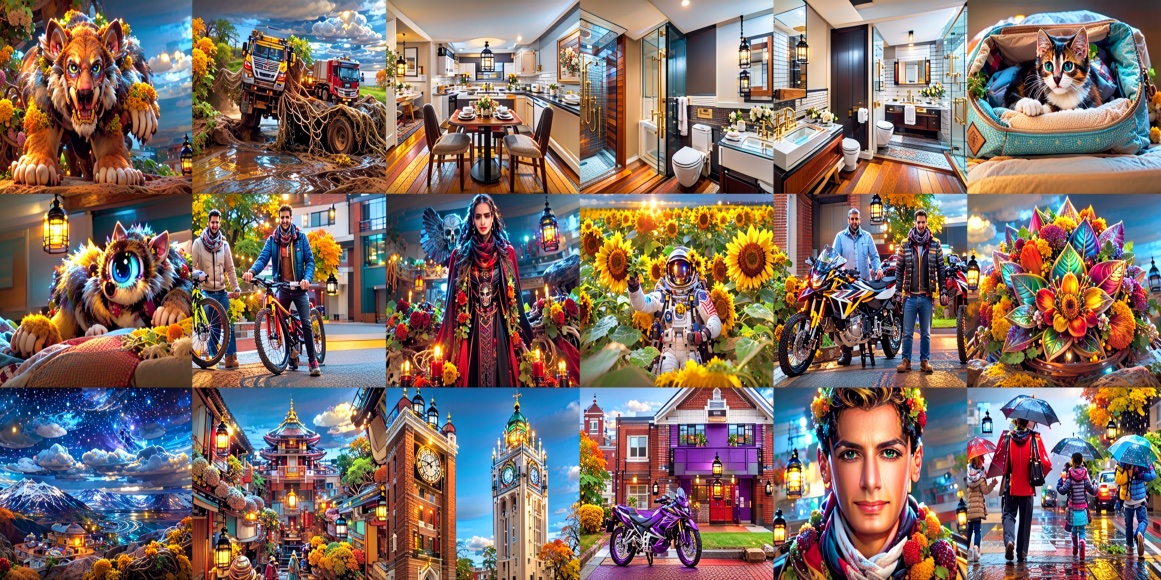}
    \vspace{-0.2cm}
    \caption{Uncurated samples from Stable Diffusion fine-tuned using DRaFT-1 for the HPSv2 reward~\citep{wu2023human}.}
    \label{fig:uncurated-hpsv2}
\end{figure}

\begin{figure}[H]
    \centering
    \includegraphics[width=0.95\linewidth]{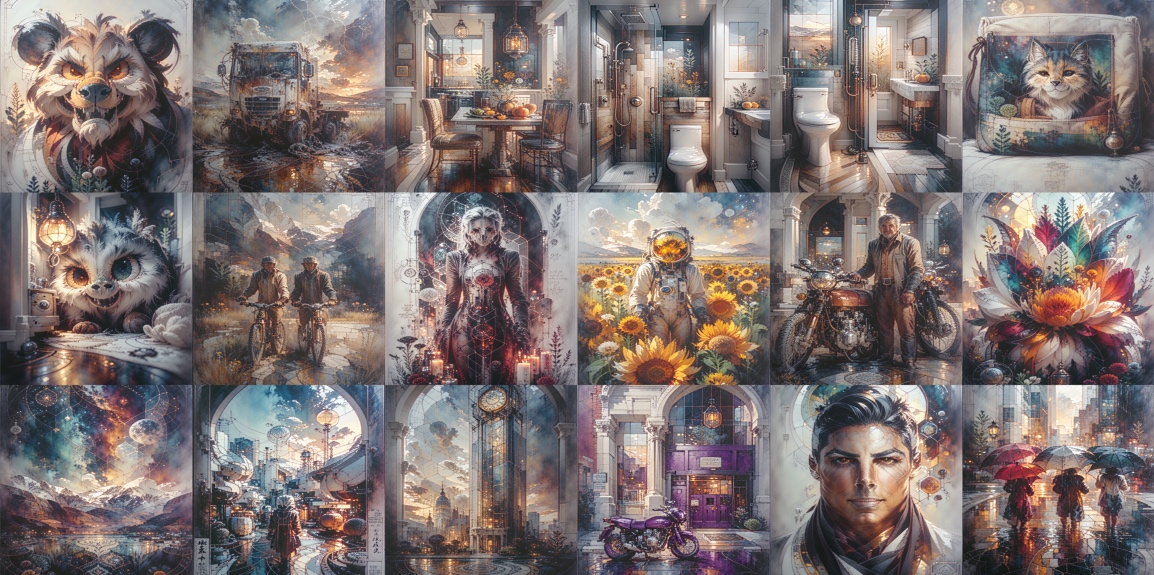}
    \vspace{-0.2cm}
    \caption{Uncurated samples from Stable Diffusion fine-tuned using DRaFT-1 for the PickScore reward~\citep{kirstain2023pick}.}
    \label{fig:uncurated-pickscore}
\end{figure}

\begin{figure}[H]
    \centering
    \vspace{-5mm}
    \includegraphics[width=0.98\linewidth]{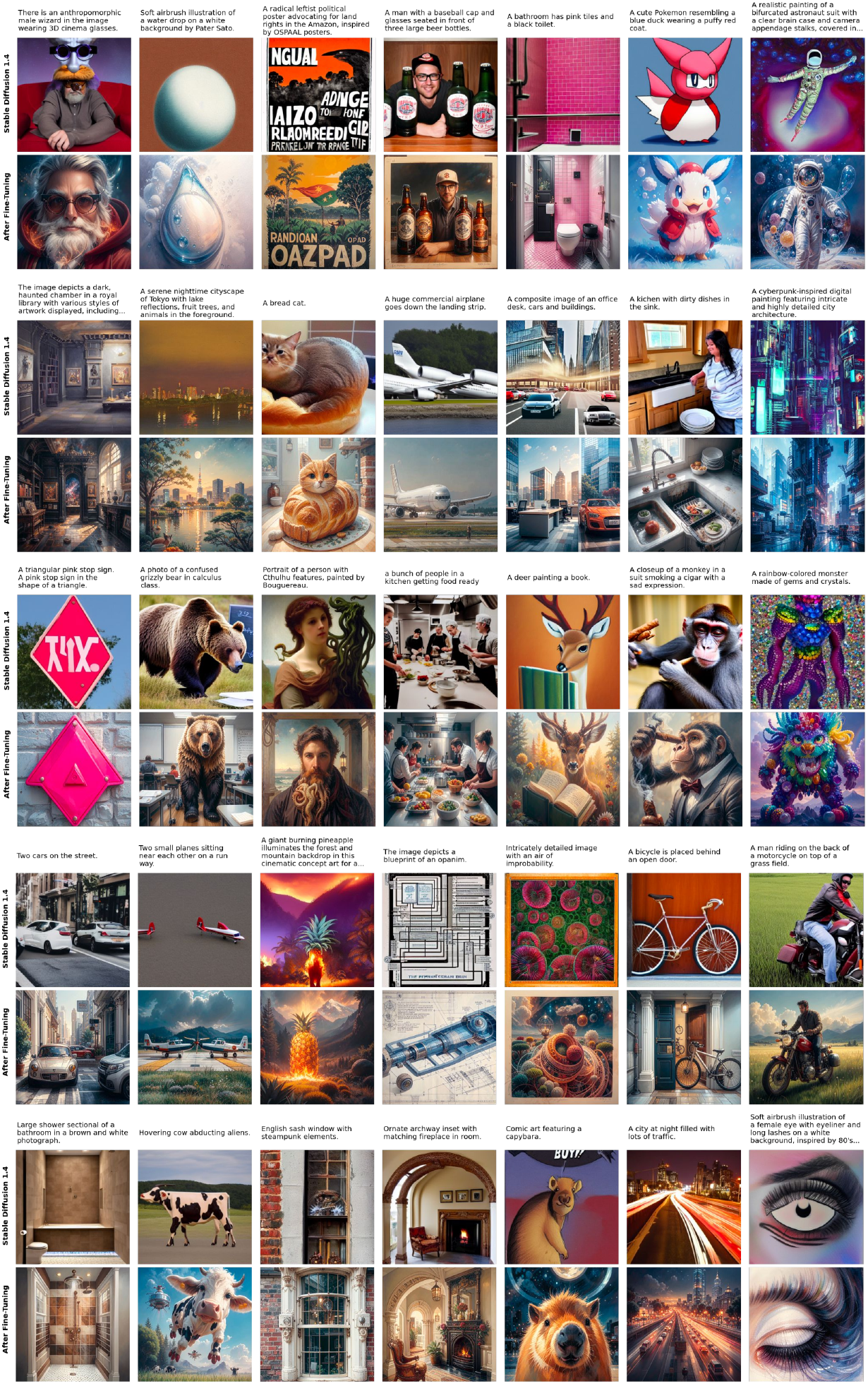}
    \caption{\small Uncurated samples from pre-trained Stable Diffusion 1.4 compared with DRaFT fine-tuned Stable Diffusion.
    We use DRaFT-LV on a combined reward of $\text{PickScore}=10, \text{HPSv2}=2, \text{Aesthetic}=0.05$.
    After training, LoRA weights are scaled down by a factor of 0.75 to reduce reward overfitting.
    Prompts are from the HPSv2 benchmark. }
    \label{fig:uncurated-combined}
\end{figure}

Here, we show uncurated samples from models fine-tuned with DRaFT for different reward functions: Human Preference Score v2 (Figure~\ref{fig:uncurated-hpsv2}), PickScore (Figure~\ref{fig:uncurated-pickscore}), and a combined reward (Figure~\ref{fig:uncurated-combined}).
The first two are overfit to the reward functions, which decreases diversity and photorealism. In the third, we mitigate overfitting through LoRA scaling.

\end{document}